\documentclass{article} 
\usepackage{iclr2027_conference,times}
\usepackage{soul}
\usepackage{amsmath,amssymb,textcomp}
\usepackage{amsthm}
\usepackage{graphicx}
\usepackage{wrapfig}

\makeatletter
    \long\def\masshl#1{\@masshl#1\par\@nil}
    \long\def\@masshl#1\par#2\@nil{%
      \ifx&#1&\else\hl{#1}\par\fi
      \ifx&#2&\else\@masshl#2\@nil\fi
    }
\makeatother

\usepackage{booktabs}
\usepackage{enumitem}

\colorlet{shadecolor}{blue!20}

\usepackage[dvipsnames]{xcolor}

\usepackage[colorlinks=true,allcolors=MidnightBlue]{hyperref}

\DeclareRobustCommand{\mb}[1]{\boldsymbol{#1}}

\DeclareMathOperator*{\argmin}{argmin}
\DeclareMathOperator*{\Tr}{Tr}

\renewcommand{\mid}{~\vert~}
\newcommand{\given}{\,|\,}

\newcommand{\mbi}{\mb{i}}

\newcommand{\mbv}{\mb{v}}
\newcommand{\mbw}{\mb{w}}
\newcommand{\mbx}{\mb{x}}
\newcommand{\mby}{\mb{y}}

\newcommand{\mbC}{\mb{C}}

\newcommand{\mbG}{\mb{G}}
\newcommand{\mbH}{\mb{H}}
\newcommand{\mbI}{\mb{I}}

\newcommand{\mbK}{\mb{K}}

\newcommand{\mbU}{\mb{U}}
\newcommand{\mbV}{\mb{V}}

\newcommand{\mbX}{\mb{X}}
\newcommand{\mbY}{\mb{Y}}

\newcommand{\mbDelta}{\mb{\Delta}}

\newcommand{\E}{\mathbb{E}}
\newcommand{\Var}{\mathbb{V}\textrm{ar}}
\newcommand{\Cov}{\mathbb{C}\textrm{ov}}

\newcommand{\cD}{\mathcal{D}}

\newcommand{\cF}{\mathcal{F}}
\newcommand{\cG}{\mathcal{G}}

\newcommand{\cI}{\mathcal{I}}

\newcommand{\cP}{\mathcal{P}}

\newcommand{\cS}{\mathcal{S}}

\newcommand{\cU}{\mathcal{U}}

\newcommand{\reals}{\mathbb{R}}
\newcommand{\expect}{\mathbb{E}}

\renewcommand{\arccos}{\cos^{-1}}

\newcommand{\plugin}{{\mathrm{pi}}}
\newcommand{\CKA}{\mathrm{CKA}}
\newcommand{\CKD}{\mathrm{CKD}}
\newcommand{\HSIC}{\mathrm{HSIC}}
\newcommand{\Proc}{\mathrm{Proc}}
\newcommand{\song}{\mathrm{sc}}
\newcommand{\sanc}{\mathrm{sanc}}

\newtheorem{theorem}{Theorem}[section]
\newtheorem{lemma}{Lemma}[section]

\newtheorem{proposition}{Proposition}[section]

\newtheorem{corollary}{Corollary}[section]

\makeatletter
\AddToHook{cmd/appendix/before}{\crefalias{section}{appendix}}
\makeatother

\title{Centered Kernel Distance:\\Dimension-Free Error Bounds for Comparing Neural Representations}

\author{%
Shujun Xiong \\
New York University \\
\texttt{jx2837@nyu.edu}
\And
Sarah Harvey \\
Simplex, Astera Institute \\
\texttt{sarah.harvey@astera.org}
\AND
Jacob A. Zavatone-Veth \\
Harvard University\thanks{Present address: The Rockefeller University. Work performed while at Harvard.}\\ 
\texttt{jzavatoneveth@rockefeller.edu}
\And
Alex H. Williams \\
New York University \\
Flatiron Institute \\ 
\texttt{alex.h.williams@nyu.edu}
}

\iclrfinalcopy

\usepackage{cleveref}

\begin{document}

\maketitle

\begin{abstract}
The asymptotic error of representational similarity metrics such as centered kernel alignment (CKA) and representational similarity analysis (RSA) has been well-characterized, but there are no realistic bounds on how estimates of representational similarity $\textit{simultaneously}$ depend on the number of sampled neurons ($N$) and experimental conditions ($M$). 
We study the natural metric counterpart of CKA and RSA, which we call the Centered Kernel Distance (CKD). 
We show that CKD admits an unbiased estimator, whereas several popular similarity measures including CKA have no unbiased estimator. 
Under mild assumptions, we show that estimation error of CKD decays in proportion to $1/M + 1/N$. 
Strikingly, the rate in this bound does not depend on the dimensionality of the space of experimental conditions.
Thus, similarity estimation does not suffer from the curse of dimensionality, which plagues similar problems in high-dimensional statistics. 
We verify our results empirically in random shallow neural networks, trained neural networks, and macaque visual cortex data.
Ultimately, this paper advances a framework for predicting the error of similarity estimators in experiments by relaxing existing, biologically-unrealistic assumptions. Our theoretical and empirical predictions produce specific prescriptions on how to most efficiently estimate the similarity between multiple neural systems.
\end{abstract}
\section{Introduction}

Comparative analysis of neural representations is an active topic of research in neuroscience \citep{kriegeskorte2019peeling} and AI \citep{sucholutsky2023getting, klabunde2025similarity}. In neuroscience, initiatives such as BrainScore or the Algonauts competition apply similarity analyses to understand how computational models align with neural data \citep{schrimpf2018brain, gifford2025algonauts2025}. 
In machine learning, similarity analysis plays an important role in model stitching \citep{BansalNB21}, and more generally in studying how different training and architectural decisions impact learned representations \citep{kornblith2019similarity, raghu2021do,huh2024platonic}.
Practitioners often quantify similarity via scalar scores such as centered kernel alignment \citep[CKA;][]{kornblith2019similarity} and the closely related framework of representational similarity analysis \citep[RSA;][]{kriegeskorte2008representational}---see \citet{williams2024equivalence} for a precise connection between the two.

In practice, representational similarity scores are estimated from limited data.
Even in artificial network models, it is impossible to exhaustively sample most input spaces (\textit{e.g.}, natural images).
In biological systems, we must additionally contend with sub-sampling neural dimensions since experimental recordings only capture a small fraction of neurons within a given brain region  \citep{stevenson2011advances}.
Furthermore, biological responses are noisy and variable from trial-to-trial.
Prior work has separately analyzed the effect of subsampling stimulus inputs \citep{pospisil2023estimating, szekely2014partial} and subsampling neurons \citep{chun2025estimating, kang2025spectral} in representational similarity measures.
However, to our knowledge, a non-asymptotic theory that simultaneously accounts for subsampling along both axes (stimuli and neurons) has not been outlined.
Hence, application of similarity analysis to neural data is difficult, as the statistical theory thus far deals with asymptotic regimes that experimenters are unable to access.
Given the difficulties of recording neural data, can we uncover which experimental parametrizations are amenable for similarity analysis?

Here, we define and analyze the \textbf{Centered Kernel Distance (CKD)}: a distance metric that is closely associated with the popular Centered Kernel Alignment \citep[CKA;][]{kornblith2019similarity} similarity score.
The metric itself is not new; CKD is distance on the inner product space defined by the Hilbert--Schmidt Independence
Criterion (HSIC; \citealt{gretton2005measuring}).
\citet{harvey_what_2024} also analyzed CKD (under a different name), and gave the metric operational meaning by showing that it equals the average distance between decoded signals over random tasks.
We focus on CKD as its structure makes statistical analysis under double-subsampling tractable, revealing several favorable properties:

\begin{itemize}[leftmargin=*]

\item We show that the square of CKD admits an unbiased estimator, $\widehat{\mathrm{CKD}}{}^2$, as a three-sample U-statistic \textit{where inputs as well as neurons are subsampled}.
Alternative measures including CKA and the squared Procrustes distance are not unbiasedly estimable (see Appendix~\ref{app:no-unbiased}).

\item We derive non-asymptotic bounds on $\Var[\widehat{\mathrm{CKD}}{}^2]$ in terms of the number of sampled inputs $M$ and neurons $N_X$ and $N_Y$. Writing $\epsilon := 1/M + 1/N_X + 1/N_Y$, the variance scales as $\mathrm{CKD}^2 \cdot \epsilon + \epsilon^2$. The bound is \emph{signal-adaptive}, tightening as the two representations converge. Interestingly, the dimension of the input space does not appear in this rate. Hence, there is no curse of dimensionality, and $\CKD^2$ can be accurately estimated over complex domains such as natural images even in experiments with limited trials and neuron counts.

\item We show that unbiased estimation of CKD with the same favorable variance decay is possible in the presence of trial-to-trial variability. If noise is correlated across neurons, an extension of the basic U-statistic estimator allows unbiased estimation from two trials per stimulus.

\item Metrics satisfying the triangle inequality enable rigorous comparison of multiple neural systems at once (e.g. three or more animal subjects or artificial network models; see \citealp{williams2021generalized}). 
Recognizing that CKD is built from an inner product between kernels, we leverage classical results \citep[see][]{dokmanic2015euclidean} to perform closed-form metric repair and low-dimensional embedding.

\end{itemize}

    
    

\section{Problem Setup and Background}
\label{sec:background}

\subsection{Sampling model}
\label{subsec:sampling-model}

We model neural responses as functions or ``tuning curves'' mapping inputs
(e.g.\ images) onto activation level. Formally, let $\cF$ and $\cG$ denote
two stochastic processes representing distributions over tuning curves, and
let $P$ denote a distribution over network inputs, which live in a space with some intrinsic dimension $D$. 
Intuitively, differences
in the geometry of neural representations stem from underlying differences
in $\cF$ and $\cG$, measured according to the distribution of inputs
specified by $P$. 
Throughout, we write
\begin{equation}
k(u_1, u_2) = \E_{f \sim \cF}\left[ f(u_1)f(u_2) \right]
\qquad \text{and} \qquad
h(u_1, u_2) = \E_{g \sim \cG}\left[ g(u_1)g(u_2) \right]
\end{equation}
for the covariance operators of $\cF$ and $\cG$.

We analyze an experiment in which $N_X$ and $N_Y$ tuning curves are randomly
sampled from each system, along with $M$ inputs fed to both:
\begin{equation}
\label{eq:sampling-model}
f_1, \dots, f_{N_X} \sim \cF, \quad
g_1, \dots, g_{N_Y} \sim \cG, \quad \text{and} \quad
u_1, \dots, u_M \sim P,
\end{equation}
independently. We thus observe $M \times N_X$ and $M \times N_Y$ random
matrices $X_{mn} = f_n(u_m)$ and $Y_{mn} = g_n(u_m)$. Crucially,
\textit{we assume both axes---neurons and inputs---are randomly sampled},
and our focus is to elucidate how these two sources of sampling error
jointly impact estimates of representational similarity. Note that the $M$
inputs are shared between the two systems, while the neurons are drawn
independently.
While i.i.d. sampling of biological neurons is not necessarily possible in practice, the data generating process in \cref{eq:sampling-model} captures the essence of how neural sampling is treated when testing experimental hypotheses. 

\subsection{Centered Kernel Alignment and Centered Kernel Distance}

Let $\widehat{\mbK}(\mbX) = \frac{1}{N_X}\mbX\mbX^\top$
and $\widehat{\mbH}(\mbY) = \frac{1}{N_Y}\mbY\mbY^\top$ denote the empirical
linear kernel matrices, normalized by the number of sampled neurons in each
system. 
To prevent notational clutter, we treat these matrices as implicit functions, writing $\widehat{\mbK}$ in place of $\widehat{\mbK}(\mbX)$ and $\widehat{\mbH}$ in place of $\widehat{\mbH}(\mbY)$.

\textit{Centered Kernel Alignment} (CKA; \citealt{cortes2012algorithms}) is
a popular framework for quantifying similarity in hidden layer
representations \citep{kornblith2019similarity}. 
Writing $\mbC = \mbI - \tfrac{1}{M}\mb{1}\mb{1}^\top$ for the
$M \times M$ centering matrix, the plug-in linear CKA score is
\begin{equation}\label{eq:def-CKA}
\CKA_\plugin(\mbX, \mbY)
= \frac{S_\plugin(\mbX, \mbY)}
       {\sqrt{S_\plugin(\mbX, \mbX)\,
              S_\plugin(\mbY, \mbY)}}
\quad \text{where} \quad
S_\plugin(\mbX, \mbY)
= \frac{1}{M^2}\Tr\big[\widehat{\mbK}\mbC\widehat{\mbH}\mbC\big] .
\end{equation}
Here $S_\plugin$ is the plug-in HSIC score, which targets the
\emph{Hilbert--Schmidt inner product} between the centered covariance
functions,
\begin{equation}
\label{eq:hsic-limit}
\lim_{\min(M, N_X, N_Y) \to \infty}
S_\plugin(\mbX, \mbY)
= \langle \bar{k}, \bar{h} \rangle_{\mathrm{HS}}
= \E_{u, u' \sim P}\big[\, \bar{k}(u, u')\,\bar{h}(u, u') \,\big],
\end{equation}
where $\bar{k}$ and $\bar{h}$ are the doubly centered kernel functions
\begin{align}
\bar{k}(u_1, u_2) &= k(u_1, u_2) - \E_{a \sim P}\left[ k(a, u_2) \right]
- \E_{b \sim P}\left[ k(u_1, b) \right]
+ \E_{a, b \sim P}\left[ k(a, b) \right],
\label{eq:kbar} \\
\bar{h}(u_1, u_2) &= h(u_1, u_2) - \E_{a \sim P}\left[ h(a, u_2) \right]
- \E_{b \sim P}\left[ h(u_1, b) \right]
+ \E_{a, b \sim P}\left[ h(a, b) \right].
\label{eq:hbar}
\end{align}

Here we focus on a related quantity, which we call the \textbf{Centered
Kernel Distance (CKD)}. Whereas CKA normalizes the HSIC inner product, CKD
is the \emph{distance} that inner product induces:
\begin{equation}
\label{eq:CKD-estimand}
\CKD^2
\;:=\; \big\Vert \bar{k} - \bar{h} \big\Vert_{\mathrm{HS}}^2
\;=\; \langle \bar{k}, \bar{k}\rangle_{\mathrm{HS}}
+ \langle \bar{h}, \bar{h}\rangle_{\mathrm{HS}}
- 2\langle \bar{k}, \bar{h}\rangle_{\mathrm{HS}} .
\end{equation}
We focus on estimating the \emph{squared} distance throughout since it is unbiasedly estimable (see \cref{sec:ckd-theory}).
Substituting \cref{eq:hsic-limit} into \cref{eq:CKD-estimand}, the natural
plug-in estimator is
\begin{equation}
\label{eq:CKD-plugin}
\CKD_\plugin^2(\mbX, \mbY)
= S_\plugin(\mbX, \mbX)
+ S_\plugin(\mbY, \mbY)
- 2\,S_\plugin(\mbX, \mbY),
\end{equation}
which is consistent, since $\CKD_\plugin^2 \to \CKD^2$ as
$\min(M, N_X, N_Y) \to \infty$ by applying the large numbers along each sampling axis.
Notably, the order of the limits does not matter for HSIC or CKD, but \textit{does} in general matter for $\CKA$; jointly taking $M, N_X, N_Y \to \infty$ in CKA can produce different behavior than taking one axis to infinity first \citep[see Appendix~\ref{app:cka-order-of-limits} and][]{chun2025estimating}.
This already hints that $\CKD$ is a more nicely behaved quantity than $\CKA$.

In Appendix~\ref{app:MMD-CKD}, we show that CKD can be viewed as a special case of a maximum mean discrepancy (MMD; \citealp{Gretton2012}) with a quadratic kernel.
This places CKD within a well-developed body of theory, which we expect to be useful beyond the results developed here.
\citet{harvey_what_2024} also previously studied the same quantity as CKD and showed that it can be interpreted as the average difference between linearly decoded signals.
Finally, CKD is closely related to Representational Similarity Analysis (RSA; \citealt{kriegeskorte2008representational}).
In particular, Proposition~1 in \citet{williams2024equivalence} implies that the CKD between $\widehat{\mbK}$ and $\widehat{\mbH}$ is proportional to the distance between doubly centered representational dissimilarity matrices (RDMs) with a squared Euclidean base distance.
Thus, all of our results on CKD immediately extend to an equivalent form of RSA.

The present paper is \emph{not} interested in demonstrating a general preference for CKD over any other metric.
Rather, we seek to precisely characterize how easy it is to estimate this metric with subsampled inputs as well as subsampled feature dimensions (i.e., subsampled neurons).
We show that CKD is statistically well-behaved, motivating its use in datasets with limited samples.

\subsection{Bias-Variance Decomposition and U-statistics}

Our exposition relies on the classical bias-variance decomposition of point estimators.
Given a target estimand of interest, $\theta$, and an estimator, $\widehat{\theta}$, the decomposition of the mean squared error is
\begin{equation}
\E \big[ \big( \theta - \widehat{\theta} \, \big)^2 \big] = \underbrace{\big( \theta - \E \big[ \widehat{\theta} \big] \big)^2}_{\mathrm{bias}^2}  + \underbrace{\E \big[ \big( \widehat{\theta} - \E \big[ \,  \widehat{\theta} \, \big] \, \big)^2 \big]}_{\mathrm{variance}}
\end{equation}
While the bias of several similarity metrics have been previously characterized \citep{pospisil2023estimating,chun2025estimating}, their variances remain underexplored. 
We aim to resolve this issue in $\CKD^2$.

A key step of our analysis is to show that $\CKD^2$ can be estimated as a \emph{U-statistic} \citep[see][for a textbook reference]{lee1990ustatistics}.
At a high level, the goal of a U-statistic is to estimate a parameter that can be written as $\theta = \E [\phi(z_1, \dots, z_k)]$, where $\phi$ is some nonlinear function that is symmetric in its inputs and $z_1, \dots, z_k$ are independent samples from some distribution $P$.
Given $n \geq k$ independent samples from $P$, a one-sample, order-$k$ U-statistic estimator is
\begin{equation}
\widehat{\theta} = \binom{n}{k}^{-1} \sum_{\mbi \in \cI_{n,k}} \phi(z_{i_1}, \dots, z_{i_k})
\end{equation}
where $\cI_{n,k}$ is the set of all length $k$ subsets of unique indices from $\{1, \dots, n\}$.
Since each multi-index, $\mbi \in \cI_{n,k}$, has no repeated elements, the expectation of each term in the sum is $\theta$.
Hence, the U-statistic, being an average of these terms, is also unbiased $\E [\widehat{\theta}] = \theta$.
Moreover, among all unbiased estimators it is generally the optimal one---in the absence of additional assumptions about $P$, the Lehmann–Scheff\'e theorem implies that the U-statistic estimator has minimal variance.
\citet{song2012feature} derived a U-statistic estimator for HSIC under the assumption of a fixed feature space.
\citet{chun2025estimating} extended this estimator to account for sampling feature dimensions (i.e. sampling neural axes).
We leverage these constructions heavily in our analysis.

U-statistic theory was also leveraged in recent work by \cite{soares2026scalable}, who proposed Triplet (TSI) and Quadruplet (QSI) Similarity Indices as new measures of neural representational similarity.
Like most prior work, they considered only subsampling stimuli (i.e., subsampling $M$ not $N$).
Here, we focus on CKD as an unnormalized counterpart to CKA and RSA, both which are currently more established and popular measures than TSI/QSI.


\section{Useful properties of CKD}
\label{sec:ckd-theory}

\subsection{\texorpdfstring{$\CKD^2$}{CKD} admits an unbiased estimator}
\label{sec:estimators}

Many popular measures of representational similarity are incompatible with unbiased estimation.
Indeed, in Appendix~\ref{app:no-unbiased}, we show that no unbiased estimators exist for the squared Procrustes distance or for $\CKA$.
In both cases, the bias can be mitigated (often at the cost of increased variance), but it cannot be entirely eliminated.
See \citet{pospisil2023estimating} for a discussion of bias in the Procrustes distance and \citet{chun2025estimating} for a discussion of bias in $\CKA$.

The plug-in estimator $\CKD_\plugin$ from \Cref{eq:CKD-plugin} is a linear combination of three HSIC estimators, each of which is biased. 
The bias arises from two sources.
First, $S_\plugin$ averages over \emph{all} pairs of inputs, including the coinciding pairs.
Second, the self-terms $S_\plugin(\mbX, \mbX)$ and
$S_\plugin(\mbY, \mbY)$ additionally average over pairs of coinciding neurons.
In both cases, a pair of coinciding samples---the same input twice, or the same neuron twice---compares a quantity to itself, and so contributes a spuriously large value (upward bias).
We propose an unbiased estimator of CKD by combining several existing approaches for bias-correction in HSIC.

To remove the first source, we use a \emph{stimulus-corrected} HSIC estimator due to \citet{song2012feature}:
\begin{equation}
\label{eq:song-HSIC}
S_\song(\mbX, \mbY)
= \frac{1}{M(M-3)}\left[
\Tr\big[\tilde{\mbK}\tilde{\mbH}\big]
+ \frac{\big(\mb{1}^\top\tilde{\mbK}\mb{1}\big)
        \big(\mb{1}^\top\tilde{\mbH}\mb{1}\big)}{(M-1)(M-2)}
- \frac{2}{M-2}\,\mb{1}^\top\tilde{\mbK}\tilde{\mbH}\mb{1}
\right],
\end{equation}
where $\tilde{\mbK}$ and $\tilde{\mbH}$ are versions of $\widehat{\mbK}$ and $\widehat{\mbH}$ with their diagonals set to zero.
This estimator is unbiased in the setting of fixed feature dimensions (i.e. sampling $M$ inputs acrossed fixed neural dimensions), but in our setting of interest the self-terms $S_\song(\mbX, \mbX)$ and $S_\song(\mbY, \mbY)$ incur a bias as described above, but the cross term $S_\song(\mbX, \mbY)$ remains unbiased.

To remove the source of bias in the self-terms, we use a \textit{stimulus-and-neuron-corrected} HSIC estimator due to \citet{chun2025estimating}. We can write this estimator as:
\begin{equation}
S_\sanc(\mbX) = \frac{1}{N_X (N_X - 1)} \sum_{n \neq n'} S_\song(\mbx_n \mbx_n^\top, \mbx_{n'} \mbx_{n'}^\top)
\label{eq:stim-and-neuron-corrected-HSIC}
\end{equation}
where $\mbx_1, \dots, \mbx_{N_X} \in \reals^{M}$ are the columns of $\mbX$.
Intuitively, $S_\sanc$ skips all terms in $S_\song$ in which the same neuron appears twice in order to remove the upward bias.
A direct calculation of $S_\sanc$ as written above requires $O(N_X^2)$ calls to $S_\song$, but in Appendix~\ref{sec:appendix-b} we describe an approach that scales linearly with the number of neurons.
Putting these components together, we arrive at our estimator:
\begin{equation}
\label{eq:ckd-ustat}
\widehat{\CKD}{}^2 = S_\sanc(\mbX) + S_\sanc(\mbY) - 2 S_\song(\mbX, \mbY).
\end{equation}

Constructing an unbiased estimator of similarity is important to answering key questions in the field.
For example, the question of whether two distributions are distinct can be posed as a question of the statistical significance of the estimated similarity score (e.g. CKD, CKA, RSA, etc.).
To evaluate statistical significance, we need a grasp on both the bias and variance of the given estimate.
Prior work has focused on finding an empirical ``zero-point" via bootstrapping to correct for estimators which are biased in the $M$ and/or $N$ axes.
While this approach has been able to address overconfident claims about similarity \citep{huh2024platonic, groger2026revisiting}, it fails to account for fundamentally biased subsampling of stimuli or neural space.
The construction of an unbiased estimator allows us to sidestep these issues entirely, allowing for more accurate and efficient null hypothesis testing.
This is of particular importance in neuroscience experiments, which are heavily limited in both $M$ and $N$ directions, and for whom bootstrapping rather than test stimuli is necessary in evaluation.

\subsection{The variance of \texorpdfstring{$\widehat{\CKD}{}^2$}{CKD} can be explicitly bounded with a favorable rate}

Our main result assumes that neural responses have finite fourth moments for every $u \in \mathrm{supp}(P)$.
This is guaranteed \textit{in vivo} (a neuron cannot fire infinitely) and \textit{in silico} (values overflow). 
\begin{theorem}[Variance Bound, Appendix~\ref{sec:appendix-b}]
\label{th:bound-CKD}
For the estimator given in \Cref{eq:ckd-ustat},
\begin{equation} \label{eq:bound-CKD}
    \mathbb{V}\mathrm{ar} (\widehat{\mathrm{CKD}}{}^2) < C \biggl [ \CKD^2 \biggl ( \frac1M + \frac{1}{N_X} + \frac{1}{N_Y} \biggr )  + \biggl ( \frac1M + \frac{1}{N_X} + \frac{1}{N_Y} \biggr )^2 \biggr ], 
\end{equation}
where $C=c\max(q,q^2)$ for $c$ an absolute constant and 
\begin{equation}
q = \max \Big\{ \sup_u \expect_f\big[(f(u) - \bar f)^4\big], \ \sup_u \expect_g\big[(g(u) - \bar g)^4\big] \Big\}
\label{eq:fourth-moment-defn}
\end{equation}
the maximum fourth-moment of the neural tuning curves. 
\end{theorem}
We provide the proof in Appendix~\ref{app:proof-of-bound-CKD}, and derive a similar scaling law for HSIC with slightly different constants (\Cref{sec:appendix-U-stats-HSIC}.
We show empirical results for HSIC and CKA in \Cref{app:hsic-cka-empirics}).
The proof follows from a construction of $\widehat{\CKD}{}^2$ as a 3-sample U-statistic, followed by standard strategies for decomposing and upper bounding the variance of a U-statistic. 

\textbf{Dimension-independence of \Cref{th:bound-CKD}.}
Let $D$ denote the intrinsic dimension of the input space---e.g., the dimension of the manifold on which natural images are supported.
Many analyses of neural population geometry are cursed by $D$.
For example, estimating a single neural tuning curve, sampled as $f_n \sim \cF$, is a nonparametric regression problem whose estimation error decays proportional to $(1/M)^{2\beta/(2\beta + D)}$ where $\beta$ is a smoothness parameter \citep[see][]{gyorfi2002distribution,tsybakov2009introduction}.
Thus, the number of samples $M$ must increase exponentially to compensate for an increase in $D$, which is the ``curse of dimensionality.''

\Cref{th:bound-CKD} shows that estimating the $\CKD$ between two representations is a fundamentally easier problem: The rate in \cref{eq:bound-CKD} does not degrade as the input dimension increases.
Remarkably, the constant $C$ in \cref{eq:bound-CKD} also has a dimension-independent upper bound under typical conditions.
In particular, any dimension dependence enters through $q$, which can be bounded in at least three ways: 

First, biological neurons have minimal and maximal firing rates.
We can therefore rescale units so that neural responses are bounded---i.e., $f(u) \in [0, 1]$ and $g(u) \in [0, 1]$ for every $u \in \mathrm{supp}(P)$ with probability one, yielding $q \leq 1$.

Second, even if neural activations are unbounded (e.g., ReLU activations in a deep network), we can bound firing rates by incorporating minimal smoothness assumptions.
For example, if the support of $P$ is a bounded set with diameter $\mathrm{diam}(P)$ and neural tuning functions are $L$-Lipschitz with probability one, then $|f(u) - \bar f| \leq  L \cdot \mathrm{diam}(P)$ (and likewise for $g$).
Then, $q \leq L^4 \cdot \mathrm{diam}(P)^4$.
If $\mathrm{supp}(P)$ is a unit cube, then the diameter grows sublinearly, proportional to $\sqrt{D}$.
If a normalization layer is included, then $\mathrm{diam}(P)$ is fully independent of the dimension. 

Third, if $f$ and $g$ are drawn from Gaussian processes with uniformly bounded covariance kernels, then $q$ is bounded even if the support of $P$ is not compact and even if the kernels are not smooth enough to ensure Lipschitzness of $f$ and $g$. Concretely, suppose that $f \sim \mathcal{GP}(0,k)$ with $|k(u,u')| \leq \kappa$ for all $u,u'$. Then, for any $u$, $f(u)-\bar{f}$ is Gaussian with mean zero and variance at most $4 \kappa$ (by the triangle inequality), so we have $\sup_{u} \mathbb{E}_{f}[ (f(u)-\bar{f})^{4} ] = 3 \sup_{u} \mathbb{E}_{f}[ (f(u)-\bar{f})^{2} ]^2 \leq 48 \kappa^{2}$. 

\textbf{Estimating the variance in practice.}
We have not attempted to sharpen the absolute constant $c$ appearing in \Cref{th:bound-CKD}; the primary focus of this paper is to characterize and interpret the rate.
In practice, we recommend estimating the variance of $\widehat{\CKD}{}^2$ by bootstrapping simultaneously over sampled inputs and neurons.
Bootstrap methods are well studied for U-statistics \citep[see][Ch.~5]{lee1990ustatistics} and for data with multiple crossed sampling factors \citep{owen2007pigeonhole,schutt2023statistical}.
One exception is when $\CKD$ is at or near zero: the U-statistic is then degenerate and the bootstrap is unreliable, so testing $\CKD = 0$ requires different approaches such as a permutation test or wild boostrap resampling \citep{Schrab2023}.

\subsection{Estimating CKD in the presence of trial-to-trial noise}
\label{subsec:noise-main}

Neural responses are noisy in biological systems---spike counts can differ dramatically across nominally identical trials (see e.g. \citealt{goris2014partitioning}).
Noise is also sometimes present in artificial networks (e.g. dropout layers), and past work has introduced specialized metrics to compare the geometry of noise across neural systems \citep{duong2023representational,nejatbakhsh2025comparing}.
Here, we focus on comparing the geometry of the average response from limited measurements.
In particular, how does the scale of noise impact our estimate of CKD?

To address this question, we modify the sampling model of \Cref{subsec:sampling-model} to allow for multiple presentations of the same stimulus input.
Let $L \geq 1$ denote the number of repetitions, so that the response of neuron $n$ to input $i$ on repetition $\ell$ is given by $f_n(u_i) + \xi_{ni\ell}$ where $f_n \sim \cF$ is a sampled tuning curve, $u_i \sim P$ is a sampled input, and $\xi_{ni\ell}$ denotes mean zero random noise with finite fourth moments.
In biological experiments, the number of repetitions, $L$, and the number of unique stimuli, $M$, are inversely coupled to each other due to constraints on the experiment's duration.
The ability to keep the repetition number $L$ small therefore allows one to sample a greater number of inputs $M$.


If trial-to-trial noise is independent across neurons, the U-statistic estimator $\widehat{\CKD}{}^2$ turns out to remain unbiased even when $L = 1$ (see \Cref{prop:noise-unbiased} in \Cref{app:noise-theory}), and the variance bound in \Cref{th:bound-CKD} applies essentially verbatim, with the constant $C$ depending on the maximum fourth moments of the observed (noise-corrupted) responses rather than the fourth moments of the underlying tuning curves (see Appendix~\ref{app:noise-variance-upper-bound-note}).
In fact, the constants appearing in the first-order expansion of the variance are completely insensitive to trial-to-trial noise (see \Cref{prop:noise-first-order} in \Cref{app:noise-theory}).
Thus, when accounting for independent noise, the asymptotic rate of convergence for $\Var [\widehat{\CKD}{}^2 ]$ only has an increased constant.

If trial-to-trial noise is correlated across neurons, as is often the case biologically \citep{kohn2005stimulus, ponce2013stimulus}, then an extension of the U-statistic estimator leveraging $L = 2$ trials per condition is necessary and sufficient to fully debias estimates of HSIC and CKD (see \Cref{app:corr-noise}).
The required modification is fairly intuitive and directly analogous to the motivation provided in \Cref{sec:estimators} for the previous U-statistic estimator.
In short, HSIC is formulated as a large sum over pairs of neurons, stimuli, and trials, and we remove bias by skipping any terms that involve coinciding pairs of neural responses from the same system, same stimulus, and same trial number.
The resulting estimator is unbiased with respect to $M, N_X$, and $N_Y$, with a variance law that only differs by a constant factor on the $1/M$ term, with otherwise no change in rate or asymptotic behavior. 

These results have experimental implications, because many large-scale neuroscience datasets use as little as a single presentation per image \citep{Madan2024MacaqueITBench, papale_extensive_2025}.
Our analysis shows that have $L=1$ presentation per stimulus is likely to add some bias, although the magnitude may be negligible if one assumes that noise is nearly decorrelated across neurons.
Practitioners who want a strict guarantee of unbiased estimation should use $L=2$ presentations per stimulus and the associated estimator we describe in \Cref{app:corr-noise}.

\subsection{Closed form metric repair and embedding for multi-systems analysis}
\label{sec:metric-repair}

We have thus far considered estimating the distance between a single pair of neural systems.
However, we are often interested in comparing across multiple systems---e.g. across several network architectures, layers, or subjects---in order to elucidate how variation in the structure of models induces variation in their function.
When comparing a collection of $T$ representations, the $T \times T$ matrix of pairwise distance scores $\mbDelta$ is the object of interest.
When this matrix is constructed with respect to an underlying \textit{metric space}, it contains rich information about the space of representations (e.g. whether systems are clustered into discrete groups or distributed uniformly; see \citealp{williams2021generalized}).
Taken one step further, one can measure geodesic paths and angles between systems to trace how factors such as network depth or data augmentation move a representation through this space \citep{lange2023deep, he2026augmentation}.

Formally, let $\bar{k}_1, \dots, \bar{k}_T$ denote the centered kernel functions of the $T$ systems, and let $\mbG$ be the $T \times T$ Gram matrix based on the HSIC inner product.
Then,
\begin{equation}
\label{eq:gram}
G_{st} = \big\langle \bar{k}_s, \bar{k}_t \big\rangle_{\mathrm{HS}},
\qquad
\Delta_{st} = G_{ss} + G_{tt} - 2 G_{st} ,
\qquad
\text{for all}~t, s \in \{1, \dots, T\}.
\end{equation}
Given observations $\mbX_i \in \reals^{M \times N_t}$ for $t \in \{1, \dots, T\}$, we can form an entrywise unbiased estimate $\widehat{\mbG}$ of $\mbG$ using the techniques described in \Cref{sec:estimators}. Specifically, we take $\widehat{G}_{tt} = S_{\sanc}(\mbX_t, \mbX_t)$ and $\widehat{G}_{st} = S_{\song}(\mbX_s, \mbX_t)$ for $s \neq t$.

Since $\mbG$ is a Gram matrix, it is positive semidefinite.
However, due to sampling noise, $\widehat{\mbG}$ is generally not positive semidefinite.
The corresponding estimate of CKD distances, given by $\widehat{\Delta}_{st} = \widehat{G}_{ss} + \widehat{G}_{tt} - 2\widehat{G}_{st}$, can have negative entries and triangle inequality violations. We can repair this by projecting $\widehat{\mbG}$ onto the feasible cone $\cS_+^T$ of $T \times T$ positive semidefinite matrices, which fortuitously can be done in closed form \citep{higham1988computing}:
\begin{equation}
\label{eq:psd-projection}
\mbG^\star \;=\; \argmin\nolimits_{\mbG \in \cS_+^T}
\big\Vert \mbG - \widehat{\mbG} \big\Vert_F^2 = \mbU \max(\Lambda, 0) \mbU^\top,
\end{equation} %
%
where $\mbU \Lambda \mbU^\top$ denotes the eigendecomposition of $\widehat{\mbG}$ and $\max(\Lambda, 0)$ is applied elementwise to clip negative eigenvalues to zero.
Projection onto a convex set containing the truth is non-expansive, so 
\begin{equation}
\label{eq:contraction}
\big\Vert \mbG^\star - \mbG \big\Vert_F
\;\leq\; \big\Vert \widehat{\mbG} - \mbG \big\Vert_F ,
\end{equation}
for every realization of data. 
The price is that $\mbG^\star$ is no longer element-wise unbiased since the projection operation is nonlinear---thus, we accept a small amount of estimator bias in exchange for a larger reduction in variance.
The improved estimate of the $T \times T$ matrix of CKD values has elements $\Delta^*_{st} = G^*_{ss} + G^*_{tt} - 2 G^*_{st}$.

The eigendecomposition that repairs $\mbG$ in \cref{eq:psd-projection} also provides an embedding.
Writing $\mbV = \mbU \max(\Lambda, 0)^{1/2}$, we have
$\mbG^\star = \mbV\mbV^\top$, so the rows $\mbv_1, \dots, \mbv_T$ of $\mbV$ are coordinates satisfying
\begin{equation}
\label{eq:embedding}
\big\Vert \mbv_s - \mbv_t \big\Vert_2^2 \;=\; \Delta^\star_{st}
\qquad \text{for all } s, t \in \{1, \dots, T\}.
\end{equation}
This is the classical multidimensional scaling (MDS) construction
(\citealt{gower1966distance}; see \citealt{dokmanic2015euclidean} for a modern
review of Euclidean distance matrices). 

It is worth emphasizing that \cref{eq:embedding} is an exact embedding of $\mbDelta^{\star}$.
Thus, the coordinates in $\mbV$ can be directly passed to essentially any method for clustering, dimensionality reduction, regression, or classification without any loss of information.
Alternative metrics including the Procrustes distance and $\arccos \left( \CKA \right)$ satisfy the triangle inequality but do not admit an inner product.
Analyses are then restricted either to approximate MDS embeddings or to methods operating on raw distances, such as hierarchical clustering and nearest-neighbor prediction (as discussed in \citealp{williams2021generalized}).
Closed form solutions for optimal metric repair, as in \cref{eq:psd-projection} for $\CKD$, are also not available for most alternative metrics.

\section{Applications and Experiments}

\subsection{Validation on Random Neural Networks}
\label{sec:random-NNs}


\begin{figure}
    \centering
    \includegraphics[width=1.0\linewidth]{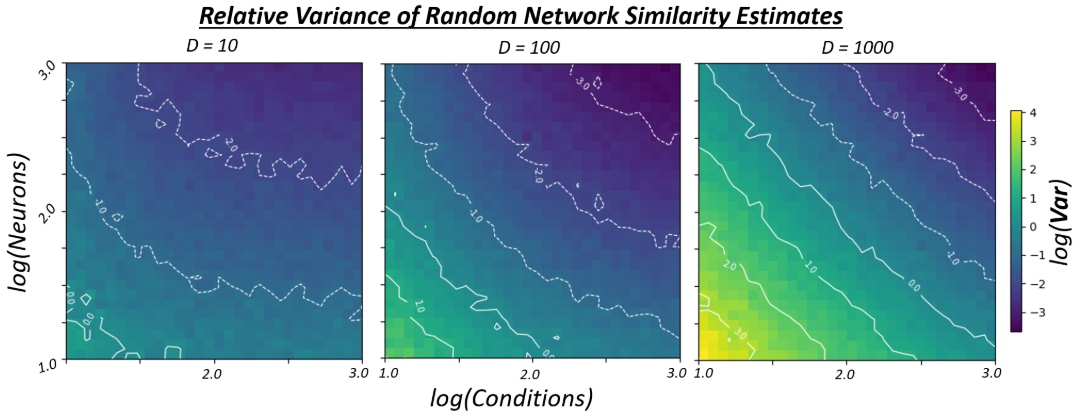}
    \caption{Heatmap of empirical relative variance in estimating CKD between random neural networks for $D = 10, 100, 1000$. Level curves for variance are shown in white. Note that as sampling increases, the relative variance converges across $D$.}
    \label{fig:randomNN}
\end{figure}

    

We first confirm our theoretical predictions in one layer neural networks with random weights (see Appendix~\ref{app:randomNN} for details).
For simplicity, we compare networks of equal width, $N = N_X = N_Y$, but comment on the case of $N_X \neq N_Y$ in Appendix~\ref{app:diffN}.
We report the normalized variance of the U-statistic estimator, $\Var[\widehat{\CKD}{}^2] / \CKD^2$, where the numerator is estimated across multiple simulations and the denominator is given in closed form based on prior work (\citealt{bietti2021deep}).
Equation~\ref{eq:bound-CKD} predicts that this quantity decays at a rate proportional $\mathcal{O}(\frac1M + \frac1N)$.

\Cref{fig:randomNN} shows simulations varying the number of sampled inputs, $M$, and number of sampled neurons, $N = N_X = N_Y$.
Counterintuitively, as $D$ increases, the normalized variance decreases in the top right corner (when $M, N$ are large).
The reason for this is because of the signal adaptivity of the variance.
When the network parameters are fixed, increasing $D$ decreases CKD. 
Hence, larger $D$ in this setting means that the estimator spends more time in the faster rate regime, which allows us to obtain \emph{more} accurate estimates of CKD.
This is most apparent when considering the difference in spacing between the contour lines for $D = 10$ and $D = 1000$.
One can also see the visible widening of the contour lines when $D = 100$ as $M, N \to 1000$.
This is the signal adaptation occurring.

Furthermore, as $D$ increases, \Cref{fig:randomNN} also shows that the contour lines flatten out, transitioning from an apparent $\mathcal{O}(\frac1M + \frac1N)$ scaling law to a $\mathcal{O}(\frac{1}{MN})$ scaling law.
We have two comments on this.
First, we propose that this effect is due to a "zoomed-in" degree of sampling; the "flatter" regions are simply a more local view, relative to the scale of the problem, of an order $\mathcal{O}(\frac1M + \frac1N)$ contour lines.
This is suggested by the slight bending of the contour lines near the edges of the heatmaps.
Second, this still verifies that our upper bound holds, as $\mathcal{O}(\frac{1}{MN})$ scales faster than the upper bound of order $\mathcal{O}(\frac1M + \frac1N)$.
Our proof in \Cref{sec:ckd-theory} accounts for this: the reason for this ``flatness'' is because the coefficients of the leader order terms, which depend on the covariance statistics of the neural tuning curves (\Cref{sec:appendix-b}), vanish, leaving second-order coefficients on the $\frac{1}{MN}$ term to dominate.

\subsection{Validation on Trained Deep Artificial Networks and Biological Data}
\label{sec:deep-NNs}

We consider the error in estimating similarities in deep neural networks by training deep convolutional networks on the ImageNet dataset \citep{deng2009imagenet} and evaluating the variance of their pairwise similarity estimates (see Appendix~\ref{app:model-training} for details).
In Fig.~\ref{fig:models-monkeys}A, we verify that the $\mathcal{O}(\frac1M + \frac1N)$ scaling law still holds. 
While we cannot explicitly modify the dimensionality of the input space, we can study this by proxy by presenting different distributions of test images. 
Evaluating the variance in similarity estimates on test images from the ImageNet dataset (i.e. in-distribution) versus images from the MacaqueITBench dataset (i.e. out-of-distribution; see Appendix~\ref{app:datasets} for details) reveals similar scaling for CKD.
Hence, in line with our theoretical results, the difficulty in estimating the similarity between two models is largely independent of the dataset over which evaluation occurs or changes in the underlying similarity estimate.


To validate our predictions on biological data, in Fig.~\ref{fig:models-monkeys}(B) we measure $\Var(\CKD)$ across regions between two monkeys in the THINGS Ventral Stream Dataset (\cite{papale_extensive_2025}; see Appendix~\ref{app:activation-extraction} for details). Again, $\Var(\CKD)$ follows evenly-spaced L-shaped level curves. 
We use the different regions as a proxy for studying the effect of dimensionality, under the assumption that the effective dimensionality likely varies across the visual hierarchy. 
The scale of variance in all three region-to-region comparisons is consistent, as our theory predicts. This holds also for comparisons between regions (e.g. V4 to IT; see Appendix~\ref{app:cross-region}), and for different image classes (Appendix~\ref{app:datasets}).
Overall, our results suggest a central experimental prescription: the most efficient way to reduce error in CKD estimates (up to cost of recording) is to keep $M \approx N$.
Conversely, increasing sampling in $M$ but not $N$ (and vice versa) will eventually fail to substantially reduce error.

Last, in Fig.~\ref{fig:models-monkeys}C, we to verify the signal-adaptivity of our bound.
We show close adherence of empirical behavior to the signal adaptation in our theory: when $\frac1M + \frac1N$ is small, scaling decelerates.


\begin{figure}
    
    \centering
    \includegraphics[width=1.0\linewidth]{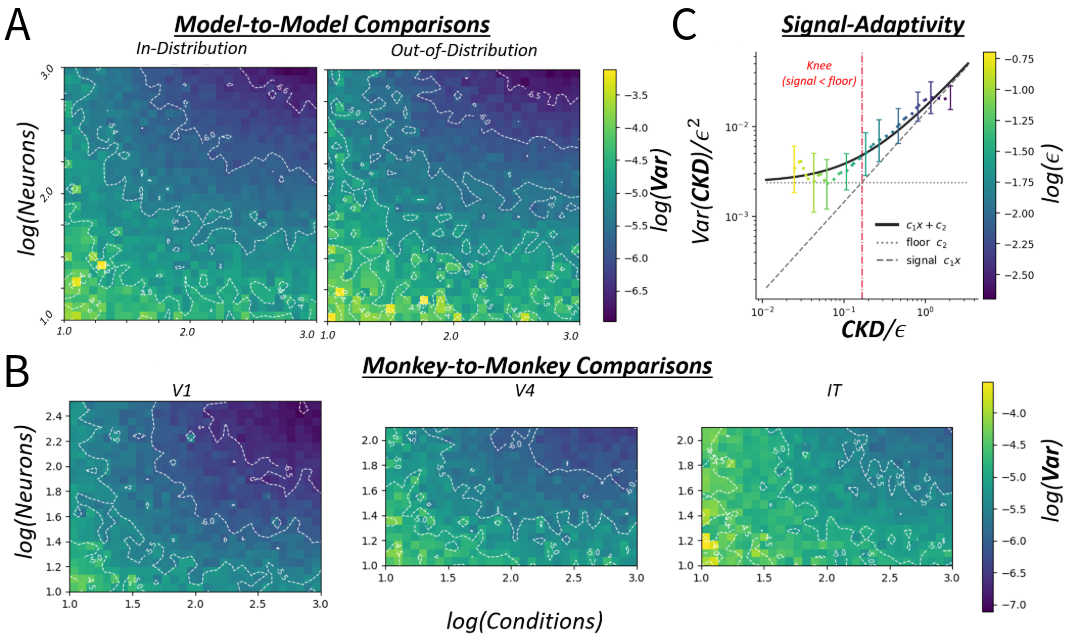}
    \caption{Variance of similarity estimates in deep networks and monkeys. \textbf{(A)} Models presented in-distribution versus out-of-distribution test images. Variance scales independent of dataset choice. \textbf{(B)} Variance of CKD across three brain regions of two monkeys scales independently from region identity. Multi-units are considered distinct ``neurons.'' \textbf{(C)} Representative sample of signal-adaptivity of variance, with decreased rate of variance reduction when $\frac1M + \frac1N$ small.}
    \label{fig:models-monkeys}
    
\end{figure}

\subsection{Similarity Estimates across Multiple Systems}
\label{sec:TVSD}

We simultaneously study the relationships between representations formed by 3 brain regions replicated across 2 monkey subjects \citep{papale_extensive_2025} and 4 trained neural networks from Section~\ref{sec:deep-NNs}.
We might expect brain regions V1, V4, and IT to have distinct representations because they fall along different parts of the visual hierarchy.
However, poorly parametrized sampling regimes may make evaluating this hypothesis impossible.
Furthermore, we don't know the exact variability that different neural regions and models exhibit.
Our theory suggest that $M \approx N$ most efficiently reduces pairwise estimation error.
Do this hold when comparing multiple systems simultaneously?

\begin{figure}    
    \centering
    \includegraphics[width=1\linewidth]{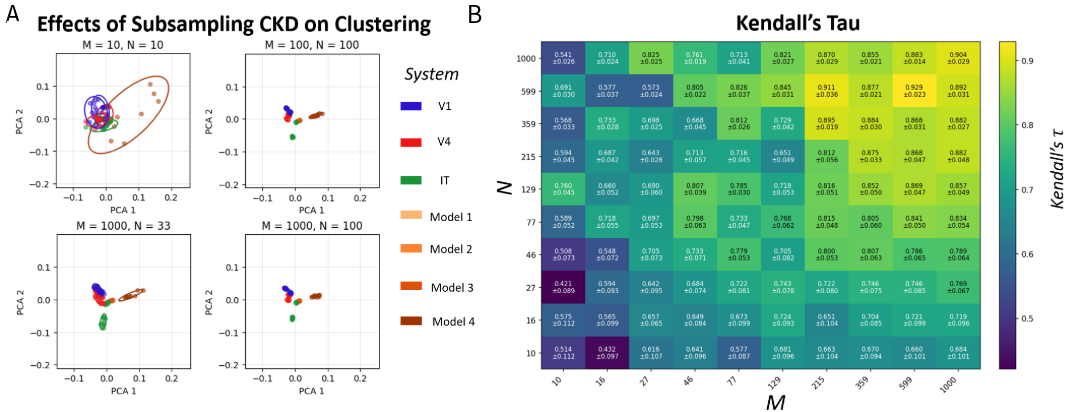}
    \caption{Comparisons between monkeys and similarity analysis between multiple neural systems simultaneously. (\textbf{A}) CKD between 4 deep models and 6 neural recordings from 3 different brain areas. MDS is performed with 10 embedding dimensions, then points are projected into the same PC space across the panels. Each ellipse denotes an estimated region in PC space of a single neural system's random subsampling at certain values of $M$ and $N$. (\textbf{B}) Estimates of Kendall's Tau on different degrees of subsampling, measuring the error in rankings of similarity between all subjects.}
    \label{fig:clusters}
\end{figure}

In Fig.~\ref{fig:clusters}A we produce low-dimensional Euclidean embeddings using multidimensional scaling to visualize the effect of subsampling on the estimated distances among systems 
(see procedure in Appendix~\ref{app:tvsd-sampling}).
Overlaps in ellipsoids indicate probable scenarios where false equivalences may be drawn between two neural systems, while highly elongated ellipsoids indicate probable circumstances where differences between systems would be falsely overstated.

When $M$ and $N$ are small, estimates of distance are noisy and the estimated relationships between neural systems are variable depending on the subsample. 
Sampling more images and neurons significantly reduces this uncertainty. 
Remarkably, however, we can attain \textit{even better} estimates of relationships even when significantly undersampling the number of stimuli ($M = 100$) by simply mildly increasing the number of recorded neurons ($N = 100$). 
This is in line with our prior prescription that the most efficient way to reduce error in estimating similarity is to keep $M \approx N$ (up to costs of additional recording in either axis), and that oversampling only $M$ but not $N$ (and vice versa) does not significantly improve one's estimate. 

This result holds in pre-projection space as well.
In Figure~\ref{fig:clusters}B, we analyze Kendall's $\tau$ measured when subsampling across multiple models.
This measures the divergence between the estimated rankings of similarity between multiple models with the true rankings of similarity between models in the pre-projection distance estimates.
A larger Kendall's $\tau$ indicates greater alignment of similarity ranking between subsampled and true measurements, and thus a more accurate read on the pairwise relationships between models.
Similar to our previous results, a large Kendall's $\tau$ is attained when $M \approx N$, achieving approximately equal or better performance than when only either $M$ or $N$ has a degree of magnitude more samples for marginal trade-off along the other axis.

\section{Discussion}

We show that the CKD exhibits a range of favorable statistical properties with respect to sampling of neurons and stimuli: it can be estimated without bias, its variance decays with a dimension-independent rate, and it enables principled metric repair and embedding.
To our knowledge, no such theory exists to account for the simultaneously subsampling of both stimuli $M$ and neurons $N$.
Our work represents a first step towards resolving this gap in the literature, as we also show that more common measures such as CKA lack such utile properties (see Appendix~\ref{app:cka_differences}).

Our goal is not to advocate for CKD as a novel or ``better'' similarity measure than others.
We agree with \cite{klabunde2025similarity} that practitioners should ideally utilize multiple measures in conjunction with one another.
Our focus on CKD estimators is a reflection of their mathematical tractability, which enabled a detailed account of its statistical properties.
We currently do not have a similarly comprehensive analysis of alternative metrics such as CKA and the Procrustes distance, but it would be highly valuable to pursue this in future work.

Our results have important implications for experimental design. 
Crucially, our theory demonstrates that one's ability to estimate CKD from limited samples does not rapidly degrade as a function of task complexity. 
That is, estimation does not suffer from a curse of dimensionality, unlike related estimation problems in high-dimensional geometry, thus underwriting confidence in similarity estimates for complex stimuli spaces.

\paragraph{Limitations.}
While CKD's lack of normalization provides it powerful properties for analysis, its sensitivity to the scale of a neural representation may be argued to be an undesirable property \citep[see e.g.][]{kornblith2019similarity}.
Incorporating normalization into analyses of CKD and related measures is an important direction of future research.
Simultaneously, more work is required to characterize and improve estimation of metrics like CKA that cannot be formulated as U-statistics. 

\section*{Acknowledgments}
We would also like to thank Ansh Soni for useful conversations in assistance of this work.
SX was supported by the NSF Graduate Research Fellowship Program and the NYU MacCracken Fellowship. 
SH was supported by the Flatiron Institute. 
JAZ-V was supported by the Office of the Director of the NIH under Award Number DP5OD037354, and by a Junior Fellowship from the Harvard Society of Fellows.
AW was supported by NIH NIDA R01DA066648 and the McKnight Foundation.

\clearpage
\section{AI Disclosure}
In this work, we used generative AI tools (LLMs; Claude Opus 5.5, GPT 5.6 Sol, GPT 6 Astra) substantially in developing the mathematical results and providing feedback on research methodology.
The authors derived an initial bound on the same order for HSIC without the assistance of LLMs, and the inspiration to formulate our claims via U-statistics was also done without LLMs.
LLMs were used to assist in learning the deeper mathematical properties of U-statistics as an accompanying guide to \citet{lee1990ustatistics}.
Afterwards, LLMs assisted in formulating precise statements of technical claims for generalization after the authors had specified the quantity to be analyzed and the kind of result sought (a bound on the same order as the initial approach) and in drafting and revising the written proofs.
The authors heavily revised and edited all mathematical proofs and extended the LLM-generated output to cover background material and broader context.

The conceptual framing of the work is due to the human authors. This includes the motivating problem of how similarity estimates depend jointly on the number of sampled neurons and stimuli, the choice to study the Centered Kernel Distance and its unbiased U-statistic estimator, and the decision of which properties to establish. 
The interpretation of all theoretical and empirical results, including their implications for experimental design in neuroscience and machine learning, is likewise due to the human authors.
We did not use generative AI tools to propose hypotheses or design experiments, but LLMs were used to generate code for simulations and aid in formatting figures.
Translation and qualitative data analysis are not applicable to this work.

We have reviewed all AI-assisted work. 
All AI-assisted derivations were checked line-by-line by all the authors.
The resulting bounds were independently checked against numerical simulations in random networks, trained networks, and macaque recordings (Section~4); similarly, AI-generated code was checked against our manual implementations and validated by the numerical experiments in these simulations.
We take responsibility for the final content of this work, including text, claims, and artifacts produced with the aid of generative AI.

\clearpage
\bibliography{iclr2027_conference}
\bibliographystyle{iclr2027_conference}


\clearpage
\appendix

\section{Appendix Table of Contents}

Here, we provide an overview of the appendix sections to orient the reader.

\Cref{sec:appendix-U-stats-HSIC} constructs a U-statistic for HSIC and derives an upper bound on the variance.
This section also outlines the core properties of U-statistics and how to apply known theory about the variance of U-statistics.
For those unfamiliar with U-statistics, or hope to apply our approach elsewhere, we recommend this section, which is written with such readers in mind.

\Cref{sec:appendix-b} constructs a U-statistic for CKD and proves \cref{th:bound-CKD}.

\Cref{app:noise-theory,app:corr-noise} provide theoretical analysis of the CKD estimator in the presence of noise, supporting the results outlined in \cref{subsec:noise-main}.
In \Cref{app:noise-theory}, we assume that noise is independent across neurons and trials.
In \Cref{app:corr-noise} we partly relax this assumption by allowing for ``noise correlations'' across neurons.
The take home message of these sections is that unbiased estimation of CKD is still possible and the dimension-free rate of $(1/M + 1/N_X + 1/N_Y)$ proved in \cref{th:bound-CKD} carries over in the presence of noise.


In \Cref{app:cka_differences}, we present auxiliary results on CKA and the squared Procrustes distances to highlight points of contrast with CKD.
In particular, we show that CKA and Procrustes do not admit unbiased estimators, and that the consistency of the plug-in estimator of CKA depends on the order that one takes the limits of $M$ and $N$ to infinity.
We also provide some experiments to document the empirical behavior of HSIC and CKA in estimators in \Cref{app:hsic-cka-empirics}.


In \Cref{app:diffN}, we document the empirical behavior of CKD when the number of neurons is mismatched across systems, $N_X \neq N_Y$.
Our theory accounts for this case, but the experiments shown in the main paper assume $N_X = N_Y$ to simplify the exposition.

To show generality of our results regardless of region identity in pairwise CKD estimates, we plot the variance of our CKD estimate for all possible pairs of regions within and between monkeys in \Cref{app:cross-region}

Last, to show that our results work across biology and architecture, in section \cref{app:model-to-monkey} we plot the variance of CKD between the trained deep networks and macaque visual cortex data. Whe show that our theoretical predictions still hold.

\Cref{app:randomNN} details how we constructed the random neural network experiments in \cref{sec:random-NNs}.

\Cref{app:datasets} details the image datasets we used in \cref{sec:deep-NNs}, and the differences in their structure. 
We also present results showing that our scaling law holds even when conditioning on specific image categories.

\Cref{app:model-training} details the model architecture and training of the models we evaluate in \cref{sec:deep-NNs}.

\Cref{app:tvsd-sampling} details the clustering procedure we apply in \cref{fig:clusters}.

\section{U-statistic estimators of HSIC}
\label{sec:appendix-U-stats-HSIC}

We begin with a formalization of HSIC as a U-statistic.
Following the construction of such an estimator, we then follow standard arguments on the variance of U-statistics \citep{lee1990ustatistics} to derive a dimension-free upper bound.
In Appendix~\ref{sec:appendix-b}, the techniques applied for HSIC will be re-applied, albeit in a more complex fashion, for the CKD estimator. 
For those unfamiliar with the theory of U-statistics, this is a useful primer. 

\subsection{A finite-sample estimator}



Recall the definition of the inner product between two symmetric functions $k$ and $h$ as
\begin{equation*}
\langle k, h \rangle_{\mathrm{HS}} = \E [k(u, u') h(u, u')]
\end{equation*}
where the expectation is over two independent samples $u$ and $u'$ from a distribution $P$ (which, in our case, is the distribution over inputs to the network).
Plugging in $\bar k$ and $\bar h$ as defined in \cref{eq:kbar,eq:hbar} and expanding leads to a sixteen term expression with many cancellations.
Ultimately, the expression simplifies to
\begin{equation}
\langle \bar{k}, \bar{h} \rangle_{\mathrm{HS}} = \expect_{u_1, u_2, u_3, u_4} [k(u_1, u_2) h(u_1, u_2) + k(u_1, u_2) h(u_3, u_4) - 2 k(u_1, u_2) h(u_1, u_3)] 
\label{eq:HSIC-uncentered-kernels}
\end{equation}
which is equivalent to Equation 3 in \citet{song2012feature}.
Thus, the HSIC estimand can be represented as an expectation over four independent samples, $u_1, u_2, u_3, u_4 \sim P$, of an expression based on the uncentered kernel functions.


Here, we are interested in comparing kernels that are defined by neural tuning curve distributions, $\cF$ and $\cG$, over the space of functions mapping $\cU \mapsto \reals$.
Specifically, we define
\begin{align}
k(u_1, u_2) &= \expect_{f \sim \cF} [ f(u_1) f(u_2) ] \\
h(u_1, u_2) &= \expect_{g \sim \cG} [ g(u_1) g(u_2) ]
\end{align}
Now plugging these equations into \cref{eq:HSIC-uncentered-kernels}, we obtain
\begin{equation}
\langle \bar{k}, \bar{h} \rangle_{\mathrm{HS}} = \expect_{f, g, u_1, u_2, u_3, u_4} \left[
\begin{aligned}
    &f(u_1) f(u_2) g(u_1) g(u_2) \\
    &+ f(u_1) f(u_2) g(u_3) g(u_4) \\
    &- 2 f(u_1) f(u_2) g(u_1) g(u_3)
\end{aligned}
\right]
\end{equation}
Let $\Pi$ denote the set of all $4!$ permutations of $(1, 2, 3, 4)$, and define
\begin{equation}
\phi(f, g, u_1, u_2, u_3, u_4) =
\frac{1}{4!} \sum_{(i,j,\ell,p) \in \Pi}
\left[
\begin{aligned}
    &f(u_i) f(u_j) g(u_i) g(u_j) \\
    &+ f(u_i) f(u_j) g(u_\ell) g(u_p) \\ 
    &- 2 f(u_i) f(u_j) g(u_i) g(u_\ell)
\end{aligned}
\right]
\label{eq:hsic-u-statistic-kernel-defn}
\end{equation}
Note that $\phi$ is symmetric in its final four arguments (i.e. invariant to permuting the order of $u_1$, $u_2$, $u_3$, and $u_4$).
It is additionally symmetric in its first two arguments, $\phi(f,g;\cdot)=\phi(g,f;\cdot)$, since swapping $f$ and $g$ induces a permutation over the input indices which is averaged out.

We assume that we have access to $N$ independently sampled neural tuning curves from each system, $f_1, \dots, f_{N_X} \sim \cF$ and $g_1, \dots, g_{N_Y} \sim \cG$, alongside $M$ independently sampled inputs, ${u_1, \dots, u_M \sim P_{\cU}}$.
Then,
\begin{equation}
S_\song = \frac{1}{N_XN_Y} {\binom{M}{4}}^{-1}  \sum_{n=1}^{N_X} \sum_{n'=1}^{N_Y} \sum_{i<j<\ell<p} \phi(f_n, g_{n'}, u_{i}, u_{j}, u_{\ell}, u_{p})
\label{eq:hsic-as-a-u-statistic}
\end{equation}
is an unbiased estimator of $\textrm{HSIC}_{\infty}$.
Indeed, it is a \emph{three-sample U-statistic} of degree $(1,1,4)$: {order one in each of the $\cF$ and $\cG$ samples, and order four in the inputs.
Note that here we investigate specifically the stimulus-corrected est}imator $S_\song$.
Neuron-corrections for bias is necessary when the underlying distributions $\cF, \cG$ are identical (e.g. comparing data from two sessions in the same animal-region). 
Fortunately, our approach in this section can be similarly applied to a U-statistic for $S_\sanc$.


\subsection{The variance as a sum over collisions}

The properties of U-statistics have been extensively characterized \citep{lee1990ustatistics}.
In particular, we can write down an explicit expression for the variance of the estimator.
To do this it is convienent to package a set of inputs into a single symbol:
\begin{equation}
    \sigma = \{ f_n, g_{n'}, u_i, u_j, u_\ell, u_p \}
\end{equation}
Then \cref{eq:hsic-as-a-u-statistic} becomes
\begin{equation}
    S_\song = \frac{1}{N_XN_Y} {\binom{M}{4}}^{-1}  \sum \phi(\sigma)
\end{equation}
where the sum is understood to range over all unique tuples.
Thus, we can write the variance of the estimator as
\begin{equation}
\Var [ S_\song ] = \Var \left [ \frac{1}{N_XN_Y} {\binom{M}{4}}^{-1}  \sum \phi(\sigma) \right ] = \frac{1}{N^2_XN^2_Y} {\binom{M}{4}}^{-2} \sum_{\sigma, \sigma^\prime} \Cov(\phi(\sigma), \phi(\sigma^\prime))
\label{eq:hsic-variance-giant-sum}
\end{equation}
where the sums over $\sigma$ and $\sigma^\prime$ independently span over all unique tuples.

Next, we invoke a standard identity for U-statistic covariances \citep[Section 1.4, Theorem 3,][]{lee1990ustatistics}:
\begin{equation} 
\Cov(\phi(\sigma), \phi(\sigma^\prime)) = \Var_{\sigma \cap \sigma^\prime}\Big[ \, \expect \big[ \, \phi(\sigma) \, \big| \, \sigma \cap \sigma^\prime \, \big] \, \Big] .
\label{eq:collision-covariance}
\end{equation}
Intuitively, this relation states that the covariance between $\phi(\sigma)$ and $\phi(\sigma^\prime)$ grows with the size of their overlapping elements $\sigma \cap \sigma^\prime$.
The inner expectation denotes partial averaging---all variables that are \emph{not overlapping} are averaged out, and the variance is computed with to respect to the remaining terms.
When $\sigma \cap \sigma^\prime = \emptyset$ (no overlap), the inner expectation is with respect to all variables in $\sigma$ and evaluates to a constant; thus, we have:
\begin{equation}
\label{eq:no-collisions-implies-zero-variance}
\sigma \cap \sigma^\prime = \emptyset \quad \Rightarrow \quad \Cov(\phi(\sigma), \phi(\sigma^\prime)) = 0.
\end{equation}
In the other extreme, when $S  = S'$ (full overlap), nothing is averaged out in the inner expectation and we recover $\Cov(\phi(\sigma), \phi(\sigma^\prime)) = \Var[\phi(\sigma)]$.

To proceed, we enumerate all possible overlap patterns between $\sigma$ and $\sigma^\prime$ in the sum appearing in \cref{eq:hsic-variance-giant-sum}.
We use the term ``collison'' to denote the event that one of the variables is part of the overlap and we introduce three index variables to account for these collisions:
\begin{align}
\alpha \in \{0, 1\} ~ &= ~ \text{\# of collisions in $\cF$ samples} \\
\beta \in \{0, 1\} ~ &= ~ \text{\# of collisions in $\cG$ samples} \\
\gamma \in \{0, 1, 2, 3, 4\} ~ &= ~ \text{\# of collisions in input samples}
\end{align}
For example, $\sigma = (f_1, g_1, u_1, u_2, u_3, u_4)$ and $\sigma^\prime = (f_1, g_2, u_2, u_3, u_4, u_5)$ corresponds to a overlap pattern of $(\alpha, \beta, \gamma) = (1, 0, 3)$; one collision in the $\cF$ neural sample, no collisions in the $\cG$ neural sample, and three collisions in the input samples.
Note that the \emph{position} of the four input samples can be ignored since $\phi$ is a symmetric function by our construction in \cref{eq:hsic-u-statistic-kernel-defn}.

Now let $C_{\alpha \beta \gamma}$ denote the number of collisions of each type occuring in \cref{eq:hsic-variance-giant-sum}, and let
\begin{equation}\label{eq:Cov-Var-U-stats}
V_{\alpha \beta \gamma} = \Cov(\phi(\sigma), \phi(\sigma^\prime)) = \Var_{\sigma \cap \sigma^\prime} \Big[ \, \expect \big[ \, \phi(\sigma) \mid \sigma \cap \sigma^\prime \, \big] \, \Big],
\end{equation}
denote the variance of each corresponding term.
Note that the variance only depends on the number of collisions in each block of variables due to the i.i.d. assumption of samples and the blockwise symmetry of $\phi$.
Thus, we can compactly re-express the variance of the estimator as:
\begin{equation}
\Var [ S_\song ] = \frac{1}{N^2_XN^2_Y} {\binom{M}{4}}^{-2} \sum_{\alpha, \beta, \gamma} C_{\alpha \beta \gamma} V_{\alpha \beta \gamma}
\end{equation}
where the sum ranges over all $2 \times 2 \times 5 = 20$ possible collision patterns.
Note that, for large $N$ and $M$, the vast majority of terms will have no collisions, and these terms do not contribute to the variance since $V_{000} = 0$ as already established in \cref{eq:no-collisions-implies-zero-variance}.

\subsection{Counting collisions}

We now derive an explicit expression for $C_{\alpha \beta \gamma}$ using some basic combinatorial calculations.
For each block of variables ($\cF$ neurons, $\cG$ neurons, and inputs), both $\sigma$ and $\sigma^\prime$ draw a subset of fixed size $\nu$ without replacement from a population of $P$ elements ($\nu = 1$, $P = N$ for each neuron sample; $\nu = 4$, $P = M$ for the inputs).
Conditioned on whichever $\nu$ elements are drawn by $\sigma$, the overlap follows a hypergeometric distribution---i.e. the number of times one of the $\nu$ elements in $\sigma$ is drawn by $\sigma^\prime$ for its own $\nu$ draws without replacement.
The number of ordered pairs of subsets that overlap in exactly $c$ elements is therefore
\begin{equation}
\underbrace{{\binom{P}{\nu}}}_{S\text{'s subset}} \; \underbrace{{\binom{\nu}{c}} {\binom{P-\nu}{\nu-c}}}_{\text{hypergeometric weight for } S'} ,
\label{eq:hypergeometric-count}
\end{equation}
where ${\binom{\nu}{c}}$ counts the marked elements that $\sigma^\prime$ reuses and ${\binom{P-\nu}{\nu-c}}$ the remaining ones it draws from the unmarked population; dividing through by ${\binom{P}{\nu}}$ recovers the hypergeometric probability mass function.
Multiplying \cref{eq:hypergeometric-count} across the three samples---with $(\nu, P, c) = (1, N, \alpha)$ for $\cF$, $(1, N, \beta)$ for $\cG$, and $(4, M, \gamma)$ for the inputs---yields
\begin{equation}
C_{\alpha \beta \gamma} = \underbrace{\binom{N_X}{1} \binom{1}{\alpha} \binom{N_X-1}{1-\alpha}}_{\cF} \; \underbrace{\binom{N_Y}{1} \binom{1}{\beta} \binom{N_Y-1}{1-\beta}}_{\cG} \; \underbrace{\binom{M}{4} \binom{4}{\gamma} \binom{M-4}{4-\gamma}}_{\text{inputs}} .
\end{equation}
This simplifies simply to 
\begin{equation}
C_{\alpha \beta \gamma} = N_X (N_X-1)^{1-\alpha} \; N_Y (N_Y-1)^{1-\beta} \; \binom{M}{4} \binom{4}{\gamma} \binom{M-4}{4-\gamma} .
\label{eq:collision-counts-exact}
\end{equation}
It is helpful to normalize $C_{\alpha \beta \gamma}$ by the number of terms in the sum.
Thus, we have
\begin{equation}
\Var [ S_\song ] = \sum_{\alpha, \beta, \gamma} \widetilde{C}_{\alpha \beta \gamma} V_{\alpha \beta \gamma}
\label{eq:hsic-variance-normalized-Ctilde}
\end{equation}
where
\begin{equation}
\widetilde{C}_{\alpha \beta \gamma} = \frac{1}{N^2_XN^2_Y} {{M}\choose{4}}^{-2} C_{\alpha \beta \gamma} = \underbrace{\frac{(N_X-1)^{1-\alpha}}{N_X}}_{\cF} \; \underbrace{\frac{(N_Y-1)^{1-\beta}}{N_Y}}_{\cG} \; \underbrace{\frac{{{4}\choose{\gamma}} {{M-4}\choose{4-\gamma}}}{{{M}\choose{4}}}}_{\text{inputs}} .
\label{eq:normalized-collision-counts}
\end{equation}
Each factor is the probability mass function of a hypergeometric distribution, so $\tilde{C}_{\alpha \beta \gamma}$ is the probability that two independently drawn tuples collide in pattern $(\alpha, \beta, \gamma)$; in particular ${\sum_{\alpha, \beta, \gamma} \tilde{C}_{\alpha \beta \gamma} = 1}$, and the variance is the expected collision covariance, $\Var[S_\song] = \expect[V_{\alpha \beta \gamma}]$.


Next, we assume that the random variables $f(u)$ (over $f \sim \cF$) and $g(u)$ (over $g \sim \cG$) are sub-Gaussian.
This assumption is desired because it guarantees that their fourth moments are bounded, $\expect_f[f(u)^4] \le C \sigma_{\cF}^4$ and $\expect_g[g(u)^4] \le C \sigma_{\cG}^4$, uniformly in $u$ (overloading notation somewhat, and definition $\sigma_{\cF}^4, \sigma_{\cG}^4$ as the fourth moment of their respective distributions).
Note that this is a looser restriction than bounded firing rates, which are guaranteed in both biological and computational networks, so we do not restrict ourselves of consideration in any practical settings. 
We show that the variance decreases at a rate proportional to $1/M + 1/N$ with a multiplicative constant that is independent of the dimensionality of the inputs or smoothness of the tuning curves.

\subsection{Variance bound on HSIC}
\label{app:variance-upper-bound-on-HSIC}

First, we note that 
\begin{equation}
V_{\alpha \beta \gamma} \le V_{114} = \Var[\phi(\sigma)] ,
\label{eq:collision-variance-monotone}
\end{equation}
for all collision patterns indexed by $(\alpha, \beta, \gamma)$.
This is because $V_{\alpha \beta \gamma} = \Var_{\sigma \cap \sigma^\prime}\big[ \expect[ \phi(\sigma) \mid \sigma \cap \sigma^\prime ] \big]$ averages out non-shared arguments in the inner expectation, and such partial averaging can only decrease the variance.

The core idea is to bound $V_{114} = \Var[\phi(\sigma)] \le \expect[\phi(\sigma)^2] = \| \phi \|_2^2$, where $\| X \|_2 := \expect[X^2]^{1/2}$ denotes the $L^2$ norm of a scalar random variable.
Using the triangle inequality associated with this norm, on both the average over permutations in \cref{eq:hsic-u-statistic-kernel-defn} and the three terms with coefficients $(1, 1, -2)$, we have
\begin{equation}
\| \phi \|_2 \leq 4 \max_{a \neq b,\ c \neq d} \big\| f(u_a) f(u_b) \, g(u_c) g(u_d) \big\|_2 ,
\end{equation}
where the maximum is over the finitely many index patterns appearing in $\phi$ (the indices $a, b$ need not be distinct from $c, d$).
Unlike a maximum over permutations inside the norm, this is a maximum of fixed numbers, so no worst-case permutation needs to be selected.

Now fix any such pattern and condition on the inputs.
Due to independence of $f \sim \cF$ and $g \sim \cG$, we have:
\begin{equation}
\expect_{f, g} \big[ f(u_a)^2 f(u_b)^2 \, g(u_c)^2 g(u_d)^2 \big] = \expect_f[f(u_a)^2 f(u_b)^2 ] \cdot \expect_g[g(u_c)^2 g(u_d)^2]
\end{equation}
Each term is bounded by Cauchy--Schwarz, which we can then bound due to the sub-Gaussian assumption on the tuning curves:
\begin{equation}
\expect_f[f(u_a)^2 f(u_b)^2 ] \leq \sqrt{ \expect_f[ f(u_a)^4 ] \; \expect_f[ f(u_b)^4 ] } \le C \sigma_{\cF}^4 ,
\end{equation}
and likewise $\expect_g[ g(u_c)^2 g(u_d)^2 ] \le C \sigma_{\cG}^4$. As these bounds hold for every realization of the inputs, averaging over the inputs gives $\| f(u_a) f(u_b) \, g(u_c) g(u_d) \|_2^2 \le C^2 \sigma_{\cF}^4 \sigma_{\cG}^4$, so we conclude that
\begin{equation}
V_{114} = \Var[\phi(\sigma)] \le \expect[\phi(\sigma)^2] \le 16 \, C^2 \sigma_{\cF}^4 \, \sigma_{\cG}^4 = C' \, \sigma_{\cF}^4 \, \sigma_{\cG}^4
\label{eq:kernel-variance-bound}
\end{equation}
for a universal constant $C$ (absorbing the numerical factors).

Now, starting from \cref{eq:hsic-variance-normalized-Ctilde} and the fact that $V_{000}=0$, we have
\begin{align}
   \Var[S_\song] &= \sum_{(\alpha, \beta, \gamma) \neq 0} \widetilde{C}_{\alpha \beta \gamma} V_{\alpha \beta \gamma} \\
   &= \frac{1}{N_X}V_{100} + \frac{1}{N_Y}V_{010} + \frac{16}{M}V_{001} + \mathcal{O} \left ( \left( \frac{1}{N_X} + \frac{1}{N_Y} + \frac{1}{M} \right)^2 \right ) \\
   &\lesssim \left( \frac{1}{N_X} + \frac{1}{N_Y} + \frac{16}{M} \right) V_{114} = C \, \sigma_{\cF}^4 \, \sigma_{\cG}^4 \left(\frac{1}{N_X} + \frac{1}{N_Y} + \frac{16}{M} \right)
\end{align}

\section{Theoretical results on \texorpdfstring{$\CKD^2$}{CKD}}
\label{sec:appendix-b}

\subsection{Maximum Mean Discrepancy}
\label{app:MMD-CKD}
CKD is a special case of the maximum mean discrepancy (MMD; \citealp{Gretton2012}) with quadratic kernel $\kappa (f, g) = \langle \bar f, \bar g \rangle^2$, where $\bar{f}, \bar{g}$ represent the mean-subtracted (i.e. centered) tuning curves and the inner product is $L_2(P_\mathcal{U})$.
Given any kernel $\kappa$, its MMD is defined as 
\begin{equation*}
    \mathrm{MMD}^2(\mathcal F, \mathcal G) = \expect \biggl [ \kappa(f, f^\prime) \biggr ] + \expect \biggl [ \kappa(g, g^\prime) \biggr ] - 2\expect \biggl [ \kappa(f, g) \biggr ]
\end{equation*}
with tuning curves $f, f^\prime \sim \mathcal{F}, g, g^\prime \sim \mathcal{G}$ drawn independently from their respective distributions.
Hence, we get $\langle \bar f \otimes \bar f, \bar g \otimes \bar g \rangle_{HS} = \langle \bar f, \bar g \rangle^2 = \kappa(f, g)$ and $\mathrm{MMD}^2 = \mathrm{CKD}^2$.

\subsection{The CKD Estimator}
\label{sec:CKD-estimator}

The main difference with the CKD estimator is the two self-terms $S_\sanc$ involved. 
The self terms require a small modification: $\langle \bar k, \bar k \rangle$ is estimated by feeding $\phi$ \emph{two independent neurons from the same system},
\begin{equation}
\expect_{f_1, f_2 \sim \cF, \; u_{1:4}} \big[ \phi(f_1, f_2, u_{1:4}) \big] = \langle \bar k, \bar k \rangle ,
\end{equation}
which holds because $\phi$ is unbiased for $\langle \bar k, \bar h \rangle$ whenever its two neuron arguments are independent.

Collecting the three pieces, and symmetrizing the cross term over which neuron is drawn from each system, define the kernel
\begin{equation}
\psi(f_1, f_2, g_1, g_2, u_{1:4}) = \phi(f_1, f_2, u_{1:4}) + \phi(g_1, g_2, u_{1:4}) - \frac{1}{2} \sum_{a=1}^{2} \sum_{b=1}^{2} \phi(f_a, g_b, u_{1:4}) ,
\label{eq:d2-kernel-defn}
\end{equation}
which satisfies $\expect[\psi] = \langle \bar k, \bar k \rangle - 2 \langle \bar k, \bar h \rangle + \langle \bar h, \bar h \rangle = \CKD^2$ and is symmetric within each block of arguments (the two $\cF$-neurons, the two $\cG$-neurons, and the four inputs).
Given $N \ge 2$ neurons from each system and $M$ inputs, the corresponding U-statistic is
\begin{equation}
\widehat{\mathrm{CKD}}{}^2 = {{N}\choose{2}}^{-2} {{M}\choose{4}}^{-1} \sum_{n_1 < n_2} \sum_{n_1' < n_2'} \sum_{i < j < \ell < p} \psi(f_{n_1}, f_{n_2}, g_{n_1'}, g_{n_2'}, u_i, u_j, u_\ell, u_p) ,
\label{eq:d2-as-a-u-statistic}
\end{equation}
an unbiased estimator of $\CKD^2$.
It is a three-sample U-statistic of degree $(2, 2, 4)$: order two in each neuron sample---one higher than for HSIC, since each self term reuses its system---and order four in the inputs.

We can efficiently compute our estimator $\widehat{\mathrm{CKD}}{}^2$ in practice by leveraging the expression of $S_\song$ in \cref{eq:song-HSIC}, due to \citet{song2012feature}.
In particular, every term appearing on the right hand side of \cref{eq:song-HSIC} is separately linear in $\widehat{\mbK}$ and $\widehat{\mbH}$.
Thus, setting the two inputs to the estimator equal to $\widehat{\mbK} = N_X^{-1} \sum_n \mbx_n \mbx_n^\top$ and rearranging, we get:
\begin{equation}
S_\song(\mbX, \mbX) = \frac{1}{N_X^2} \; \sum_{n=1}^{N_X} \sum_{n'=1}^{N_X} \; S_\song(\mbx_n \mbx_n^\top, \mbx_{n'} \mbx_{n'}^\top) .
\end{equation}
We need to remove the diagonal terms where $n = n'$.
Subtracting off these terms and rescaling to achieve the proper normalization we get
\begin{equation}
S_\sanc(\mbX, \mbX) = 
\frac{1}{N_X (N_X - 1)} \left( N_X^2 S_\song(\widehat{\mbK}, \widehat{\mbK}) - \sum_{n} S_\song( \mbx_n \mbx_n^\top , \mbx_{n} \mbx_{n}^\top) \right)
\label{eq:hsic-self-neuron-diagonal-deleted}
\end{equation}
To efficiently calculate $S_\sanc$ between two rank one matrices, we do not form the $M \times M$ matrices explicitly.
Instead, one can verify that 
\begin{equation}
S_\sanc(\mbv \mbv^\top, \! \mbv \mbv^\top)
  \! = \! \frac{1}{M(M-3)} \! \left[ \left( z_2^2 - z_4 \right)
    + \frac{ \left( z_1^2 - z_2 \right)^2 }{(M-1)(M-2)}
    - \frac{2\left(z_1^2 z_2 + z_4 - 2 z_1 z_3 \right)}{M-2} \right],
\end{equation}
for any vector $\mbv \in \reals^M$ and $z_1 = \sum_i v_i$, $z_2 = \sum_i v_i^2$, $z_3 = \sum_i v_i^3$, and $z_4 = \sum_i v_i^4$.
Thus, computing $S_\sanc$ on two identical rank-one matrices requires only $O(M)$ calculations.
Repeating this calculation across all neurons in \cref{eq:hsic-self-neuron-diagonal-deleted} requires only $O(M N_X)$ computation.

\subsection{The variance as a sum over collisions}

The variance decomposes exactly as in \crefrange{eq:hsic-variance-giant-sum}{eq:normalized-collision-counts}, now for the degree-$(2, 2, 4)$ kernel $\psi$.
The only change is that the two neuron samples now contribute degree two, so the collision indices run over
\begin{equation}
\alpha, \beta \in \{ 0, 1, 2 \}, \qquad \gamma \in \{ 0, 1, 2, 3, 4 \} ,
\end{equation}
giving a $3 \times 3 \times 5$ table of patterns.
Writing $V_{\alpha \beta \gamma}$ for the projection variance of each pattern (with $V_{000} = 0$ as before),
\begin{equation}
\Var[\widehat{\mathrm{CKD}}{}^2] = \sum_{\alpha, \beta, \gamma} \tilde{C}_{\alpha \beta \gamma} V_{\alpha \beta \gamma} ,
\end{equation}
where $\tilde{C}_{\alpha \beta \gamma}$ is again a product of three hypergeometric probabilities, now with degree $d = 2$ for each neuron sample,
\begin{equation}
\tilde{C}_{\alpha \beta \gamma} = \frac{{{2}\choose{\alpha}} {{N_X-2}\choose{2-\alpha}}}{{{N_X}\choose{2}}} \cdot \frac{{{2}\choose{\beta}} {{N_Y-2}\choose{2-\beta}}}{{{N_Y}\choose{2}}} \cdot \frac{{{4}\choose{\gamma}} {{M-4}\choose{4-\gamma}}}{{{M}\choose{4}}} .
\end{equation}
The leading single-collision probabilities again follow the hypergeometric mean overlap $d^2 / P$ of two degree-$d$ draws from a population of size $P$: a single neuron collision occurs with probability $\tilde{C}_{100} \to 4/N_X, \tilde{C}_{010} \to 4/N_Y$ (degree $d = 2$), and a single input collision with $\tilde{C}_{001} \to 16/M$ (degree $d = 4$). The neuron rate is thus raised from the $1/N$ of the preceding section to $4/N$ (for both $N_X$ and $N_Y$.

\subsection{Proof of \cref{th:bound-CKD}}
\label{app:proof-of-bound-CKD}

For each neuron, write $\bar f = \expect_u[f(u)]$ and $\bar g = \expect_u[g(u)]$ for its stimulus-averaged response.
Then consider the expansion of $\expect_{u, u'}[(f(u) - \bar f)(f(u') - \bar f)\, h(u, u')]$, given by
\begin{align*}
\expect_{u, u'}[f(u) f(u') h(u, u')] 
+ \bar f^2\, \expect_{u, u'}[h(u, u')] 
- 2 \bar f\, \expect_{u, u'}[f(u) h(u, u')] ,
\end{align*}
where the cross terms combine by the symmetry of $h$.
Substituting $h(u, u') = \E_g [g(u) g(u')]$ and $\bar f = \expect_u[f(u)]$ into the right hand side and pulling out the expectation, we get:
\begin{equation*}
\expect_{g,\, u_{1:4}}\big[ f(u_1) f(u_2) g(u_1) g(u_2) + f(u_1) f(u_2) g(u_3) g(u_4) - 2 f(u_1) f(u_2) g(u_1) g(u_3) \big] = \expect[\phi(\sigma) \mid f] ,
\end{equation*}
where $\sigma = (f_n, g_{n'}, u_i, u_j, u_\ell, u_p)$ denotes a set of inputs to $\phi$, as in Appendix~\ref{sec:appendix-U-stats-HSIC}.
The equality holds because the inputs are i.i.d., so the permutation average in \cref{eq:hsic-u-statistic-kernel-defn} does not change the expectation.

Since $f(u) - \bar f$ has zero mean over $u$, the centering terms of $\bar h$ integrate against it to zero, so $h$ may be replaced by $\bar h$:
\begin{equation}
\expect[\phi(\sigma) \mid f] = \expect_{u, u'}\big[ (f(u) - \bar f)(f(u') - \bar f) \, \bar{h}(u, u') \big] = \langle \bar{k}_f, \bar{h} \rangle ,
\label{eq:projection-onto-f}
\end{equation}
where $\bar{k}_f(u, u') = (f(u) - \bar f)(f(u') - \bar f)$ is the rank-one centered kernel of the single neuron $f$.
The same holds when the other neuron is drawn from any distribution, with $\bar h$ replaced by that distribution's centered kernel.

Turning to the CKD estimator, we use $\Sigma = (f_{n_1}, f_{n_2}, g_{n_1'}, g_{n_2'}, u_i, u_j, u_\ell, u_p)$ to denote a set of inputs to the U-statistic kernel $\psi$, defined in \cref{eq:d2-kernel-defn}.
We write $\Sigma$ and $\Sigma^\prime$ as a given pair of these inputs, analogous to how $\sigma$ and $\sigma^\prime$ denote a pair of inputs to HSIC U-statistic kernel $\phi$.

Consider $V_{100}$, the covariance of $\psi(\Sigma)$ and $\psi(\Sigma^\prime)$ when they share exactly one $\cF$-neuron $f$; by \cref{eq:Cov-Var-U-stats}, this equals $\Var_f\big[\expect[\psi(\Sigma) \mid f]\big]$.
We condition $\psi$ on $f$ and average over the remaining neurons and inputs, applying \cref{eq:projection-onto-f} to each of the three terms of \cref{eq:d2-kernel-defn}.
The self-$\cF$ term projects onto $\bar k$ (the population kernel of the second $\cF$-neuron), the self-$\cG$ term does not involve $f$ at all, and among the four cross terms only the two that contain $f$ retain it:
\begin{equation}
\begin{aligned}
\expect[ \psi(\Sigma) \mid f ]
&= \underbrace{\langle \bar{k}_f, \bar k \rangle}_{\phi(f, f_2)}
\;+\; \underbrace{\langle \bar h, \bar h \rangle}_{\phi(g_1, g_2)}
\;-\; \frac{1}{2} \Big( \underbrace{2 \langle \bar{k}_f, \bar h \rangle}_{f_a = f} + \underbrace{2 \langle \bar k, \bar h \rangle}_{f_a = f_2} \Big) \\
&= \langle \bar{k}_f, \, \bar k - \bar h \rangle + \langle \bar h, \bar h \rangle - \langle \bar k, \bar h \rangle \\
&= \langle \bar{k}_f, \bar k - \bar h \rangle + \textrm{const} .
\end{aligned}
\label{eq:d2-projection-onto-f}
\end{equation}
The terms constant in $f$ do not affect the variance, leaving only the gap-dependent part.
This projection is \emph{linear} in $\bar k - \bar h$, so by Cauchy--Schwarz
\begin{equation}
V_{100} = \Var_f \big[ \langle \bar{k}_f, \bar k - \bar h \rangle \big] \le \expect_f \big[ \| \bar{k}_f \|^2 \big] \, \| \bar k - \bar h \|^2 = \expect_f \big[ \| \bar{k}_f \|^2 \big] \, \CKD^2 ,
\label{eq:d2-V100}
\end{equation}
and likewise for $V_{010}$ with the roles of the $\cF$- and $\cG$-neurons exchanged.
For a shared input, the three terms of $\psi$ instead complete the square:
\begin{equation}
\expect[ \psi(\Sigma) \mid u ] = \frac{1}{2} \expect_{u'} \big[ ( \bar k - \bar h )(u, u')^2 \big] + \textrm{const} .
\label{eq:d2-projection-onto-u}
\end{equation}
This is quadratic in $\bar k - \bar h$, so that $V_{001} = \tfrac{1}{4} \Var_u\big[ \expect_{u'} (\bar k - \bar h)(u, u')^2 \big]$.

We now bound each contribution explicitly in terms of the \emph{maximal fourth moment} $q$ of a neuron's deviation from its own stimulus-averaged rate,
\begin{equation}
q = \max \Big\{ \sup_u \expect_f\big[(f(u) - \bar f)^4\big], \ \sup_u \expect_g\big[(g(u) - \bar g)^4\big] \Big\},
\label{eq:fourth-moment-defn}
\end{equation}
where $f \sim \cF$ and $g \sim \cG$.
Note that these this is a moment about each neuron's stimulus-averaged rate $\bar f$, not central moments across neurons.
In the main text, we list several sufficient conditions under which $q$ is finite. For example, bounded tuning functions or Lipschitz tuning functions and compactly-supported $P$ suffice. 

By \cref{eq:d2-V100}, followed by Jensen's inequality and the definition of $q$, we have
\begin{equation*}
V_{100} \le \expect_f\Big[\big(\expect_u (f(u) - \bar f)^2\big)^2\Big]\,\CKD^2
\le \expect_u \expect_f\big[(f(u) - \bar f)^4\big]\,\CKD^2
\le q \, \CKD^2 ,
\end{equation*}
and likewise $V_{010} \le q\, \CKD^2$.

We next turn to the input collision term, $V_{001}$.
Since $\bar k(u, u') = \expect_f[(f(u) - \bar f)(f(u') - \bar f)]$, Cauchy--Schwarz combined with the elementary inequality $(a - b)^2 \leq 2a^2 + 2 b^2$ gives
\begin{equation}
(\bar k - \bar h)(u,u')^2 \le 2\bar k(u,u')^2 + 2\bar h(u,u')^2 \leq 4q
\end{equation}
Therefore,
\begin{align*}
V_{001} &\le \tfrac14\, \expect_u\Big[\big(\expect_{u'} (\bar k - \bar h)(u,u')^2\big)^2\Big] \\
& \le \tfrac14 \Big(\sup_u \expect_{u'} (\bar k - \bar h)(u,u')^2\Big)\, \expect_{u,u'}\big[(\bar k - \bar h)(u,u')^2\big] \\
&\le q\, \CKD^2.
\end{align*}

It remains to control the higher-order collision terms. 
As before, we use a crude bound.
Recall that $\Var[\psi] = V_{224}$, the covariance of $\psi(\Sigma)$ and $\psi(\Sigma^\prime)$ when $\Sigma = \Sigma^\prime$, is the maximal $V_{\alpha, \beta, \gamma}$ term since conditioning on fewer shared arguments can only reduce the variance of the projection.

Here, we use a trick to evaluate the individual monomials of $\phi$ using the stimulus-centered responses.
To accomplish this, we provide the following lemma.

\begin{lemma}[shift invariance of $\phi$ and $\psi$]
\label{lem:ckd-phi-invar-const}
$\phi(f + c, g) = \phi(f, g)$, and by the symmetry of $\phi$ the same holds for a shift of $g$.
By definition of $\psi$, this holds for $\psi$ as well.
\end{lemma}

\begin{proof}
To see the cancellation explicitly, replace $f$ by $f + c$ for a constant $c$. 
In each of the three monomials of \cref{eq:hsic-u-statistic-kernel-defn}, the $f$-pair expands as
\begin{equation*}
  (f(u_i) + c)(f(u_j) + c) = f(u_i) f(u_j) + c f(u_i) + c f(u_j) + c^2 .
\end{equation*}
The first piece reproduces $\phi(f, g)$ that we know. 
We will show that the other three pieces vanish under the permutation average.


\emph{The $c^2$ terms.} They are $c^2 \big( g(u_i) g(u_j) + g(u_\ell) g(u_p) - 2 g(u_i) g(u_\ell) \big)$. Each of the three products is $g$ evaluated at two distinct inputs. Cycling through all the possible pairs makes this argument vanish.

\emph{The terms linear in $c$.} Each such product evaluates $f$ at one input and $g$ at two distinct inputs, and has one of two shapes. Either $g$ takes two inputs different than $f$ (e.g. $g(u_p), g(u_q)$) or $g$ shares one input with $f$.
  Write the first pattern as $A$, write the second pattern as $B$. 
  The terms containing $c f(u_i)$ are $c \big( f(u_i) g(u_i) g(u_j) + f(u_i) g(u_\ell) g(u_p) - 2 f(u_i) g(u_i) g(u_\ell) \big)$. 
  This contributes a factor of $c\,(A + B - 2A) = c\,(B - A)$.
  The terms containing $c f(u_j)$ are $c \big( f(u_j) g(u_i) g(u_j) + f(u_j) g(u_\ell) g(u_p) - 2 f(u_j) g(u_i) g(u_\ell) \big)$. 
  This contributes a factor of $c\,(A + B - 2B) = c\,(A - B)$.
  Together, the two contributions cancel each other out, and the whole term vanishes.
\end{proof}

We may therefore evaluate $\psi$ on the stimulus-centered responses $f - \bar f$ and $g - \bar g$.
By abuse of notation, we denote these \emph{independent and centered} neurons $f, g$ as well.
Thus, every summand of $\phi$ in \cref{eq:hsic-u-statistic-kernel-defn} has the form $f(u_i)f(u_j)\,g(u_\ell)g(u_p)$ for two \emph{independent and centered} neurons $f, g$. 
This allows us to use the bound with respect to $q$.

Conditioning on the inputs, independence of $f$ and $g$ followed by Cauchy--Schwarz applied to each neuron separately gives
\begin{equation}
\expect\big[f(u_i)^2 f(u_j)^2\, g(u_\ell)^2 g(u_p)^2\big]
= \expect\big[f(u_i)^2 f(u_j)^2\big]\,\expect\big[g(u_\ell)^2 g(u_p)^2\big]
\le q \cdot q = q^2 .
\label{eq:q-bound-1}
\end{equation}

The triangle inequality over the permutation average and the coefficients $(1, 1, -2)$ then gives $\|\phi(a, b, \cdot)\|_2 \le 4 q$.
Hence, replacing the terms in \cref{eq:d2-kernel-defn},
\begin{equation*}
\|\psi\|_2 \le 4 q + 4 q + \tfrac12 \cdot 4 \cdot 4 q = 16\, q ,
\label{eq:q-bound-2}
\end{equation*}
and thus $V_{\alpha\beta\gamma} \le \Var[\psi] \le 256\, q^2$.

We complete the bound by applying a counting argument on the higher order terms (i.e. $\Sigma, \Sigma^\prime$ pairs with at least \textit{two} overlapping terms).
For ease of notation, let $\epsilon := \tfrac1M + \tfrac1{N_X} + \tfrac1{N_Y}$.
Next, let the number of overlapping terms be $T \geq 2$. 
Single-collision patterns occur with probability $\tilde C_{100} \le 4/N_X$, $\tilde C_{010} \le 4/N_Y$, and $\tilde C_{001} \le 16/M$.

Next, note that a hypergeometric overlap of two size-$\delta$ draws from $P$ items has $\expect[X(X-1)] = \delta^2(\delta-1)^2 / (P(P-1)) \le 2 (\expect X)^2$.
In our case and $\mathbf{1}\{T \ge 2\} \le T(T-1)/2$, this gives
\begin{equation*}
\Pr(T \ge 2) \le \Big( \tfrac{4}{N_X} + \tfrac{4}{N_Y} + \tfrac{16}{M} \Big)^2 \le 256\, \epsilon^2 .
\end{equation*}

Combining these bounds,
\begin{align}
\Var[\widehat{\CKD}{}^2]
&\le \tfrac{4}{N_X} V_{100} + \tfrac{4}{N_Y} V_{010} + \tfrac{16}{M} V_{001} + \Pr(T \ge 2)\, \Var[\psi] \nonumber \\
&\le q\, \CKD^2 \left(\tfrac{4}{N_X} + \tfrac{4}{N_Y} + \tfrac{16}{M} \right) + 256\,\epsilon^2 \cdot 256\, q^2 \nonumber \\
&\le 16\, q\, \CKD^2\, \epsilon \;+\; 2^{16}\, q^2\, \epsilon^2 .
\label{eq:d2-explicit-bound}
\end{align}
Each term is homogeneous of degree eight in the responses, since $\CKD^2$ scales as $q$.
This implies \cref{eq:bound-CKD} with $C = 2^{16} \max(q, q^2)$.
The numerical constant $2^{16}$ has not been optimized, and we expect the sharp constant to be considerably smaller. 

%

\section{Independent Trial-to-trial noise}
\label{app:noise-theory}

In this section, we analyze how the U-statistic estimator of $\mathrm{CKD}^2$ behaves when neural responses are corrupted by trial-to-trial noise.
Here, we assume that noise is independent across neurons and trials.
In appendix~\ref{app:corr-noise}, we relax these assumptions to allow pairs of neurons to exhibit ``noise correlations'' from trial-to-trial.

In this appendix section, we show the following.
First, we show that the estimator remains unbiased (\Cref{app:noise-unbiased}). 
Second, we show that the estimator's variance is insensitive to first-order: the leading rate remains $1/M + 1/N_X + 1/N_Y$ and the constants $V_{001}$, $V_{010}$, and $V_{100}$ that scale each term remain the same (\Cref{app:noise-insensitive}).

\subsection{Sampling model with noise}
\label{app:noise-model}

We denote noise variables by $\xi$ (system $X$) and $\zeta$ (system $Y$). 
For a tuning curve $f$ we write $\bar f := \mathbb{E}_{u}[f(u)]$.
Neurons and inputs are sampled as in Section~2.1: $f_1,\dots,f_{N_X}\sim\mathcal F$, $g_1,\dots,g_{N_Y}\sim\mathcal G$, and $u_1,\dots,u_M\sim P$, all independent. We collect these noiseless sampling variables into
\begin{equation*}
  \mathcal D := (f_{1:N_X},\, g_{1:N_Y},\, u_{1:M}).
\end{equation*}
Each input is presented $L \ge 1$ times. On repetition $\ell$ we observe
\begin{equation}
  X_{ni\ell} = f_n(u_i) + \xi_{ni\ell},
  \qquad
  Y_{ni\ell} = g_n(u_i) + \zeta_{ni\ell}.
\end{equation}
We make the following assumptions.
\begin{enumerate}
  \item[(N1)] \emph{Conditional mean zero noise.} $\mathbb{E}[\xi_{ni\ell}\mid\mathcal D] = 0$ and $\mathbb{E}[\zeta_{ni\ell}\mid\mathcal D]=0$.
  \item[(N2)] \emph{Noise has bounded fourth moments.} We assume $\E [ \xi_{ni\ell}^4 \mid\mathcal D ]$ and $\E [ \zeta_{ni\ell}^4 \mid\mathcal D ]$ are uniformly upper bounded by a finite constant for all neurons and possible network inputs. 
  \item[(N3)] \emph{Conditional independence.} Conditional on $\mathcal D$, all noise variables $\{\xi_{ni\ell}\}$ and $\{\zeta_{ni\ell}\}$ are mutually independent across neurons, inputs, repetitions, and systems. Moreover, the conditional law of $\xi_{ni\ell}$ depends on $\mathcal D$ only through $(f_n, u_i)$, and likewise for $\zeta_{ni\ell}$ through $(g_n,u_i)$.
\end{enumerate}
Heteroscedastic noise is allowed, so long as the fourth moments are uniformly bounded.
The estimator is computed from trial-averaged responses
\begin{equation}
  \bar X_{ni} := \frac1L\sum_{\ell=1}^L X_{ni\ell} = f_n(u_i) + \bar\xi_{ni},
  \qquad
  \bar Y_{ni} := \frac1L\sum_{\ell=1}^L Y_{ni\ell} = g_n(u_i) + \bar\zeta_{ni}.
\end{equation}
The averaged noise terms $\bar\xi_{ni}$ and $\bar\zeta_{ni}$ satisfy (N1)--(N2) with the variance reduced by a multiplicative factor of $1/L$.

\subsection{The U-statistic estimator remains unbiased under independent noise}
\label{app:noise-unbiased}
 
Recall from \Cref{sec:appendix-U-stats-HSIC} and \Cref{sec:appendix-b} that,
\begin{equation}
  \phi(f,g;u_{1:4}) = \frac{1}{4!}\sum_{\pi\in S_4}
  \left[
  \begin{aligned}
    &f(u_{\pi_1}) f (u_{\pi_2}) g(u_{\pi_1} g(u_{\pi_2}) \\
    &+ f(u_{\pi_1}) f (u_{\pi_2}) g(u_{\pi_3}) g(u_{\pi_4}) \\
    &- 2 f(u_{\pi_1}) f(u_{\pi_2}) g(u_{\pi_1}) g(u_{\pi_3}) \\
    \end{aligned}
  \right],
  \label{eq:noise-phi}
\end{equation}
\begin{equation}
  \psi(f_1,f_2,g_1,g_2;u_{1:4})
  = \phi(f_1,f_2;u_{1:4}) + \phi(g_1,g_2;u_{1:4})
  - \frac12\sum_{a=1}^2\sum_{b=1}^2 \phi(f_a,g_b;u_{1:4}),
  \label{eq:noise-psi}
\end{equation}
and $\widehat{\CKD}{}^2$ averages $\psi$ over all index tuples
\begin{equation*}
  \Sigma = (n_1<n_2;\; n_1'<n_2';\; i_1<i_2<i_3<i_4).
\end{equation*}
With noisy data, $\psi$ is evaluated on observed entries. That is, each $f_{n}(u_i)$ is replaced by $\bar X_{ni}$ and each $g_{n}(u_i)$ by $\bar Y_{ni}$.
In this appendix section, we write $\widehat{\CKD}{}^2_{\mathrm{avg}}$ to denote this estimator.
As before, we use $\widehat{\CKD}{}^2$ to denote the ``noise-free'' estimator that we previously analyzed in \Cref{sec:appendix-b} and which is our primary focus in the main text.
In the context of this appendix section, $\widehat{\CKD}{}^2$ can be viewed as idealized estimator that has direct access to the neural tuning functions.

\begin{proposition}[Unbiasedness under decorrelated noise]
\label{prop:noise-unbiased}
Assume (N1) and (N3) and the moment conditions of \Cref{th:bound-CKD}. Then for every $L\ge1$,
\begin{equation}
  \mathbb{E}\big[\widehat{\CKD}{}^2_{\mathrm{avg}}\,\big|\,\mathcal D\big] = \widehat{\CKD}{}^2,
  \qquad\text{and hence}\qquad
  \mathbb{E}\big[ \widehat{\CKD}{}^2_{\mathrm{avg}} \big] = \CKD^2.
\end{equation}
\end{proposition}
 
\begin{proof}
Applying iterated expectations, $\E \big[\widehat{\CKD}{}^2_{\mathrm{avg}} \big] = \E_{\cD} \left[ \E_{\xi, \zeta} \left[ \widehat{\CKD}{}^2_{\mathrm{avg}} \mid \mathcal D \right]\right]$, with the inner expectation averaging over the noise terms.

Conditional on $\mathcal D$, the averaged noise terms are independent and mean zero by (N1)--(N2). 
Expanding each response into signal plus noise, every term that contains a noise variable contains it only once, and so has conditional mean zero. 
Hence, every term averages like:
\begin{equation*}
  \mathbb{E}\left[
  \begin{aligned}
    &\bar X_{n_1 i_1} \bar X_{n_1 i_2}\bar Y_{n_2 i_1}\bar Y_{n_2 i_2} \\
    &+ \bar X_{n_1 i_1} \bar X_{n_1 i_2}\bar Y_{n_2 i_3}\bar Y_{n_2 i_4} \\
    &- 2\, \bar X_{n_1 i_1} \bar X_{n_1 i_2}\bar Y_{n_2 i_1}\bar Y_{n_2 i_3}
  \end{aligned}
  \;\middle|\; \mathcal D\right]
  =
  \begin{aligned}
    &f_{n_1}(u_{i_1})f_{n_1}(u_{i_2})g_{n_2}(u_1)g_{n_2}(u_2) \\
    &+ f_{n_1}(u_1)f_{n_1}(u_2)g_{n_2}(u_3)g_{n_2}(u_4) \\
    &- 2\, f_{n_1}(u_1)f_{n_1}(u_2)g_{n_2}(u_1)g_{n_2}(u_3),
  \end{aligned}
\end{equation*}
including for the self-terms since the noise is assumed independent across neurons and the self-term estimator only touches distinct neuron pairs.
Thus, the inner expectation evaluates to $\E_{\xi, \zeta} \left[ \widehat{\CKD}{}^2_{\mathrm{avg}} \mid \mathcal D \right] = \widehat{\CKD}^2$ and the estimator is remains unbiased as claimed.
\end{proof}

\subsection{The variance of the U-statistic estimator is insensitive to noise to first-order}
\label{app:noise-insensitive}

Recall from Appendix~\ref{sec:appendix-b} that two index tuples $\Sigma$ and $\Sigma^\prime$ have collision pattern $(\alpha,\beta,\gamma)$ when they share $\alpha$ $\cF$-neurons, $\beta$ $\cG$-neurons, and $\gamma$ inputs. 
After incorporating noise, we need to additionally account for collisions over neuron-input pairs.
For example, suppose $\Sigma$ and $\Sigma^\prime$ both contain the $\cF$-neuron $f_1$ and both contain
the input $u_2$. Then the noisy response $\bar X_{12}$ enters both $\psi(\Sigma)$ and
$\psi(\Sigma')$, and its noise is a source of covariance between them. If instead the two
tuples share $f_1$ but no input, or share $u_2$ but no $\cF$-neuron, then the two
kernels are evaluated on disjoint sets of noisy responses, so the noise contributes nothing.

Accordingly, we now read $\Sigma$ as the full collection of random variables that $\psi(\Sigma)$ depends on: its neurons, its inputs, and the averaged noise of its sixteen neuron-input pairs. 
Under this reading, $\Sigma \cap \Sigma^\prime$ consists of the shared neurons, the shared inputs, and the noise of the shared neuron-input pairs, and our prior bound holds verbatim. 
To distinguish the two, we write 
\begin{equation}
  V^{\mathrm{avg}}_{\alpha\beta\gamma}
  := \mathrm{Var}\Big(\mathbb{E}\big[\psi(\Sigma)\,\big|\,\Sigma \cap \Sigma^\prime\big]\Big),
  \label{eq:noise-collision-var}
\end{equation}
to denote the variance terms for $\widehat{\CKD}{}^2_{\mathrm{avg}}$.

Collision probabilities depend only on the index tuples, not on the noise.
As before,
\begin{equation}
  \mathrm{Var}\big[\widehat{\CKD}{}^2_{\mathrm{avg}} \big] = \sum_{\alpha,\beta,\gamma}\tilde C_{\alpha\beta\gamma}\,V^{\mathrm{avg}}_{\alpha\beta\gamma},
  \label{eq:noise-var-sum}
\end{equation}
with the same hypergeometric weights $\tilde C_{\alpha\beta\gamma}$.

In Appendix~\ref{sec:appendix-U-stats-HSIC} and Appendix~\ref{sec:appendix-b}, we previously saw that the constants $V_{001}$, $V_{010}$, and $V_{100}$ describe the estimator's variance to first-order, and that these terms asymptotically dominate in the large-sample limit.
Here we show that these determinitive constants are insensitive to the trial-to-trial noise.

\begin{proposition}[Noise does not affect the first-order constants]
\label{prop:noise-first-order}
Assume (N1)--(N3). If a collision pattern $(\alpha,\beta,\gamma)$ has no shared
neuron-input pairs, that is if $\alpha + \beta = 0$ or $\gamma = 0$, then
\begin{equation*}
  V^{\mathrm{avg}}_{\alpha\beta\gamma} = V_{\alpha\beta\gamma}.
\end{equation*}
In particular, this holds for the three first-order patterns,
\begin{equation*}
  V^{\mathrm{avg}}_{100} = V_{100},
  \qquad
  V^{\mathrm{avg}}_{010} = V_{010},
  \qquad
  V^{\mathrm{avg}}_{001} = V_{001},
\end{equation*}
as well as for $V^{\mathrm{avg}}_{000} = V_{000} = 0$. For every other pattern,
$V^{\mathrm{avg}}_{\alpha\beta\gamma} \ge V_{\alpha\beta\gamma}$.
\end{proposition}

\begin{proof}
Write $A$ for the index part of the overlap, namely the shared neurons and shared
inputs, so that $\Sigma \cap \Sigma^\prime$ consists of $A$ together with the noise of the shared
neuron-input pairs. By the law of total variance,
\begin{equation}
  V^{\mathrm{avg}}_{\alpha\beta\gamma}
  = \mathrm{Var}\Big(\mathbb{E}\big[\psi(\Sigma) \mid A\big]\Big)
  + \mathbb{E}\Big[\mathrm{Var}\Big(\mathbb{E}\big[\psi(\Sigma) \mid \Sigma \cap \Sigma^\prime\big] \;\Big|\; A\Big)\Big].
  \label{eq:noise-total-var}
\end{equation}
Consider the first term. Since $A$ contains no noise variables, the expectation conditioned on $A$ averages over all of the noise.
Hence, by Proposition~\ref{prop:noise-unbiased}, $\mathbb{E}[\psi(\Sigma) \mid A]$
returns the kernel evaluated on noiseless responses, and the variance of this quantity is precisely what we defined as $V_{\alpha\beta\gamma}$.
The second term of eq.~\eqref{eq:noise-total-var} is nonnegative, which gives the final claim.

Now suppose the pattern has no shared neuron-input pairs. Then $\Sigma \cap \Sigma^\prime = A$, so the inner conditional expectation in the second term is a function of $A$ alone. Conditioning on $A$ leaves it constant, so its conditional
variance is zero and the term vanishes.
A neuron-input pair is shared only if its neuron and its input are
both shared, so the number of shared pairs is $\alpha\gamma$ in system $X$ and
$\beta\gamma$ in system $Y$. Both vanish  when $\alpha + \beta = 0$ or
$\gamma = 0$, which covers $(1,0,0)$, $(0,1,0)$, $(0,0,1)$, and $(0,0,0)$.
\end{proof}

\subsection{The variance bound carries over under noise}
\label{app:noise-variance-upper-bound-note}

Under (N1)--(N3), the proof of \Cref{th:bound-CKD} carries over with minimal modifications.
\Cref{prop:noise-first-order} above already established this and showed that the first-order terms in the expansion  $\Var[\widehat{\CKD}{}^2]$ are unaffected by decorrelated noise.
Thus, the only step that changes is the bound on $\Var[\psi]$ that controls the higher-order collisions.
That argument uses only the offset invariance of $\phi$ (\cref{lem:ckd-phi-invar-const}), the conditional independence of the two neurons given the inputs (which (N3) preserves), and a bound on the fourth moments of the responses.
Applying \cref{lem:ckd-phi-invar-const} with the noiseless offsets $\bar f$ and $\bar g$, the relevant responses are $f(u) - \bar f + \bar\xi$ and $g(u) - \bar g + \bar\zeta$, where $\bar\xi$ and $\bar\zeta$ denote the trial-averaged noise of a neuron at input $u$. We therefore replace $q$ with
\begin{equation}
\tilde q = \max \Big\{ \sup_u \E \big[ (f(u) - \bar f + \bar\xi)^4 \,\big|\, u \big], \ \sup_u \E \big[ (g(u) - \bar g + \bar\zeta)^4 \,\big|\, u \big] \Big\}
\label{eq:fourth-moment-defn-noise}
\end{equation}
where the expectations are jointly taken over a randomly sampled neuron and its trial-averaged noise.
Assumption (N2) and the elementary inequality $(a + b)^4 \le 8a^4 + 8b^4$ are enough to establish that $\tilde q$ is finite.
Intuitively, $\tilde q$ is controlled by the sum of the fourth moments of the centered tuning curve, $f(u) - \bar f$, and of the trial-averaged noise, $\bar\xi$, with the latter shrinking as the number of repetitions $L$ grows.

\section{Correlated trial-to-trial noise}
\label{app:corr-noise}

Here, we further extend the analysis in \Cref{app:noise-theory} to cover cases where simultaneously measured neurons share fluctuations  or ``noise correlations'' from trial-to-trial.
In this section, we  only assume that distinct repeated stimulus presentations have variability that is independent.
We show the following.
\begin{enumerate}
  \item The estimator $\widehat{\CKD}{}^2_{\mathrm{avg}}$ that we introduced in \Cref{app:noise-theory}, produces a biased result when noise is correlated across neurons. For $L$ trials per stimuli, this bias decays as $1/L$ in the number of repetitions $L$. Notably, this bias does not decay with the number of neurons or stimuli (\Cref{app:corr-noise-bias}).
  \item We construct an alternative estimator, $\widehat{\CKD}{}^2_{\mathrm{split}}$, that requires only two trials per stimuli for full debiasing of correlated noise (\Cref{app:corr-noise-split}).
  \item The variance of this new estimator still scales at the $1/M + 1/N_X + 1/N_Y$ rate, and the only constant that changes is the coefficient of the $1/M$ term. (\Cref{app:corr-noise-variance}).
\end{enumerate}

\subsection{Sampling model with correlated noise}
\label{app:corr-noise-model}

We consider the same data generating process described in \cref{app:noise-model}, but modify assumption (N3) to the following:
\begin{enumerate}
  \item[(C3)] \emph{Independent trials, arbitrary correlations within a trial.} Conditional on $\cD$, the vectors $\{\nu_{i\ell}\}_{i \le M,\, \ell \le L}$ are independent. Within a trial, $\nu_{i\ell}$ may have any dependence structure, across neurons of the same system and across the two systems.
\end{enumerate}
Note that assumption (N3) of \Cref{app:noise-theory} is the special case of (C3) in which the coordinates of $\nu_{i\ell}$ are themselves independent.
Our relaxation in (C3) allows, for instance, shared gain fluctuations, shared additive fluctuations (e.g.\ arousal or brain state), and noise correlations whose structure depends on the neurons' tuning.
The relaxation also accounts for cases where neuroal response noise is correlated \textit{across} systems (e.g. when two brain areas are simultaneously recorded in the same animal) and for input-dependent, heteroskedastic noise correlations.
We exclude cases where noise correlations are dependent on the noise structure of previous trials as well as cases where neural tuning drifts between trials.


\subsection{The standard estimator is biased under correlated noise}
\label{app:corr-noise-bias}

In Appendix~\ref{app:noise-theory}, unbiasedness followed from the fact that every noise variable enters each monomial of $\phi$ at most once. With correlated noise this argument fails: when two neural responses are measured on the same trial, the product of their noise terms has nonzero mean. The following lemma quantifies the resulting bias in $\phi$: 


\begin{lemma}
\label{lem:same-trial-cov}
Fix two distinct neurons with tuning curves $f$ and $g$. Let $f(u_i) + \xi_i$ and $g(u_i) + \zeta_i$ be their responses on the same trial of input $u_i$. Then
\begin{equation*}
  \E\big[\phi(f + \xi, g + \zeta; u_{1:4}) \,\big|\, f, g\big] = \Big(\mathrm{Cov}_u\big(f(u), g(u)\big) + \E\big[\xi_i \zeta_i \,\big|\, f, g\big]\Big)^2 .
\end{equation*}
This holds regardless of whether neurons are recorded from the same system (e.g., $f, g \sim \cF$) or from different systems ($f \sim \cF$, $g \sim \cG$).
\end{lemma}

\begin{proof}
Conditioning on $f$ and $g$ leaves $u_{1:4}$ i.i.d.\ from $P$. Recall that every monomial of $\phi$ has the form $f(u_i)\, f(u_j)\, g(u_k)\, g(u_l)$ with $i \neq j$ and $k \neq l$. Responses at distinct inputs lie on distinct trials, so by (C3) their noise terms are conditionally independent given $\mathcal D$, and by (N1) each has conditional mean zero. Expanding each response into signal plus noise, every term containing a noise factor that is not paired with a noise factor at the same input therefore vanishes. The only surviving noise terms are $\xi_i \zeta_i$, arising when $f$ and $g$ are evaluated at the same input $u_i$. Hence each such pair contributes $f(u_i) g(u_i) + \xi_i \zeta_i$. Since pairs at distinct inputs are independent and identically distributed given $f$ and $g$, averaging over the inputs and the noise, the three monomial types of $\phi$ give
\begin{align*}
  f(u_i) f(u_j) g(u_i) g(u_j) &\;\to\; \Big(\E_u[f(u) g(u)] + \E\big[\xi_i \zeta_i \,\big|\, f, g\big]\Big)^2, \\
  f(u_i) f(u_j) g(u_l) g(u_p) &\;\to\; \bar f^{\,2}\, \bar g^{\,2}, \\
  f(u_i) f(u_j) g(u_i) g(u_l) &\;\to\; \Big(\E_u[f(u) g(u)] + \E\big[\xi_i \zeta_i \,\big|\, f, g\big]\Big)\, \bar f\, \bar g .
\end{align*}
Combining these with coefficients $(1, 1, -2)$ gives
\begin{equation*}
  \Big(\E_u[f(u) g(u)] + \E\big[\xi_i \zeta_i \,\big|\, f, g\big] - \bar f\, \bar g\Big)^2 = \Big(\mathrm{Cov}_u\big(f(u), g(u)\big) + \E\big[\xi_i \zeta_i \,\big|\, f, g\big]\Big)^2 .
\end{equation*}
\end{proof}

Now let $\widehat{\CKD}{}^2_{\mathrm{avg}}$ denote the standard estimator of \cref{eq:ckd-ustat} computed on the trial-averaged responses $\bar X_{ni}$ and $\bar Y_{ni}$.
Trial-averaging reduces the effect of correlated noise by a factor of $L$: since distinct trials are independent with mean-zero noise, only the $L$ pairs of noise terms on a common trial survive, so $\E[\bar\xi_i \bar\zeta_i \mid f, g] = \E[\xi_i \zeta_i \mid f, g] / L$. 
The kernel $\psi$ of $\widehat{\CKD}{}^2_{\mathrm{avg}}$ is a combination of $\phi$ terms, so applying \cref{lem:same-trial-cov} to the pairs $(f, f')$, $(g, g')$, and $(f, g)$ appearing in $\psi$ gives the following.

\begin{corollary}
\label{cor:avg-bias}
Under (N1)--(N2) and (C3),
\begin{equation*}
  \E\big[\widehat{\CKD}{}^2_{\mathrm{avg}}\big] = \CKD^2 + \frac{2}{L}\, b_1 + \frac{1}{L^2}\, b_2,
\end{equation*}
where, with $f, f' \sim \cF$ and $g, g' \sim \cG$ independent and $\xi, \xi', \zeta, \zeta'$ denoting their respective noise on a common trial,
\begin{align*}
  b_1 &= \E\Big[\mathrm{Cov}_u\big(f(u), f'(u)\big)\, \E[\xi_i \xi'_i \mid f, f']\Big]
       + \E\Big[\mathrm{Cov}_u\big(g(u), g'(u)\big)\, \E[\zeta_i \zeta'_i \mid g, g']\Big] \\
       &\quad- 2\, \E\Big[\mathrm{Cov}_u\big(f(u), g(u)\big)\, \E[\xi_i \zeta_i \mid f, g]\Big], \\
  b_2 &= \E\Big[\E[\xi_i \xi'_i \mid f, f']^2\Big]
       + \E\Big[\E[\zeta_i \zeta'_i \mid g, g']^2\Big]
       - 2\, \E\Big[\E[\xi_i \zeta_i \mid f, g]^2\Big].
\end{align*}
\end{corollary}

\begin{proof}
By \cref{eq:noise-psi}, $\E[\psi] = \E[\phi(f, f')] + \E[\phi(g, g')] - 2\, \E[\phi(f, g)]$. Expanding the square in \cref{lem:same-trial-cov}, with the noise covariance divided by $L$, and averaging over the neurons, the terms of order $L^0$ give $\E[\mathrm{Cov}_u(f(u), f'(u))^2] + \E[\mathrm{Cov}_u(g(u), g'(u))^2] - 2\, \E[\mathrm{Cov}_u(f(u), g(u))^2] = \CKD^2$, and the terms of order $L^{-1}$ and $L^{-2}$ give $2 b_1$ and $b_2$.
\end{proof}

Three features of \cref{cor:avg-bias} are worth noting. First, the bias does not depend on $M$, $N_X$, or $N_Y$, so it is not removed by recording more stimuli or more neurons. Second, because it is a difference of within-system and across-system noise correlations, the bias can be positive or negative: correlated noise can make two systems look either more or less similar than they are. Third, under (N3), same-trial noise is uncorrelated across distinct neurons, so $b_1 = b_2 = 0$ and we recover \cref{prop:noise-unbiased}. Increasing $L$ reduces the bias at rate $1/L$, but for any fixed $L$ it remains.

\subsection{Two trials suffice: the split-trial estimator}
\label{app:corr-noise-split}

By \cref{lem:same-trial-cov}, the bias arises only when two neurons are evaluated on a common trial. Here, we show that these same-trial products can be avoided entirely whenever each input is presented $L \geq 2$ times. We focus on the case $L = 2$, both for simplicity and because two repetitions per input is the minimum needed to remove the bias. Under a fixed budget of trials, this leaves as many trials as possible for presenting distinct stimuli.

For two distinct neurons $f$ and $g$, recorded either from the same system or from different systems, let $\phi^{(1,2)}(f, g; u_{1:4})$ denote $\phi$ evaluated with every factor $f(u_i)$ in \cref{eq:hsic-u-statistic-kernel-defn} replaced by the response of $f$ on the first presentation of $u_i$, and every factor $g(u_i)$ replaced by the response of $g$ on the second presentation. Define $\phi^{(2,1)}$ analogously with the trials swapped, and
\begin{equation}
  \phi_{\mathrm{split}}(f, g; u_{1:4}) := \frac12 \Big[ \phi^{(1,2)}(f, g; u_{1:4}) + \phi^{(2,1)}(f, g; u_{1:4}) \Big] .
  \label{eq:corr-noise-phi-split}
\end{equation}
Averaging over the two assignments makes $\phi_{\mathrm{split}}$ symmetric in $(f, g)$; it is also symmetric in the four inputs.
Thus $\phi_{\mathrm{split}}$ is a valid U-statistic kernel and defines a noise-correlation-debiased HSIC estimator.
Intuitively, this treats the two repetitions as separate recording sessions and combines them so that the two neurons in each product are never read on the same trial.

The corresponding split-trial estimator of $\CKD^2$ uses the kernel
\begin{equation}
\begin{split}
  \psi_{\mathrm{split}}(f_1, f_2, g_1, g_2; u_{1:4})
  &= \phi_{\mathrm{split}}(f_1, f_2; u_{1:4}) + \phi_{\mathrm{split}}(g_1, g_2; u_{1:4}) \\
  &\quad - \frac12 \sum_{a=1}^2 \sum_{b=1}^2 \phi_{\mathrm{split}}(f_a, g_b; u_{1:4}),
\end{split}
  \label{eq:corr-noise-psi-split}
\end{equation}
and is given by
\begin{equation}
\begin{split}
  \widehat{\CKD}{}^2_{\mathrm{split}}
  &= {{N_X}\choose{2}}^{-1} {{N_Y}\choose{2}}^{-1} {{M}\choose{4}}^{-1} \\
  &\quad \times \sum_{n_1 < n_2} \; \sum_{n_1' < n_2'} \; \sum_{i_1 < i_2 < i_3 < i_4} \psi_{\mathrm{split}}\big(f_{n_1}, f_{n_2}, g_{n_1'}, g_{n_2'}; u_{i_1}, u_{i_2}, u_{i_3}, u_{i_4}\big).
\end{split}
  \label{eq:corr-noise-ckd-split}
\end{equation}
This is simply \cref{eq:d2-as-a-u-statistic} with $\phi$ replaced by $\phi_{\mathrm{split}}$.
Both the self terms (two distinct neurons of the same system) and the cross terms (one neuron from each system) are now split across trials, which, as we show next, is sufficient to debias the estimator.

\begin{proposition}[The split-trial estimator is unbiased]
\label{prop:corr-noise-unbiased}
Assume (N1), (C3), and $L = 2$ trials per input. Then $\E\big[ \widehat{\CKD}{}^2_{\mathrm{split}} \given \cD \big] = \widehat{\CKD}{}^2$, the estimator computed on noiseless responses, and hence $\E\big[ \widehat{\CKD}{}^2_{\mathrm{split}} \big] = \CKD^2$.
\end{proposition}

\begin{proof}
Each monomial $f(u_i)\, f(u_j)\, g(u_k)\, g(u_l)$ of $\phi^{(1,2)}(f, g)$ evaluates its four factors on the distinct presentations $(i, 1)$, $(j, 1)$, $(k, 2)$, $(l, 2)$, so given $\cD$ their noise terms are independent with mean zero, and every noise-containing term vanishes in expectation. The same holds for $\phi^{(2,1)}$. The second claim follows from iterated expectations and the unbiasedness of $\widehat{\CKD}{}^2$.
\end{proof}


\paragraph{Efficient computation.}
As for the standard estimator, $\widehat{\CKD}{}^2_{\mathrm{split}}$ can be computed without enumerating index tuples.
Let $\mbx^{(\ell)}_n = (X_{n1\ell}, \dots, X_{nM\ell}) \in \reals^M$ denote the responses of neuron $n$ of system $X$ on repetition $\ell$, let $\widehat{\mbK}^{(\ell)} = N_X^{-1} \sum_n \mbx^{(\ell)}_n \mbx^{(\ell)\top}_n$, and define $\mby^{(\ell)}_n$ and $\widehat{\mbH}^{(\ell)}$ analogously for system $Y$.
Since $S_\song$ in \cref{eq:song-HSIC} is linear in each argument, the reasoning behind \cref{eq:hsic-self-neuron-diagonal-deleted} gives
\begin{equation}
  \widehat{\CKD}{}^2_{\mathrm{split}}
  = S^{(12)}_\sanc(\mbX) + S^{(12)}_\sanc(\mbY)
  - S_\song\big( \widehat{\mbK}^{(1)}, \widehat{\mbH}^{(2)} \big)
  - S_\song\big( \widehat{\mbK}^{(2)}, \widehat{\mbH}^{(1)} \big) ,
  \label{eq:corr-noise-efficient}
\end{equation}
where
\begin{equation}
  S^{(12)}_\sanc(\mbX) = \frac{1}{N_X (N_X - 1)} \bigg( N_X^2\, S_\song\big( \widehat{\mbK}^{(1)}, \widehat{\mbK}^{(2)} \big) - \sum_{n} S_\song\big( \mbx^{(1)}_n \mbx^{(1)\top}_n ,\, \mbx^{(2)}_n \mbx^{(2)\top}_n \big) \bigg) ,
\end{equation}
and $S^{(12)}_\sanc(\mbY)$ is defined analogously.
The subtracted sum removes pairs in which a neuron is compared with itself across repetitions.
Each of its terms pairs two \emph{different} rank-one matrices, but can still be computed in $O(M)$ time: writing $\odot$ for the elementwise product, for $\mbv, \mbw \in \reals^M$,
\begin{equation}
\begin{split}
  S_\song(\mbv \mbv^\top, \mbw \mbw^\top)
  = \frac{1}{M(M-3)} \bigg[
  & (\mbv^\top \mbw)^2 - \|\mbv \odot \mbw\|^2 \\
  & + \frac{\big( (\mathbf 1^\top \mbv)^2 - \|\mbv\|^2 \big)\big( (\mathbf 1^\top \mbw)^2 - \|\mbw\|^2 \big)}{(M-1)(M-2)} \\
  & - \frac{2}{M - 2} \Big( (\mathbf 1^\top \mbv)(\mathbf 1^\top \mbw)(\mbv^\top \mbw) - (\mathbf 1^\top \mbv)\, \mbv^\top (\mbw \odot \mbw) \\
  & \qquad\qquad - (\mathbf 1^\top \mbw)\, (\mbv \odot \mbv)^\top \mbw + \|\mbv \odot \mbw\|^2 \Big) \bigg] ,
\end{split}
  \label{eq:corr-noise-rank-one}
\end{equation}
which reduces to the formula in \Cref{sec:CKD-estimator} when $\mbv = \mbw$.
The cost is therefore of the same order as the standard estimator.

\paragraph{More than two trials.}
If $L > 2$ trials are available, the natural extension is to extend the definition of $\phi_{\mathrm{split}}$ to average over all ordered pairs of trial indices, $\ell \neq \ell'$. The proof of \cref{prop:corr-noise-unbiased} applies verbatim to each pair. This preserves unbiasedness while reducing variance in the estimator that stems from response noise.

\paragraph{Systems recorded separately.}
If the two systems are not recorded simultaneously (e.g., in different animals or sessions), their noise is independent across systems, while noise within each system may still be correlated. By \cref{lem:same-trial-cov}, only the self terms of $\psi$ are then biased; the cross terms $\phi(f, g)$ are unbiased even on trial-averaged responses. It therefore suffices to split the self terms across repetitions and to evaluate the cross terms on trial averages, which uses all available trials and reduces variance. In \cref{eq:corr-noise-efficient}, this amounts to replacing the last two terms with $-2\, S_\song\big(\widehat{\mbK}, \widehat{\mbH}\big)$, where $\widehat{\mbK}$ and $\widehat{\mbH}$ are computed from the trial-averaged responses $\bar X_{ni}$ and $\bar Y_{ni}$.


\subsection{Variance of the split-trial estimator}
\label{app:split-variance}

The previous section showed that the split-trial estimator is unbiased.
Here, we show that its variance decays at the same rate as in our main result, \Cref{th:bound-CKD}, and that correlated noise changes only the coefficient of the $1/M$ term.
We focus on the case of $L=2$ trials or stimulus repetitions for simplicity.
It is possible to extend the results to $L > 2$ by averaging $\phi^{(\ell, \ell')}$ over pairs of distinct repetitions; we omit the details.

The core idea behind our analysis in this section is to attribute all noise correlations to latent variables.
After conditioning on these latent variables, the remaining noise is independent across neurons and trials; as we discuss below, this representation is essentially without loss of generality.
We use the law of total variance to capture contributions from the latent factors and the remaining, decorrelated noise (which is handled by our previous analysis in \Cref{app:noise-theory}).

Concretely, for each input $u_i$ and repetition $\ell \in \{1, 2\}$, let $\omega_i^{(\ell)}$ be a latent trial state that captures all noise correlations on that trial.
We assume the latent states are i.i.d.\ across inputs and repetitions and independent of the neurons and inputs. 
We then model the response of each neuron to input $u_i$ on repetition $\ell$ as
\begin{equation}
  \label{eq:latent-variable-tuning-model}
  X_{ni\ell} = f_n\big(u_i, \omega_i^{(\ell)}\big) + \xi_{ni\ell},
  \qquad
  Y_{ni\ell} = g_n\big(u_i, \omega_i^{(\ell)}\big) + \zeta_{ni\ell},
\end{equation}
where, conditional on the neurons, inputs, and trial states, the private noise variables $\xi_{ni\ell}$ and $\zeta_{ni\ell}$ are mutually independent and mean zero.
Intuitively, each tuning curve now depends jointly on the input and the latent trial state, and the usual tuning curve is recovered by averaging over the trial state: $f_n(u) = \E_\omega[f_n(u, \omega)]$, and likewise for $g_n$.

To analyze the variance of $\widehat{\CKD}{}^2_{\mathrm{split}}$, we attach both trial states of each input to the input itself, defining augmented inputs
\begin{equation}
u^{+}_i := \big(u_i, \omega_i^{(1)}, \omega_i^{(2)}\big),
\end{equation}
which are i.i.d.\ and independent of the neurons. For an index tuple $(n_1 < n_2;\; n_1' < n_2';\; i_1 < i_2 < i_3 < i_4)$, let $\Sigma$ denote the collection of random variables on which the corresponding term $\psi_{\mathrm{split}}(\Sigma)$ depends: the neurons $f_{n_1}, f_{n_2}, g_{n_1'}, g_{n_2'}$, the augmented inputs $u^{+}_{i_1}, \dots, u^{+}_{i_4}$, and the private noise of each of these neurons on each presentation of these inputs.
Two such collections $\Sigma$ and $\Sigma'$ share a latent trial state only if they share the corresponding input, and share a private noise variable only if they share both the corresponding neuron and input. All of their remaining variables are distinct and mutually independent, so $\psi_{\mathrm{split}}(\Sigma)$ and $\psi_{\mathrm{split}}(\Sigma')$ are conditionally independent given $\Sigma \cap \Sigma'$. The standard decomposition of variance for a U-statistic (see \cref{sec:appendix-U-stats-HSIC,sec:appendix-b}) therefore gives
\begin{equation}
  \Var\big[\widehat{\CKD}{}^2_{\mathrm{split}}\big] = \sum_{\alpha,\beta,\gamma} \tilde C_{\alpha\beta\gamma} V^{\mathrm{corr}}_{\alpha\beta\gamma},
  \qquad
  V^{\mathrm{corr}}_{\alpha\beta\gamma} := \Var\big(\E[\psi_{\mathrm{split}}(\Sigma) \mid \Sigma \cap \Sigma']\big),
  \label{eq:split-var-decomp}
\end{equation}
where $\tilde C_{\alpha\beta\gamma}$ is the probability that two independently chosen index tuples share $\alpha$ neurons of system $X$, $\beta$ neurons of system $Y$, and $\gamma$ inputs. These weights are the same as in \cref{sec:appendix-b}, because the block structure is unchanged.

As in \cref{app:noise-insensitive}, let $A$ denote the shared neurons and shared inputs $u$, excluding latent trial states and private noise. By the law of total variance,
\begin{equation}
  V^{\mathrm{corr}}_{\alpha\beta\gamma}
  = \Var_A\big(\E[\psi_{\mathrm{split}}(\Sigma) \mid A]\big)
  + \E_A\Big[\Var_{(\Sigma \cap \Sigma') \setminus A}\big(\E[\psi_{\mathrm{split}}(\Sigma) \mid \Sigma \cap \Sigma'] \,\big|\, A\big)\Big].
  \label{eq:split-total-var}
\end{equation}
The first term of \cref{eq:split-total-var} is the noiseless contribution, and the second is the contribution of the noise. The following proposition shows that the noise affects only collision patterns that share an input.

\begin{proposition}
\label{prop:split-first-order}
$V^{\mathrm{corr}}_{\alpha\beta\gamma} \ge V_{\alpha\beta\gamma}$ for every collision pattern, with equality whenever $\gamma = 0$. In particular, $V^{\mathrm{corr}}_{100} = V_{100}$ and $V^{\mathrm{corr}}_{010} = V_{010}$, while
\begin{equation}
  V^{\mathrm{corr}}_{001} = V_{001} + \E_u\Big[\Var_{\omega^{(1)}, \omega^{(2)}}\big(\E[\psi_{\mathrm{split}} \mid u^{+} ] \,\big|\, u\big)\Big],
  \label{eq:V001-corr}
\end{equation}
where $u^{+} = (u, \omega^{(1)}, \omega^{(2)})$ is the shared augmented input.
\end{proposition}

\begin{proof}
Consider the two terms of \cref{eq:split-total-var}.
For the first, $A$ is a function of $\cD$, so the tower property and \cref{prop:corr-noise-unbiased} give $\E[\psi_{\mathrm{split}}(\Sigma) \mid A] = \E[\psi(\Sigma) \mid A]$, with $\psi$ evaluated on noiseless responses. Hence the first term equals $V_{\alpha\beta\gamma}$.
The second term is nonnegative, which gives $V^{\mathrm{corr}}_{\alpha\beta\gamma} \ge V_{\alpha\beta\gamma}$.
If the two tuples share no input ($\gamma = 0$), they share no latent trial states and no private noise, so $(\Sigma \cap \Sigma') \setminus A$ is empty and the second term vanishes. This covers the patterns $(1, 0, 0)$ and $(0, 1, 0)$.
For the pattern $(0, 0, 1)$, the two tuples share one augmented input $u^+$ and no neurons, and therefore no private noise. Thus $A = \{u\}$ and $(\Sigma \cap \Sigma') \setminus A = \{\omega^{(1)}, \omega^{(2)}\}$. Since $\psi_{\mathrm{split}}$ is symmetric in its inputs, it does not matter which input is shared, and the second term of \cref{eq:split-total-var} reduces to the second term of \cref{eq:V001-corr}.
\end{proof}

It remains to bound the extra noise term in \cref{eq:V001-corr}, as well as the higher-order collision terms. Both are controlled by the second moment of $\psi_{\mathrm{split}}$, which in turn depends on the fourth moment of single-trial responses,
\begin{equation}
  \tilde q = \max \Big\{ \sup_u \E \big[ (X_{ni\ell} - \bar f_n)^4 \,\big|\, u_i = u \big], \ \sup_u \E \big[ (Y_{ni\ell} - \bar g_n)^4 \,\big|\, u_i = u \big] \Big\},
  \label{eq:fourth-moment-defn-corr}
\end{equation}
where the expectations are jointly over a randomly sampled neuron, a latent trial state, and private noise. This is \cref{eq:fourth-moment-defn} with each tuning value $f_n(u_i)$ replaced by a single-trial response $X_{ni\ell}$, still centered at the noiseless stimulus average $\bar f_n$. Writing $X_{ni\ell} - \bar f_n = \big(f_n(u_i) - \bar f_n\big) + \big(X_{ni\ell} - f_n(u_i)\big)$ and using $(a + b)^4 \le 8a^4 + 8b^4$ shows that $\tilde q$ is at most $8q$ plus eight times the largest fourth moment of the single-trial deviation $X_{ni\ell} - f_n(u_i)$, which combines shared and private noise.

\begin{lemma}
\label{lem:split-psi-moment}
$\Var[\psi_{\mathrm{split}}] \le 256\, \tilde q^{\,2}$.
\end{lemma}

\begin{proof}
The argument follows the corresponding bound in \cref{app:proof-of-bound-CKD}, with $q$ replaced by $\tilde q$. By \cref{lem:ckd-phi-invar-const}, we may center each neuron by its noiseless stimulus average, so assume $\bar f = \bar g = 0$. Each monomial of $\phi^{(1,2)}(f, g)$ is then a product of two responses of $f$ on repetition $1$, at inputs $u_i$ and $u_j$, and two responses of $g$ on repetition $2$, at inputs $u_k$ and $u_l$. Given the inputs, the two responses of $f$ depend only on $f$, the trial states $\omega_i^{(1)}, \omega_j^{(1)}$, and the private noise of $f$, while the two responses of $g$ depend only on $g$, the trial states $\omega_k^{(2)}, \omega_l^{(2)}$, and the private noise of $g$. Since $f \neq g$ are independent and the two neurons are read on different repetitions, these are independent, even when $i = k$ or $j = l$. (This is where the split-trial construction is needed: for the standard estimator, two neurons evaluated at the same input share its trial state.) Hence, applying Cauchy--Schwarz within each neuron,
\begin{equation*}
  \E\Big[ \big(X_{ni1} X_{nj1}\, X_{n'k2} X_{n'l2}\big)^2 \,\Big|\, u_{1:4} \Big]
  = \E\big[ X_{ni1}^2 X_{nj1}^2 \,\big|\, u_{1:4} \big]\, \E\big[ X_{n'k2}^2 X_{n'l2}^2 \,\big|\, u_{1:4} \big]
  \le \tilde q^{\,2},
\end{equation*}
with $Y$ in place of $X$ for neurons of system $Y$. Every monomial therefore has $L^2$ norm at most $\tilde q$. By the triangle inequality, $\|\phi^{(1,2)}(f, g)\|_2 \le 4 \tilde q$, and the same bound holds for $\phi^{(2,1)}(f, g)$ and hence for $\phi_{\mathrm{split}}(f, g)$. Combining the two self terms and four cross terms of $\psi_{\mathrm{split}}$ as in \cref{app:proof-of-bound-CKD} gives $\|\psi_{\mathrm{split}}\|_2 \le 16 \tilde q$, and hence $\Var[\psi_{\mathrm{split}}] \le 256\, \tilde q^{\,2}$.
\end{proof}

Since the variance of a conditional expectation is at most the variance of the original quantity, the extra term in \cref{eq:V001-corr} is at most $\Var[\psi_{\mathrm{split}}] \le 256\, \tilde q^{\,2}$, and likewise $V^{\mathrm{corr}}_{\alpha\beta\gamma} \le 256\, \tilde q^{\,2}$ for every pattern. Combining these bounds with \cref{prop:split-first-order} gives our final result.

\begin{corollary}[Variance bound for the split-trial estimator]
\label{cor:split-variance}
\begin{equation*}
  \Var\big[\widehat{\CKD}{}^2_{\mathrm{split}}\big]
  \le 16\, q\, \CKD^2 \Big(\frac1M + \frac1{N_X} + \frac1{N_Y}\Big)
  + \frac{2^{12}\, \tilde q^{\,2}}{M}
  + 2^{16}\, \tilde q^{\,2} \Big(\frac1M + \frac1{N_X} + \frac1{N_Y}\Big)^2 .
\end{equation*}
\end{corollary}

\begin{proof}
Group the terms of \cref{eq:split-var-decomp} by the number of collisions, $\alpha + \beta + \gamma$. For single collisions, $\tilde C_{100} \le 4/N_X$, $\tilde C_{010} \le 4/N_Y$, and $\tilde C_{001} \le 16/M$ (\cref{app:proof-of-bound-CKD}). By \cref{prop:split-first-order}, these terms equal their noiseless counterparts, each at most $q\, \CKD^2$, except for the extra term in $V^{\mathrm{corr}}_{001}$, which is at most $256\, \tilde q^{\,2}$. Together they contribute at most $16\, q\, \CKD^2 (1/M + 1/N_X + 1/N_Y) + 16 \cdot 256\, \tilde q^{\,2}/M$. Patterns with two or more collisions have total weight at most $256\,(1/M + 1/N_X + 1/N_Y)^2$, as in \cref{app:proof-of-bound-CKD}, and each $V^{\mathrm{corr}}_{\alpha\beta\gamma} \le 256\, \tilde q^{\,2}$.
\end{proof}

Correlated noise therefore leaves the rate unchanged and affects only the coefficient of $1/M$. Unlike the other first-order terms, however, this additional contribution is not proportional to $\CKD^2$, so it does not vanish when the two representations are identical. Intuitively, an input whose trial states happen to push the two systems in different directions adds variance to the estimator, even when the underlying tuning is the same.

\paragraph{Generality of the latent representation.}
The representation in \cref{eq:latent-variable-tuning-model} is not restrictive. 
Suppose that, on each trial, the joint law of responses is unchanged by relabeling neurons within each system, as is implicit in modeling neurons as i.i.d.\ draws from a population. 
Then, by de Finetti's theorem, the responses on that trial are conditionally i.i.d.\ given a latent variable.
The latent variable supplied by de Finetti's theorem may have a law that depends on the input $u_i$.
However, it can always be written as a measurable function of $u_i$ and an independent random variable uniformly distributed on $[0, 1]$, essentially by the inverse-CDF construction \citep[Lemma 4.22]{kallenberg2021foundations}.
We take this uniform variable as $\omega_i^{(\ell)}$, absorbing the dependence on the input into the augmented tuning function, so that $\omega_i^{(\ell)}$ is independent of the input.
Letting $f_n(u, \omega)$ denote the conditional mean of $X_{ni\ell}$ given the tuning curve of neuron $n$, $u_i = u$, and $\omega_i^{(\ell)} = \omega$, and setting $\xi_{ni\ell} := X_{ni\ell} - f_n(u_i, \omega_i^{(\ell)})$, then recovers the form above.

\subsection{Empirical Results}
\begin{figure}

    \centering
    \includegraphics[width=1.0\linewidth]{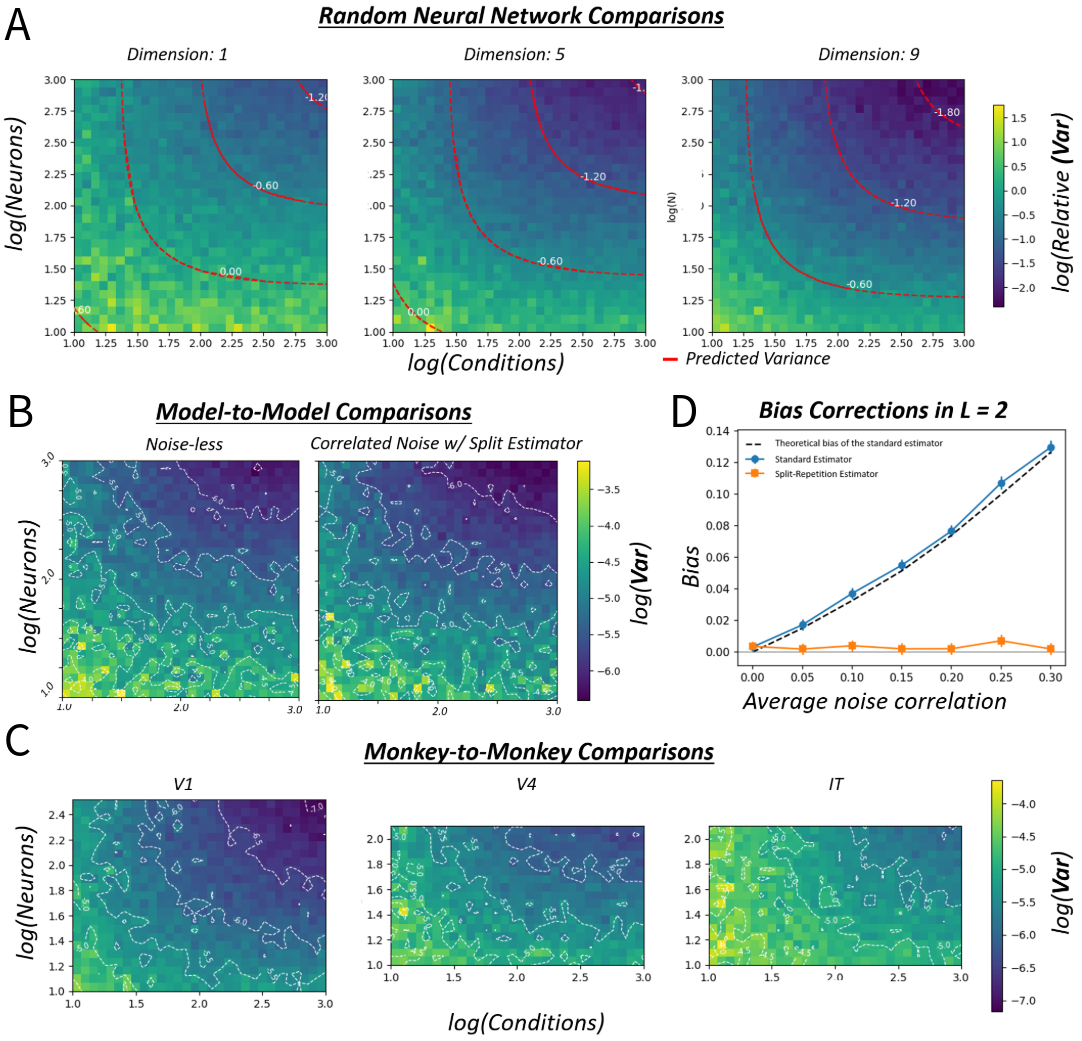}
    \caption{Reproduction of core simulations for data with noise correlations. (\textbf{A}) Variance of estimating CKD between random neural networks as in Section~\ref{sec:random-NNs}. (\textbf{B}) Variance of estimating CKD between trained neural networks as in Section~\ref{sec:deep-NNs}. On the left, we show the noiseless results we obtained in \Cref{sec:deep-NNs} to demonstrate that the variance scales almost exactly. (\textbf{C}) Variance of estimating CKD between monkey regions as in Section~\ref{sec:TVSD}. (\textbf{D}) Increasing strength of noise correlation induces bias in CKD estimates using the standard presented estimator. Results in this subfigure are in random neural network simulations.}
    \label{fig:noisecorr-empirics}
    
\end{figure}

In \Cref{fig:noisecorr-empirics}, we present empirical results supporting our theoretical predictions when noise is correlated.
Below, we apply correlated Gaussian noise to $L = 2$ trials, then estimate CKD using our split trial estimator $\widehat{\CKD}{}^2_{\mathrm{split}}$ over the two trials.
We reproduce qualitatively and numerically the same results that we presented in the main body of the paper.
In \Cref{fig:noisecorr-empirics}B, we also present the noiseless results from \Cref{sec:deep-NNs} for ease of viewing: results are near-identical when using the split-trial estimator.
In \Cref{fig:noisecorr-empirics}D, we show the empirical bias of the standard estimator we presented in the paper, which increases as the strength of noise correlations increase.
This bias follows our theory well.
This bias vanishes when using our split-trial estimator with just $L = 2$ trials, and debiases for all degrees of noise correlation.

\section{What is different about CKA?}\label{app:cka_differences}

\subsection{No Unbiased Estimators for CKA and Squared Procrustes Distance}
\label{app:no-unbiased}

We have seen that $\HSIC$ admits an unbiased U-statistic estimator, leading to an unbiased estimator of $\CKD^2$.
Here, we show that it is not possible to completely eliminate estimator bias in other popular representational similarity measures including $\CKA$ and the squared Procrustes distance.

$\CKA$ is a nonlinear combination of three $\HSIC$ terms, which are unbiasedly estimable.
However, unbiasedness is not preserved under nonlinear transformations.
The textbook example is that $1/p$ is not unbiasedly estimable from $n$ Bernoulli trials for any $n$ \citep[see][example 1.2 in Chapter 2]{lehmann1998theory}.
The proposition below constructs a similar argument for $\CKA$, showing that it is not possible to achieve unbiased estimation across all problem instances.

\begin{proposition}
\label{prop:no-unbiased-cka}
Let $\cU = [0,1]$ with $P_U$ uniform, and fix $M \geq 1$ and
$N_X, N_Y \geq 1$. There is a one-parameter family of tuning-curve
distributions such that no estimator
$\hat T$ of $\CKA$ computed from the observed response matrices, and having
finite expectation at some interior parameter value, is unbiased on any
subinterval of $[0,1]$.
\end{proposition}

\begin{proof}
Let $\varphi_1(u) = \sqrt2\cos(2\pi u)$ and
$\varphi_2(u) = \sqrt2\cos(4\pi u)$, which are centered, bounded, and orthonormal in
$L^2(P_U)$. 
Take $\cF$ to be the point mass at $\varphi_1$, and let $\cG$ place mass $p$ on $\varphi_1$ and $1-p$ on
$\varphi_2$. 
That is, for $f \sim \cF$, we have $f = \varphi_1$ with probability one, while
for $g \sim \cG$ we have $g = \varphi_1$ with probability $p$ and
$g = \varphi_2$ with probability $1-p$.

Both curves are centered, so
\begin{equation}
\bar{k} = \varphi_1\otimes\varphi_1,
\qquad
\bar{h}_p = p\,\varphi_1\otimes\varphi_1 + (1-p)\,\varphi_2\otimes\varphi_2 ,
\end{equation}
and it is easy to check that
\begin{equation}
\langle \bar{k} , \bar{h}_p \rangle = p,
\qquad
\Vert \bar{k} \Vert^2 = 1,
\qquad
\Vert \bar{h}_p \Vert^2 = p^2 + (1-p)^2 = 2p^2-2p+1.
\end{equation}
Thus we have constructed a family of problem instances where 
\begin{equation}
\CKA_\infty = \frac{p}{\sqrt{2p^2-2p+1}}
\label{eq:cka-of-p}
\end{equation}
controlled by an arbitrary parameter $p \in [0, 1]$.

Crucially, the right hand side of \cref{eq:cka-of-p} is \emph{not expressible as a polynomial} in $p$.
Below, we show that every estimator has an expected value that \emph{is a polynomial} in $p$.
The two can therefore never exactly agree over any subinterval $p \in [0, 1]$, which is sufficient to
conclude the proof.

Let $\nu$ denote the number of tuning curves drawn from $\cG$ that are equal to $\varphi_1$.
Given our construction, $\nu$ is a binomial random variable with $N_Y$ trials and success probability $p$.
Now let $\widehat{T}$ denote any estimator and let $t_\nu = \E[\widehat{T} \mid \nu ]$ denote the 
conditional expectation of the estimator given $\nu$.
Crucially, $t_\nu$ is a constant with respect to $p$ since the count variable $\nu$
contains all information about $p$ that is available to any estimator.
By the law of total expectation followed by the expression for the expected value of a binomial
random variable we have:
\begin{equation}
\E[\widehat{T}] = \E_{\nu} [ t_\nu ] = \sum_{\nu=0}^{N_Y} \binom{N_Y}{\nu} p^{\nu}(1-p)^{N_Y-\nu}\, t_\nu.
\end{equation}
This is a polynomial in $p$ of degree at most $N_Y$, thus concluding the proof.
\end{proof}

The same construction shows that the squared Procrustes distance is also
not unbiasedly estimable.
Recall that,
by the duality of \citet{harvey_duality_2023}, the squared Procrustes distance
between normalized representations is the squared Bures distance between the
centered kernel operators,
\begin{equation}
\Proc^2_\infty(\bar{k},\bar{h}) \;=\; \Tr\bar{k} \;+\; \Tr\bar{h} \;-\; 2\, 
\Tr\Big[\big(\bar{k}^{1/2}\bar{h}\,\bar{k}^{1/2}\big)^{1/2}\Big].
\label{eq:bures}
\end{equation}

\begin{proposition}
\label{prop:no-unbiased-procrustes}
Under the hypotheses of Proposition~\ref{prop:no-unbiased-cka}, and on the
same family of tuning-curve distributions, no estimator of $\cP^2$ with
finite expectation at some interior parameter value is unbiased on any
subinterval of $[0,1]$.
\end{proposition}

\begin{proof}
Take $\cF$ and $\cG$ exactly as in the proof of
Proposition~\ref{prop:no-unbiased-cka}. Since $\varphi_1$ is a unit vector,
$\bar{k} = \bar{k}^{1/2} = \varphi_1\otimes\varphi_1$.
Orthogonality of $\varphi_1$ and $\varphi_2$ then gives
\begin{equation}
\bar{k}^{1/2}\,\bar{h}_p\,\bar{k}^{1/2}
\;=\; p\;\varphi_1\otimes\varphi_1 ,
\end{equation}
whose square root is $\sqrt{p}\,\varphi_1\otimes\varphi_1$. Thus,
the Bures fidelity term is
\begin{equation}
\Tr\Big[\big(\bar{k}^{1/2}\bar{h}\,\bar{k}^{1/2}\big)^{1/2}\Big] = \sqrt{p},
\end{equation}
while $\Tr\bar{k} = 1$ and $\Tr\bar{h}_p = p + (1-p) = 1$, so that
\begin{equation}
\Proc^2_\infty \;=\; 2 - 2\sqrt{p} .
\label{eq:proc-of-p}
\end{equation}
The right-hand side of \cref{eq:proc-of-p} is not a polynomial in $p$.
The rest of the proof carries over verbatim from \cref{prop:no-unbiased-cka}:
the expectation of any estimator is a polynomial in $p$, and therefore
no estimator can be unbiased on this family of problem instances.
\end{proof}

The inability of CKA or Procrustes to be estimated with an unbiased estimator has implications for their variance as well.
Crucially, because no unbiased estimator exists for either CKA or Procrustes, then neither measure can be constructed as a U-statistic.
Hence, the approaches we take in Appendix~\ref{sec:appendix-U-stats-HSIC} and Appendix~\ref{sec:appendix-b} no longer apply.
Indeed, empirical simulations in \Cref{app:no-unbiased} reveals that the variance of CKA behaves quite nonlinearly compared to the ordered theory we derive for HSIC and CKA.

\subsection{The consistency of the plug-in estimator of CKA depends on the order of limits and the structure of the kernel functions}
\label{app:cka-order-of-limits}

In the main text, we noted that 
\begin{align}
    \lim_{\min(M,N_{X},N_{Y}) \to \infty} S_{\mathrm{pi}}(\mbX, \mbY) = \langle \bar{k},\bar{h} \rangle_{\mathrm{HS}},
\end{align}
and that in this sense the plug-in estimators of both HSIC and CKD are consistent irrespective of how one takes $M, N_X, N_Y \to \infty$. Here, we show that CKA is not in general consistent if one takes a proportional high-dimensional limit in which we take $M,N_{X},N_{Y} \to \infty$ such that
\begin{align}
    \frac{M}{N_X} \to q_{X} \in (0,\infty) \quad \textrm{and} \quad \frac{M}{N_Y} \to q_{Y} \in (0,\infty) . 
\end{align}
In particular, depending on the properties of the population kernel function, the limit of the plug-in estimator may deviate by an order-one amount from the desired limit. This result is known, for example, from \citet{chun2025estimating}, but we give a self-contained though informal derivation here. 

Consider, for simplicity, the case in which the tuning curves are Gaussian, such that the empirical kernel matrices $\widehat{\mbK}$ and $\widehat{\mbH}$ are Wishart. Expanding the definition, the plug-in estimator of the CKA is
\begin{align}
    \widehat{\mathrm{CKA}}_{\mathrm{pi}}(\mbX,\mbY) = \frac{\frac{1}{M}\Tr(\mbC \widehat{\mbK} \mbC \widehat{\mbH})}{\sqrt{\frac{1}{M}\Tr(\mbC \widehat{\mbK} \mbC \widehat{\mbK}) \frac{1}{M}\Tr(\mbC \widehat{\mbH} \mbC \widehat{\mbH}) }}. 
\end{align}
Now, using standard results on Wishart matrix moments, we have that the expectations of the three terms with respect to the Gaussian distribution of tuning curves are
\begin{align}
    \mathbb{E}_{f,g} \frac{1}{M}\Tr(\mbC \widehat{\mbK} \mbC \widehat{\mbH}) = \frac{1}{M} \Tr(\mbC \mbK \mbC \mbH),
\end{align}
\begin{align}
    \mathbb{E}_{f,g} \frac{1}{M}\Tr(\mbC \widehat{\mbK} \mbC \widehat{\mbK}) = \left(1 + \frac{1}{N_X}\right) \frac{1}{M} \Tr(\mbC \mbK \mbC \mbK) + \frac{M}{N_{X}} \left(\frac{1}{M} \Tr(\mbC \mbK) \right)^2,
\end{align}
and analogously 
\begin{align}
    \mathbb{E}_{f,g} \frac{1}{M}\Tr(\mbC \widehat{\mbH} \mbC \widehat{\mbH}) = \left(1 + \frac{1}{N_Y}\right) \frac{1}{M} \Tr(\mbC \mbH \mbC \mbH) + \frac{M}{N_{Y}} \left(\frac{1}{M} \Tr(\mbC \mbH) \right)^2. 
\end{align}
Under standard regularity assumptions on the kernels $k$ and $h$, the remaining traces are well-behaved, and moreover the three normalized traces concentrate to their expectations with respect to both neuron and stimulus sampling. This gives, somewhat schematically, 
\begin{align}
    \widehat{\mathrm{CKA}}_{\mathrm{pi}}(\mbX,\mbY) \sim  \frac{\mathrm{CKA}(\mbK,\mbH)}{\sqrt{(1+d_{X})(1+ d_{Y})}},
\end{align}
where we have defined the normalized effective dimensionalities
\begin{align}
    d_{X} = \frac{1}{N_X} \frac{\left( \Tr(\mbC \mbK) \right)^2}{\Tr(\mbC \mbK \mbC \mbK)}
    \quad \textrm{and} \quad 
    d_{Y} = \frac{1}{N_Y} \frac{\left( \Tr(\mbC \mbH) \right)^2}{\Tr(\mbC \mbH \mbC \mbH)}. 
\end{align}
This is the primary high-dimensional bias result of \citet{chun2025estimating}. Here, we have been quite informal: we have not specified the notion of asymptotic equivalence in $\sim$, nor the required regularity on the kernels. But, all of this could be made precise, and the qualitative result stands: if we consider a limit such that $d_X$ and $d_{Y}$ do not tend to zero sufficiently rapidly, then the plug-in estimator of the CKA acquires an order-one multiplicative bias. This is, importantly, dependent on the spectral properties of the kernel: roughly speaking if $k$ and $h$ are trace-class then $d_{X}$ and $d_{Y}$ should tend to zero even if $M$ is taken to infinity proportionally with $N_{X}$ and $N_{Y}$, while if their traces scale with $M$ then $M$ must be taken to infinity strictly more slowly. The latter is the usual high-dimensional case, while in the former the data have non-extensive intrinsic dimension. 

The most extreme, and most illustrative, high-dimensional example is if we considered two independent, isotropic Gaussian clouds, \textit{i.e.}, if $\mbK = \mbH = \mathbf{I}_{M}$. This corresponds to an infinitely-sharp kernel that is one if two stimuli coincide and zero otherwise. In this case, we of course have that the CKA between $\mbK$ and $\mbH$ is 
\begin{align}
    \mathrm{CKA}(\mbK,\mbH) = 1,
\end{align}
as they are the same. However, 
\begin{align}
    d_{X} = \frac{M-1}{N_{X}} \sim q_{X} \quad \textrm{and} \quad d_{Y} = \frac{M-1}{N_{Y}} \sim q_{Y} ,
\end{align}
so
\begin{align}
    \widehat{\mathrm{CKA}}_{\mathrm{pi}} \sim \frac{1}{\sqrt{(1+q_{X})(1+q_{Y})}},
\end{align}
which may be substantially biased downward from one if $q_{X}$ and $q_{Y}$ are not small.


\subsection{Experimental Results on HSIC and CKA}
\label{app:hsic-cka-empirics}

\begin{figure}

    \centering
    \includegraphics[width=1.0\linewidth]{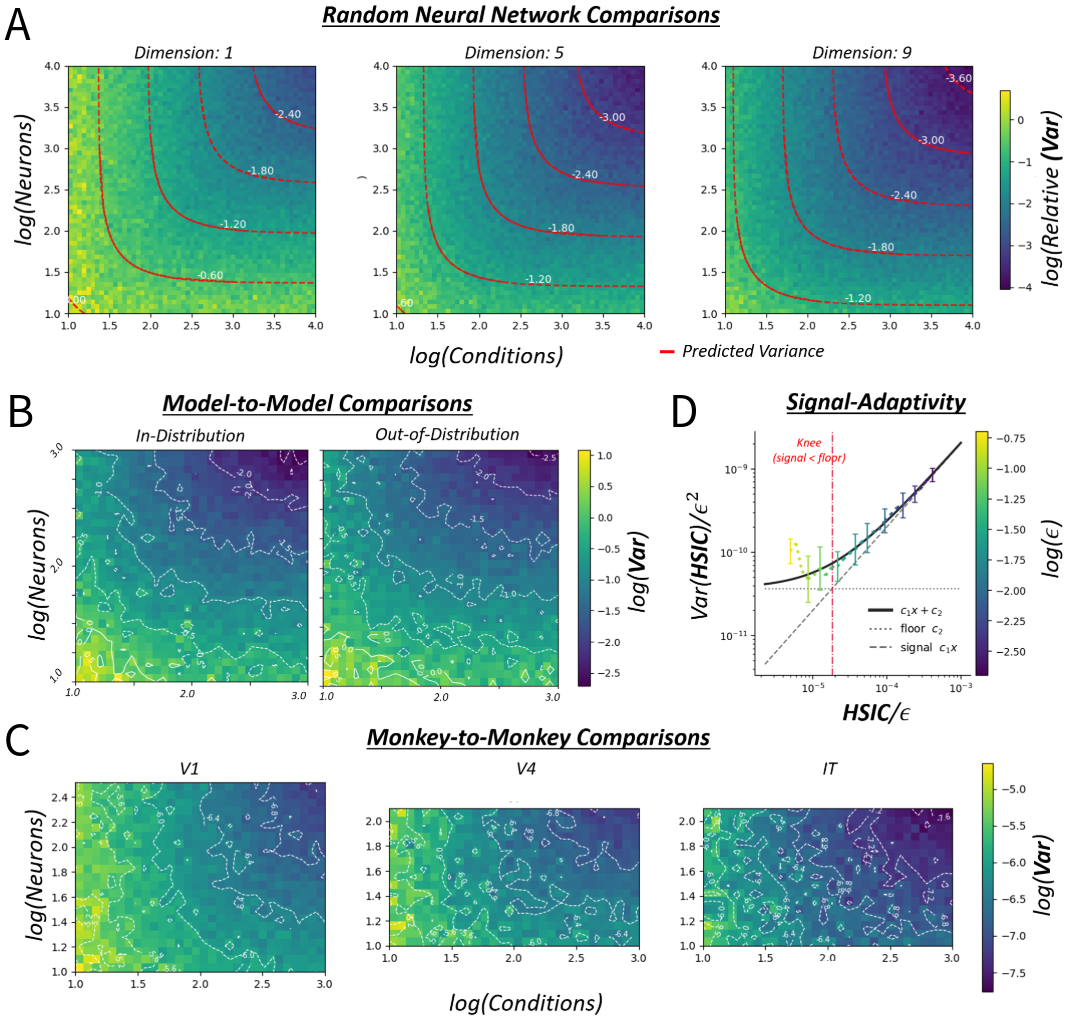}
    \caption{Reproduction of results for CKD in the context of HSIC. (\textbf{A}) Variance of estimating HSIC between random neural networks as in Section~\ref{sec:random-NNs}. (\textbf{B}) Variance of estimating HSIC between trained neural networks as in Section~\ref{sec:deep-NNs}. (\textbf{C}) Variance of estimating HSIC between monkey regions as in Section~\ref{sec:TVSD}. (\textbf{D}) Signal adaptivity of HSIC.}
    \label{fig:hsic-empirics}
    
\end{figure}
Here, we show the results for the behavior of $\Var(\mathrm{HSIC})$ in the context of the main experiments presented in the main body of the paper.
The results presented here are for the unbiased $S_\song$ estimator, but are qualitatively replicated when using the standard plug-in estimator.
Given that we derive a similar upper bound on variance scaling for HSIC as with CKD in Appendix~\ref{sec:appendix-U-stats-HSIC}, it is no surprise that we demonstrate similar qualitative and quantitative results for HSIC as in CKD.

\begin{figure}

    \centering
    \includegraphics[width=1.0\linewidth]{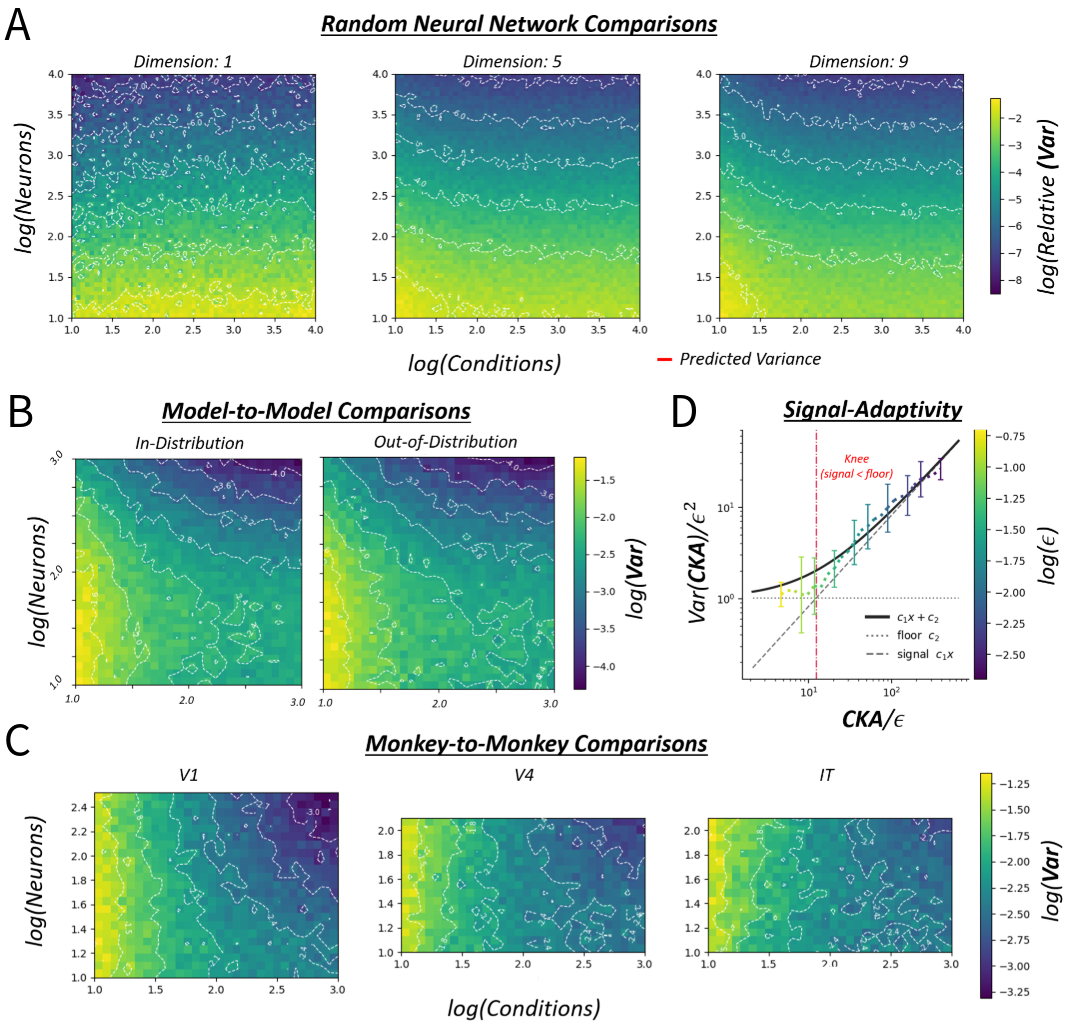}
    \caption{Reproduction of results for CKD in the context of CKA. Note scaling behavior differs wildly. (\textbf{A}) Variance of estimating CKA between random neural networks as in Section~\ref{sec:random-NNs}. (\textbf{B}) Variance of estimating CKA between trained neural networks as in Section~\ref{sec:deep-NNs}. (\textbf{C}) Variance of estimating CKA between monkey regions as in Section~\ref{sec:TVSD}. (\textbf{D}) Signal adaptivity of CKA.}
    \label{fig:cka-empirics}
    
\end{figure}
Next, we present the results for the behavior of $\Var(\mathrm{CKA})$ in the context of the main experiments from the main body of the paper.
CKA behaves markedly different in all these scenarios, where its dependence on changes in $M$ and $N$ vary wildly on both setting and magnitude.

\begin{figure}

    \centering
    \includegraphics[width=1.0\linewidth]{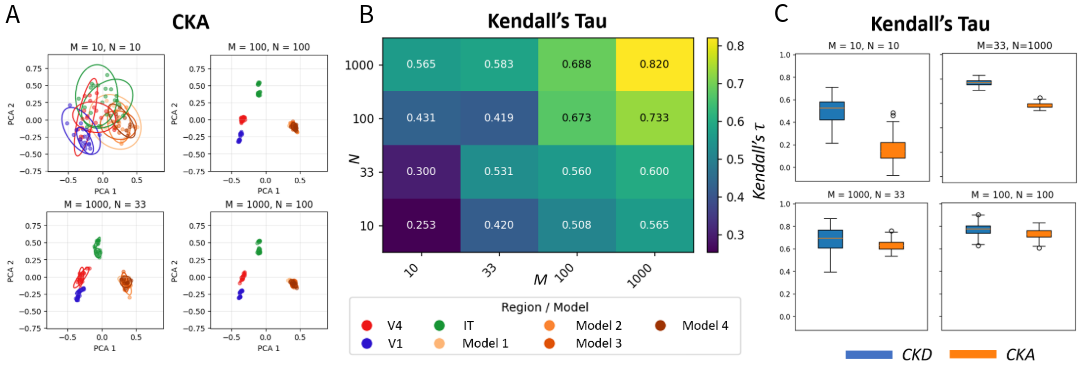}
    \caption{Clustering experiments for CKA. (\textbf{A}) Projection of estimated CKA between 4 deep models and 6 neural recordings from 3 different brain areas. (\textbf{B}) Estimates of Kendall's $\tau$ on different degrees of subsampling. (\textbf{C}) Comparisons between Kendall's $\tau$ in CKD with CKA.}
    \label{fig:cka-empirics-clusters}
    
\end{figure}

Last, we plot the same trends but for CKA (Figure~\ref{fig:cka-empirics-clusters}A-B), which produces similar qualitative results.
This is suprising because CKA admits no unbiased estimator, and can diverge qualitatively from the predicted $\mathcal{O}(\frac1M + \frac1N)$ scaling law when comparing one model to another.
This suggests that while CKA admits no unbiased estimator, variance in similarity estimation in the top 2 principal components follows the established theoretical trend well, so our experimental prescriptions may hold for a larger class of similarity metrics.

Notably, the Kendall's $\tau$ for CKA are decreased relative to CKD (Figure~\ref{fig:cka-empirics-clusters}C).
Since CKA is scale-invariant while CKD is not, this suggests that the difference in scale across multiple neural regions and models may be important in understanding multi-system relationships. 
In other words, the scale invariance of CKA, when comparing multiple monkeys and models to one another, surprisingly hurts one's ability to determine the proper ordering of the their pairwise relationships in concert, even when comparing across models and neural data.
This difference in accuracy is particularly true when recording volume is low, suggesting that using CKD (an estimator that does not normalize the scale of representations) attains more accurate reads of the relationships between the subjects than using CKA (an estimator that normalizes the scales of representations to be equal).

Our explanation for this phenomenon is not necessarily surprising, but ill-documented in the literature.
The reason for this is because estimating the true scale of a representation is not known during an experiment where finite $M$ and $N$ are sampled.
Therefore, a similarity measure that demands normalization by scale necessarily would require more samples to attain an accurate estimate.
In the case of CKA, not only does one need to get an accurate estimate of $S(K, H)$ (the numerator), but also $\sqrt{S(K, K)S(H, H)}$ (the denominator). 
Thus, when sample volume is low, errors propagate farther.
\section{Variance Scaling when \texorpdfstring{$N_X \neq N_Y$}{NX != NY}}
\label{app:diffN}

\begin{figure}

    \centering
    \includegraphics[width=1.0\linewidth]{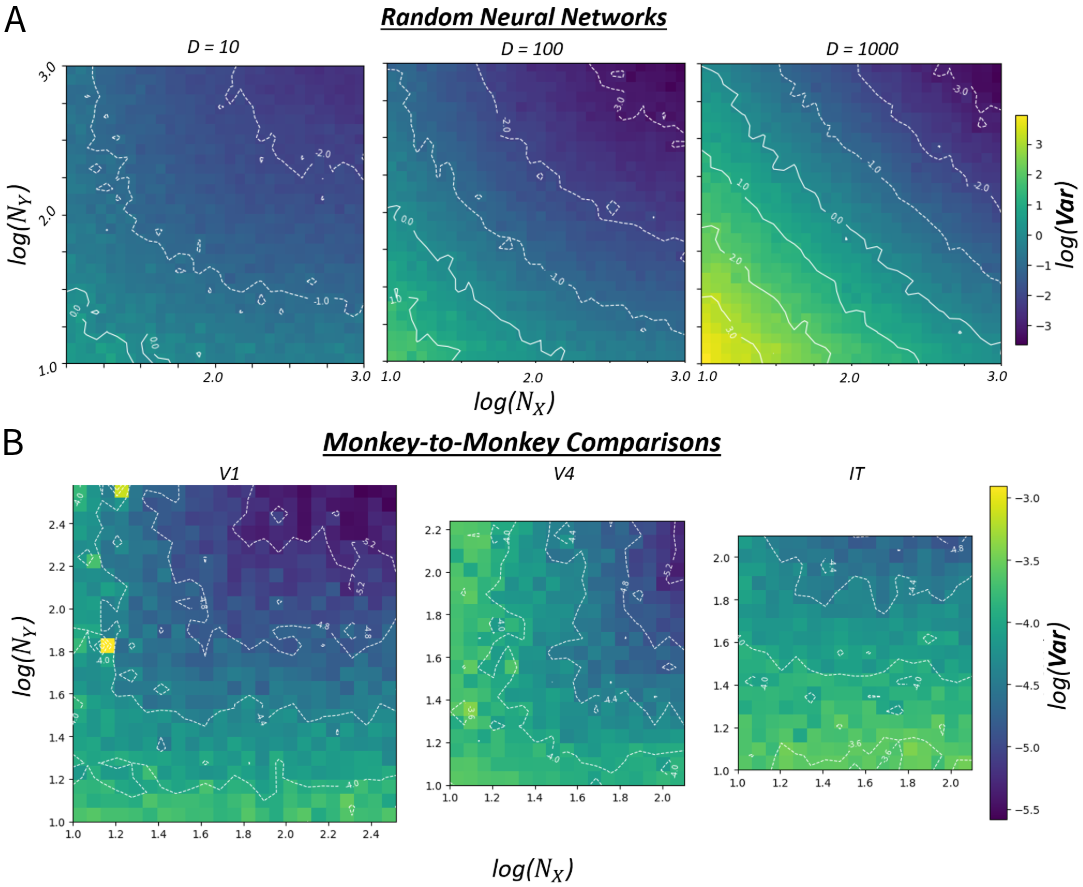}
    \caption{Heatmaps of variance scaling when $N_X \neq N_Y$ and $M = 1000$. X-axes are $\log(N_X)$, Y-axes are $\log(N_Y)$. (\textbf{A}) Variance of estimating CKD between random neural networks as in Section~\ref{sec:random-NNs}. (\textbf{B}) Monkey regions from \cite{papale_extensive_2025} as in \ref{sec:TVSD}.}
    \label{fig:diffN-rnn-monkey}
    
\end{figure}


    

In the main body of the paper, we present empirical verification of \ref{th:bound-CKD} in experiments where $N_X = N_Y$. 
For $N = N_X = N_Y$, we showed L-shaped level curves when plotting log-variance against $\log(M)$ and $\log(N)$, which are predicted by our theory of $\Var \sim \mathcal{O}(\frac1M + \frac1N)$.

In most practical neuroscience applications, $N_X \neq N_Y$. 
In Figure~\ref{fig:diffN-rnn-monkey}, we show variance scaling in this setting, fixing $M$ to be sufficiently large as to be negligible under our theory.
Similarly, our results in Figure~\ref{fig:diffN-rnn-monkey}(A) recapitulate the L-shaped curves from Section~\ref{sec:random-NNs}, verifying our theory.
Reassuringly, comparing monkey data to monkey data in Figure~\ref{fig:diffN-rnn-monkey}B also follows theory quite well.

\section{Cross-Region Comparisons in Monkey Data}
\label{app:cross-region}

\begin{figure}
    \centering
    \includegraphics[width=0.8\linewidth]{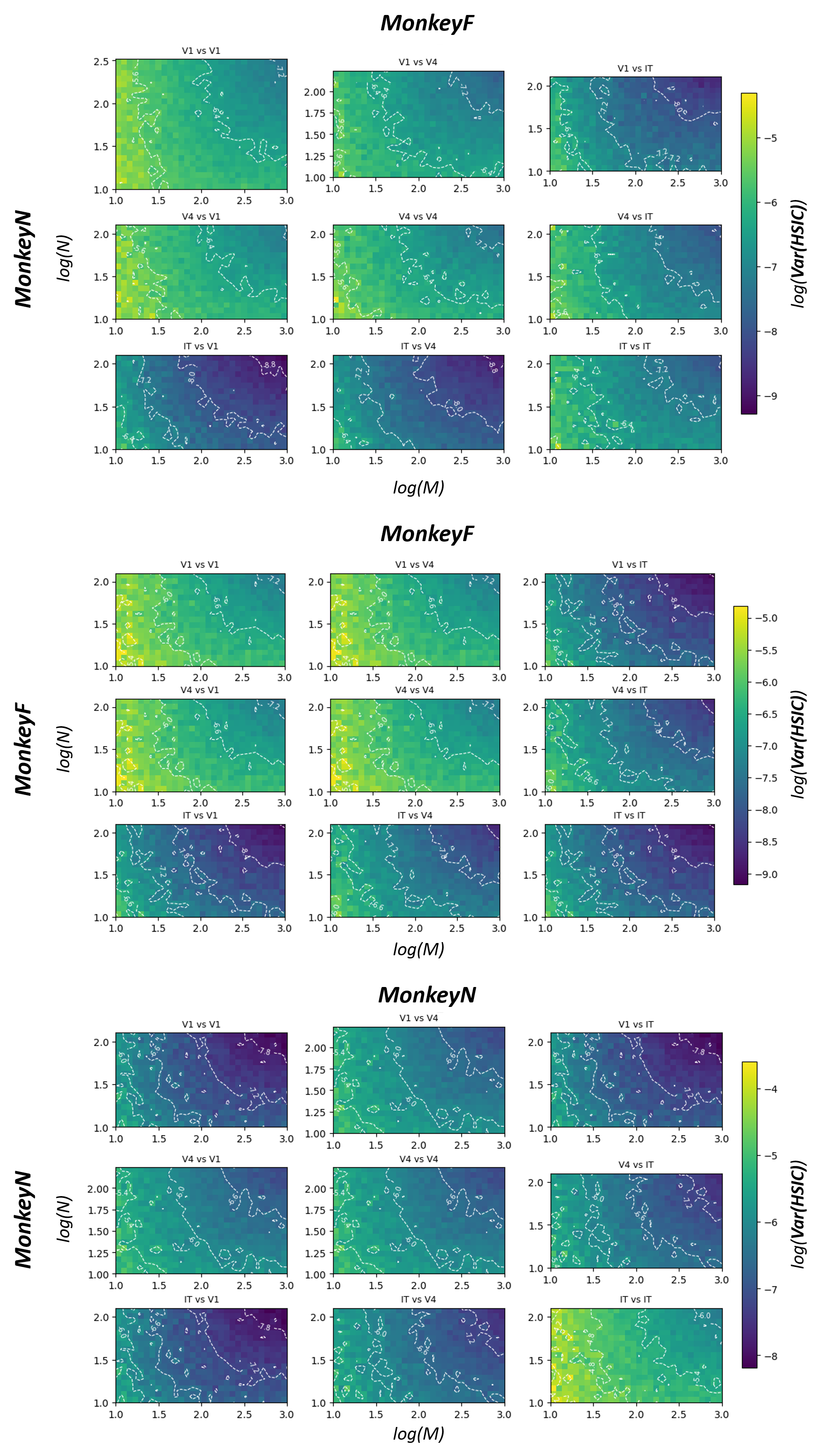}
    \caption{Variance of HSIC when subsampling across all regions between the two macaques available in the dataset.  Within a set of heatmaps, rows (left-to-right) and columns (top-to-bottom) are ordered as V1, V4, IT. Subplots are ordered as MonkeyF-to-MonkeyN, MonkeyF-to-Monkey-F, and MonkeyN-to-MonkeyN.}
    \label{fig:allTHINGS-HSIC}
\end{figure}

\begin{figure}
    \centering
    \includegraphics[width=0.8\linewidth]{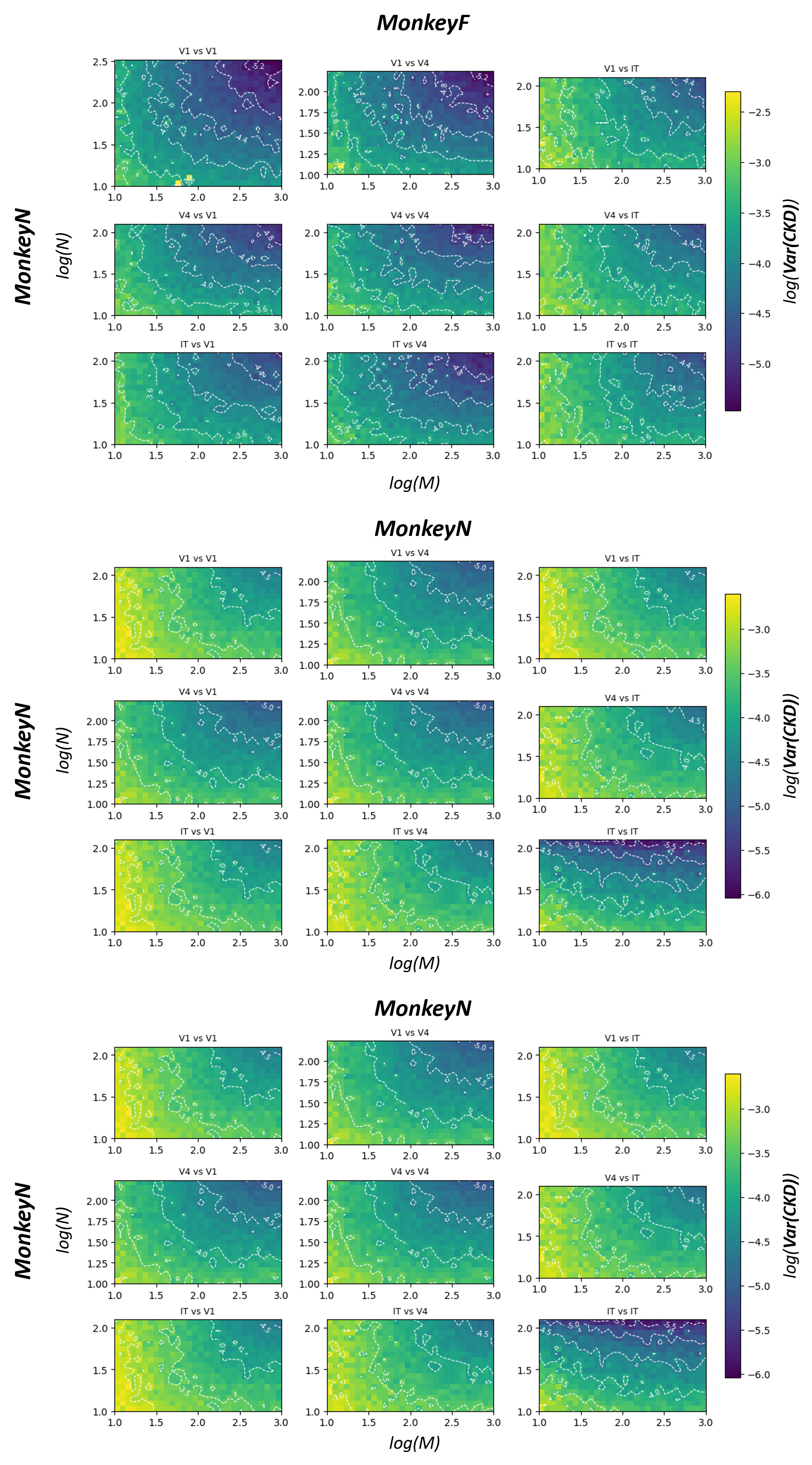}
    \caption{Variance of CKD when subsampling across all regions between the two macaques available in the dataset.  Within a set of heatmaps, rows (left-to-right) and columns (top-to-bottom) are ordered as V1, V4, IT. Subplots are ordered as MonkeyF-to-MonkeyN, MonkeyF-to-Monkey-F, and MonkeyN-to-MonkeyN.}
    \label{fig:allTHINGS-CKD}
\end{figure}

\begin{figure}
    \centering
    \includegraphics[width=0.8\linewidth]{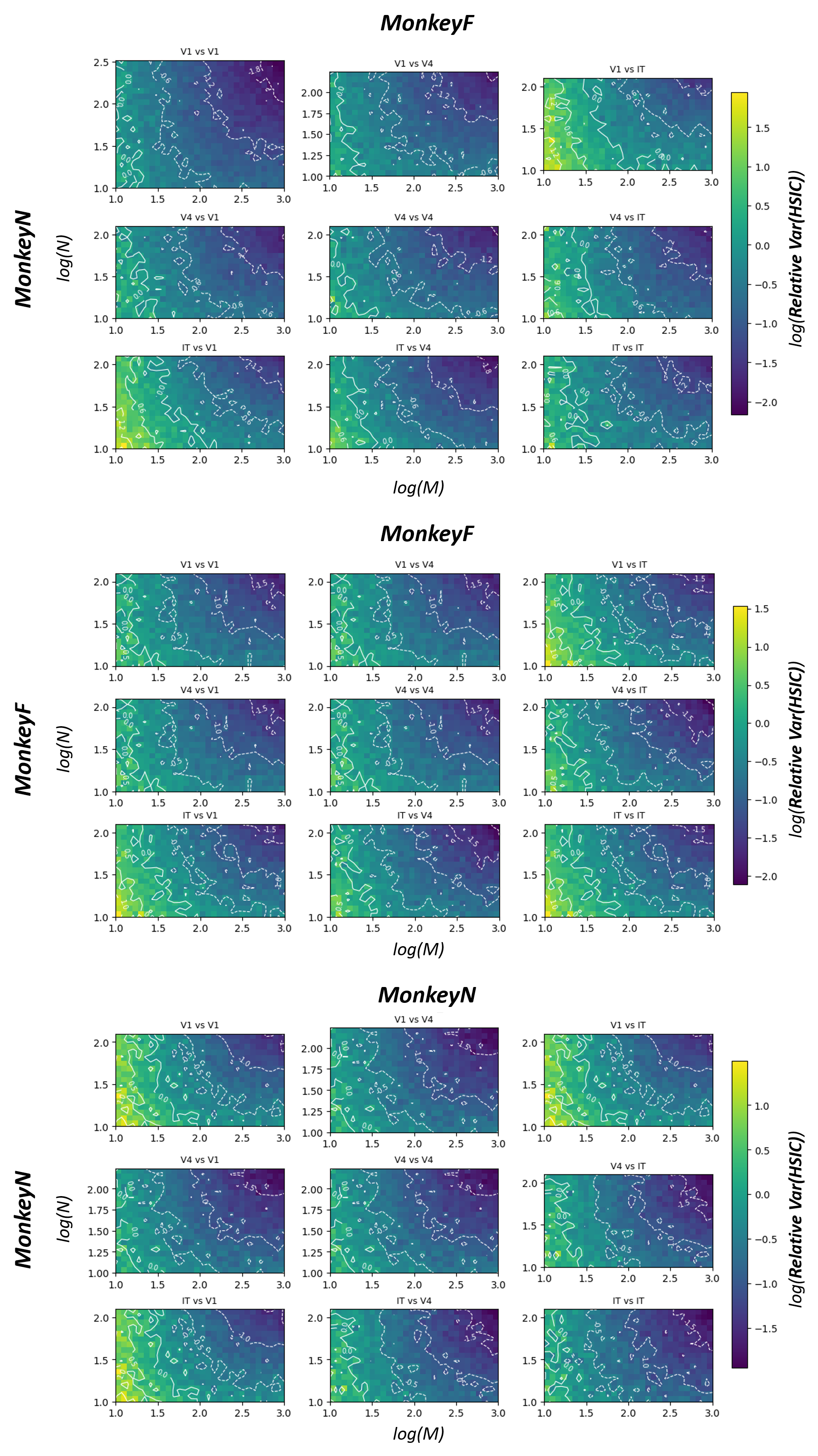}
    \caption{Relative variance of HSIC when subsampling across all regions between the two macaques available in the dataset.  Within a set of heatmaps, rows (left-to-right) and columns (top-to-bottom) are ordered as V1, V4, IT. Subplots are ordered as MonkeyF-to-MonkeyN, MonkeyF-to-Monkey-F, and MonkeyN-to-MonkeyN.}
    \label{fig:allTHINGS-HSIC-Rel}
\end{figure}

\begin{figure}
    \centering
    \includegraphics[width=0.8\linewidth]{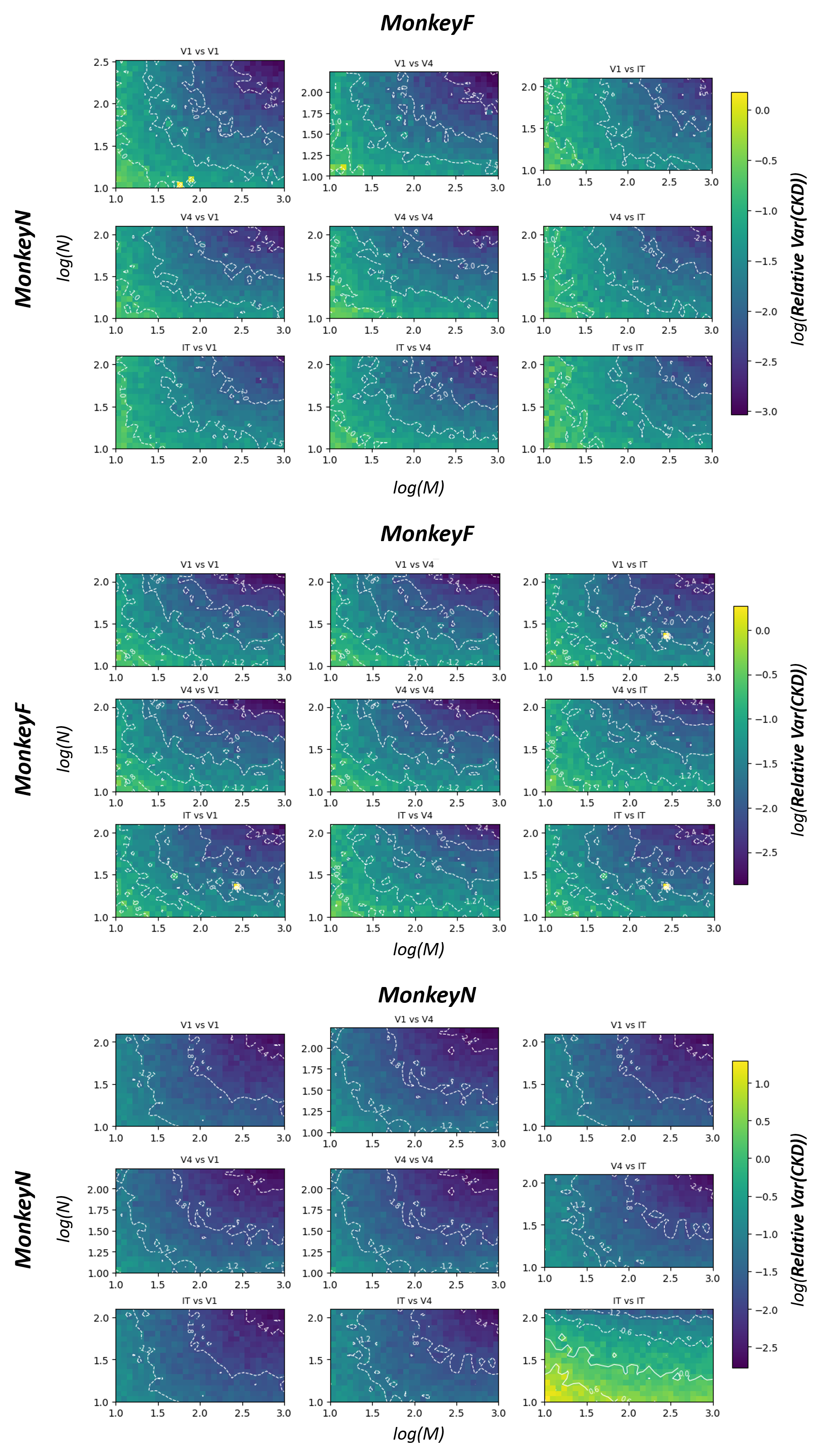}
    \caption{Relative variance of CKD when subsampling across all regions between the two macaques available in the dataset.  Within a set of heatmaps, rows (left-to-right) and columns (top-to-bottom) are ordered as V1, V4, IT. Subplots are ordered as MonkeyF-to-MonkeyN, MonkeyF-to-Monkey-F, and MonkeyN-to-MonkeyN.}
    \label{fig:allTHINGS-CKD-Rel}
\end{figure}

To further illustrate the region-independent nature of our theory, we extend our results from Section~\ref{sec:TVSD}.
In Figure~\ref{fig:models-monkeys}B, we plotted variance heatmaps of similarity estimates between the same region across two monkeys.
One may protest to our results' generality; two regions in the same species likely have similar underlying statistics.
In Figures~\ref{fig:allTHINGS-HSIC}, ~\ref{fig:allTHINGS-CKD},~\ref{fig:allTHINGS-HSIC-Rel},~\ref{fig:allTHINGS-CKD-Rel}, we show that our results can be generalized across arbitrary region-to-region estimation.



Almost universally, regardless of which region is compared to which region, regardless of whether we are comparing across monkeys or within monkeys, the variance estimates follow L-shaped curves as predicted by our theory.
Notably, the scale of relative variance appears inconstant for the most part regardless of the simulation type, as seen by traversing the corresponding heatmaps across the rows in each figure. 
Overall, these results reiterate that our theoretical predictions are reproduced in biological data, and serve as a prediction for how to efficiently collect neural data given a finite recording budget.
\section{Model-to-Monkey Comparisons}
\label{app:model-to-monkey}
\begin{figure}

    \centering
    \includegraphics[width=1.0\linewidth]{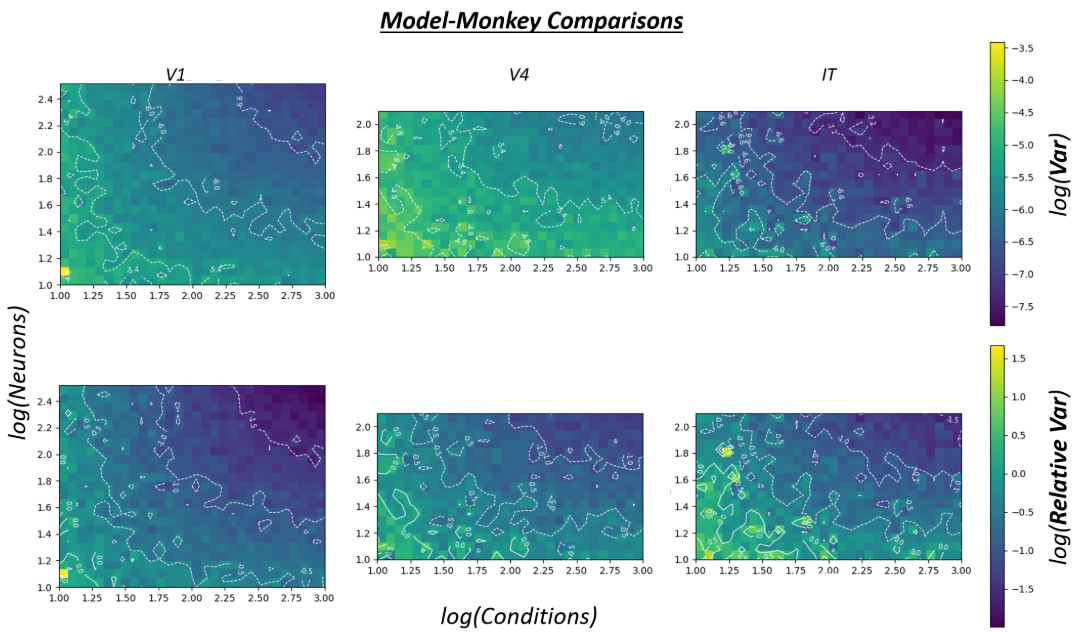}
    \caption{Variance of CKD between penultimate model layer and macaque cortex regions. Top row is pure variance of CKD, bottom row is relative variance of CKD (variance divided by true CKD value). Note that relative variance is constant across region identity.}
    \label{fig:model-to-monkey}
    
\end{figure}

To evaluate the universality of our claims, we examine the variance of CKD when comparing our trained deep networks to macaque cortex.
In \Cref{fig:model-to-monkey}, we plot the same heatmaps as before, where the top row is the variance of our CKD estimate and the bottom row is the relative variance of our CKD estimator.
Once again, we reproduce the L-shaped, evenly-spaced curves independent of region identity.
This further suggests that our theory applies universally regardless of the specific structure of the systems compared.
Furthermore, note that the specific value of relative variance appears constant regardless of region identity suggests that differences in the value of CKD variance is largely due to the scale of the underlying CKD score.
Hence, the \textit{proportion} at which one is erroneous in their CKD estimate is region-independent.

\section{Random Neural Network Construction}
\label{app:randomNN}
Consider two random neural networks $k, h$.
While their true similarity can be measured by sampling infinitely many $M$ or $N$, we want to study how subsampling affects our similarity estimate. 
We can compare the similarity between $k$ and $h$ by constructing representations $X$ and $Y$ by presenting $M$ inputs and extracting activations from only $N$ neurons, keeping the neuron identities constant through sampling. 

Consider a neural network $k$ with one hidden layer with weights; its outputs are arbitrary as we are only concerned with representations in this untrained hidden layer.
For 
\begin{align*}
    w &\sim \mathcal{N}(\mu_k, \sigma_k), \text{weights} \\
    \hat u &\sim \mathcal N (0, I_D), \text{inputs} \\
\end{align*}
over $M$ inputs and $N$ neurons.

Before presenting the inputs to the random network, we project the raw $\hat u$ onto the unit $D$-sphere (where we take $u = \frac{\hat u}{\Vert \hat u \Vert}$). 
With this, we can collect the neural representation for $k$
\begin{equation}\label{eq: rep-NN}
    X = \text{ReLU}(UW) \text{, } U = (u^\top_1, .., u^\top_M) \in \mathbb{R}^{M \times D}\text{, } W = (w_1, ..., w_N) \in \mathbb{R}^{D \times N}
\end{equation}

Similarly, we can construct neural representation for another shallow, random neural network $h$
\begin{equation*}
    Y = \text{ReLU}(UV)\text{, } V = (v_1, ..., v_N) \in \mathbb{R}^{D \times N}\text{, } v \sim \mathcal{N}(\mu_h, \sigma_h)
\end{equation*}
Thus, we can compare the similarity between $k$ and $h$ by measuring the similarity between their neural representations $X$ and $Y$ for any similarity measure of our choosing.

We choose this analysis because random neural networks in this setting have been well-studied as approximants of kernel functions in deep networks \citep{cho2009kernel}.
In this particular setting, the eigenstructure of the kernel functions has been well-studied in \cite{bietti2021deep}, which gives us access to theoretical values of HSIC (and thus CKD) in the infinite $M, N$ limit.

Last, as our inputs are bounded, the $q$-factor in our input admits a simple dimension-free bound.
It is simple to see why.
For any fixed input $u$, $w^\top u$ is Gaussian with mean at most $|\mu|$ in magnitude and variance $\sigma^2$, and since $|f(u)| \le |w^\top u|$, a short calculation gives $q \le 128\mu^4 + 384\sigma^4$ for every $D$.

\section{Image Dataset Details}
\label{app:datasets}

\begin{figure}
    \centering
    \includegraphics[width=1\linewidth]{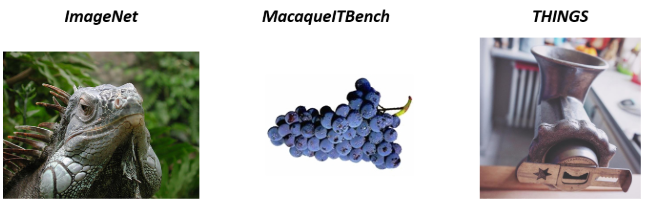}
    \caption{Representative images from the ImageNet dataset \citep{deng2009imagenet}, the MacaqueITBench dataset \citep{Madan2024MacaqueITBench}, and the THINGS Ventral Spiking Dataset \citep{papale_extensive_2025}. ImageNet and THINGS both comprise of naturalistic whole-scenes images. MacaqueITBench comprises of natural objects against a white background. Specific image categories intersect but are not identical between the three datasets.}
    \label{fig:datasets}
\end{figure}

We use three public datasets in this paper: the ImageNet dataset \citep{deng2009imagenet}, the MacaqueITBench dataset \citep{Madan2024MacaqueITBench}, and the THINGS Ventral Spiking Dataset \citep{papale_extensive_2025} (Figure~\ref{fig:datasets}).
The ImageNet dataset contains over 14 million naturalistic images with more than 20,000 unique semantic categories.
We use the ImageNet dataset to test and validate our deep networks.
Evaluation of performance and estimation of similarity between networks is done on a test set partition of the ImageNet dataset.

In Section~\ref{sec:deep-NNs}, we evaluate the effects of in- versus out-of-distribution evaluation data for similarity estimates on our deep networks.
We use the MacaqueITBench dataset as a proxy for out-of-distribution image data.
The MacaqueITBench dataset consists of 8233 unique natural images.
Since the deep models did not see these images during training or evaluation, we use the entire dataset as a pool of test images.
The images in the MacaqueITBench dataset differ markedly from the ImageNet dataset; while both datasets comprise of naturalistic images, MacaqueITBench images consist of a natural object on a white background, while ImageNet images are an entire natural scene.
The specific images and image categories also differ, so we use the MacaqueITBench dataset as a proxy for out-of-distribution natural images.

We use both image and macaque spiking data from the THINGS Ventral Spiking Dataset (TVSD) in Section~\ref{sec:TVSD}.
Image data in TVSD was collected in the THINGS database \citep{Hebart2023ThingsData}, consisting of 26,107 naturalistic visual images in whole scenes, similar to the ImageNet dataset.
The TVSD dataset presented these images to two macaques, recording from V1, V4, and IT using chronic Utah array, consisting of between 200-512 neurons per region.
The TVSD dataset is divided into a training set of 22,248 images presented 1 time each (12 images per category) and a test set of 100 images presented at 30 repetitions each.
Because our theory suggests that trial-to-trial variability does not significantly impact similarity estimates, to increase the number of $M$ we can work with, we use only images and data from the training set.

We threshold for mean reliability $> 0.3$ and use activations from the remaining neurons for comparison when subsampling; relatively few neurons are filtered out.
Recordings are taken over 4 days for one macaque and 7 days for the other, with pseudorandomized image presentations across sessions, although all images are presented.
In this paper, we treat the neurons recorded across sessions as identical.
Hence, we construct six $22248 \times N_{TOTAL}$ activation matrices, and we randomly bootstrap rows (subsampling $M$) and columns (subsampling $N$) to evaluate similarity comparisons over empirical $X \in \mathbb{R}^{M \times N_X}$ matrices.

\begin{figure}
    \centering
    \includegraphics[width=1.0\linewidth]{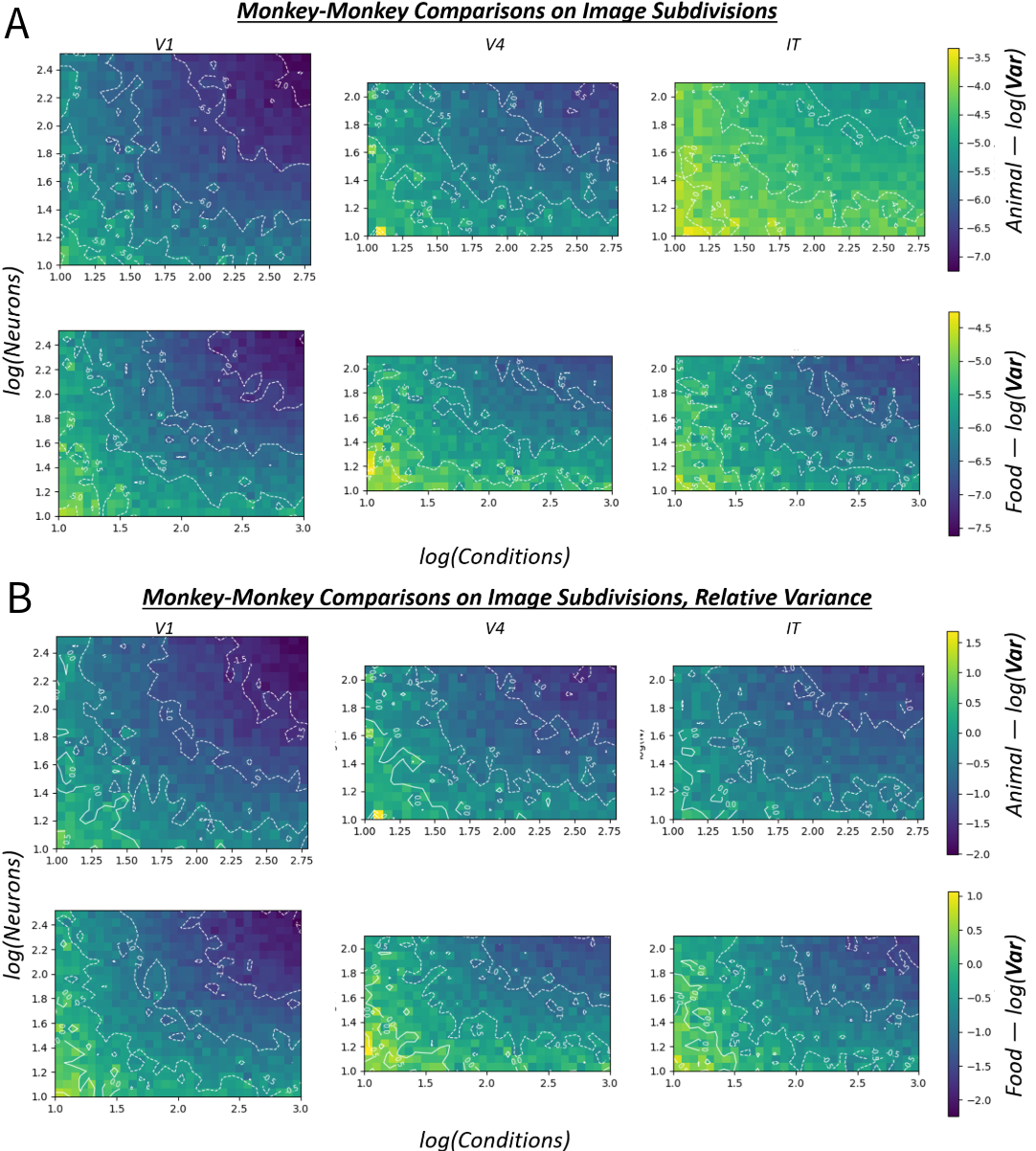}
    \caption{Variance of CKD estimates over specific image classes. (A) Variance of CKD estimates over animal (top) and food (bottom) images. L-shaped curves with consistent spacing found in \Cref{fig:models-monkeys} reproduced. (B) Relative variance of the experiments from (A).}
    \label{fig:monkey-subsets}
\end{figure}
Finally, to illustrate additional examples of how input distribution does not significantly affect variance scaling, we subset these above experiments to just certain classes of images (e.g. animals, foods, machines, toys, etc.).
Statistically, this is equivalent to conditioning our entire experiment on input distribution $\mathcal U$.
In \Cref{fig:monkey-subsets}, we show qualitative agreement with results from monkey-to-monkey simulations from \Cref{sec:deep-NNs}: L-shaped curves with consistent spacing regardless of region identity or input distribution.
Divergence between the two regions gets corrected when accounting for relative variance, so the proportion to which an estimate is "off" is near-independent of region identity; divergence can largely be chalked up to differences in teh scale of CKD between two input distributions, suggesting that the similarity in representation is not identical over different image classes.
\section{Model Training Details}
\label{app:model-training}

\subsection{Model Architecture}

We train self-supervised SimCLR models on ImageNet using an all-convolutional encoder with injected Gaussian noise after each convolution.

Let $x$ be an input image. The encoder produces a representation $h$ and the
projection head produces the contrastive embedding $z$:
\begin{align*}
h &= \mathrm{AllConvNet}(x), \\
z &= \mathrm{projection\_head}(h).
\end{align*}
Contrastive learning loss (Normalized Temperature-scaled Cross-Entropy loss, also known as NT-Xent) is computed on $z$.

After every convolutional layer, activations are injected with Gaussian noise at different parametrizations of $\sigma^2$, such that $x \to x + \mathcal{N}(0,\ \sigma^{2})$. In the main body of the paper, we focus on 5 models, with $\sigma^2 = 0, 0.01, 0.03, 0.09$.  through
Noise is sampled independently at each forward pass. 

\begin{table}[htbp]
\centering
\caption{AllConvNet encoder architecture. All convolutions use $3\times 3$
kernels and padding $1$. Channel width is $\texttt{layer\_dim}=96$. After
\texttt{conv7} the feature map is $2\times 2$ spatially; global average pooling
(\texttt{adaptive\_avg\_pool2d} to $1\times 1$) produces a $96$-D representation.}
Operations are all \texttt{Conv2d}, noise is always $\sigma^2$.
\label{tab:architecture}
\begin{tabular}{clcccc}
\toprule
Layer & In $\to$ Out & Stride & Activation \\
\midrule
1 & $3  \to 96$ & $1$ & ReLU \\
2 & $96 \to 96$ & $2$ & ReLU \\
3 & $96 \to 96$ & $2$ & ReLU \\
4 & $96 \to 96$ & $2$ & ReLU \\
5 & $96 \to 96$ & $2$ & ReLU \\
6 & $96 \to 96$ & $2$ & ReLU \\
7 & $96 \to 96$ & $2$ & --- \\
--- & Adaptive avg.\ pool & $96\times 2\times 2 \to 96$ & --- & --- & --- \\
\bottomrule
\end{tabular}
\end{table}

\subsection{Training Regime}

\begin{table}[htbp]
\centering
\caption{Training hyperparameters and compute setup.}
\label{tab:training}
\begin{tabular}{ll}
\toprule
Setting & Value \\
\midrule
Objective & SimCLR with NT-Xent (\texttt{lightly.loss.NTXentLoss}) \\
Optimizer & SGD, learning rate $0.06$, momentum $0.9$, weight decay $5\times 10^{-4}$ \\
LR schedule & \texttt{CosineAnnealingLR} over \texttt{epochs} \\
Epochs & $2$ \\
Batch size & $512$ \\
Input size & $128\times 128$ RGB \\
\bottomrule
\end{tabular}
\end{table}

Training uses the ImageNet train split via \texttt{LightlyDataset}, producing two independently augmented views per image. 
Validation uses the ImageNet val split, resized to $128\times 128$ and ImageNet-normalized, also subsampled at stride $2$.
Each batch is a pair of views $(x_0, x_1)$ of the same images. The model computes $z_0 = \mathrm{SimCLR}(x_0)$ and $z_1 = \mathrm{SimCLR}(x_1)$, then minimizes NT-Xent between positive pairs across the batch.

\subsection{Activation Extraction for Similarity Estimation}
\label{app:activation-extraction}
During evaluation of similarity estimates, we present $M$ randomly sampled images from a dataset of choice (for specifics on the datasets see Appendix~\ref{app:datasets}).
We sample $N$ random neurons from a desired layer (V1, V4, or IT in monkeys; second-to-last layer consisting of $96 \times 4 \times 4 = 1536$ neurons post-ReLU without additive Gaussian noise in trained neural networks)
Thus, we construct the $X \in \mathbb{R}^{M \times N_X}$ neural representation matrix.
The procedure for drawing neurons can be done either without replacement (subsampling) or with replacement (bootstrapping).
In this paper we present the bootstrapped variance estimates, but the results do not qualitatively differ between the two procedures, provided that $N$ is less than the total number of neurons in the layer; remarkably, even leaving just 1 out when sampling without replacement recapitulates these results qualitatively.
\section{Sampling Procedure for Section~\ref{sec:TVSD}}
\label{app:tvsd-sampling}
In Section~\ref{sec:TVSD}, at each value of $M$, we present $M$ random images to the 10 regions of study.
From each region, we randomly sample $N$ random neurons 10 times to create 10 copies of each region, and compare all pairwise estimates of HSIC, CKD, and CKA between the copies.
Hence, we construct a $100 \times 100$ matrix at each $(M, N)$ pair, for $M, N \in \{10, 33, 100, 1000\}$.
For the clusters, we apply MDS (10 dimensions) on the $100 \times 100$ matrix, then project onto the first two principal components.

Next, at each value of $(M, N)$ we randomly sample 1 copy from each model to generate a $10 \times 10$ matrix which we can compute a Kendall's $\tau$ against a reference $10 \times 10$ matrix containing the true distances between each pair of models.
We repeat this procedure 128 times and report the mean Kendall's $\tau$ in Figure~\ref{fig:clusters}.
In this paper, we use the upper-triangular Kendall's $\tau$, but using the row-wise Kendall's $\tau$ produces very similar results.

\end{document}